\documentclass[journal]{IEEEtran}

\usepackage{color, array, amsthm}
\usepackage{graphicx}
\usepackage{amssymb}
\usepackage{xcolor}

\usepackage{algorithm}
\usepackage{algpseudocode}
\usepackage{amsmath}

\begin{document}

\title{Real-Time Detection for Small UAVs: Combining YOLO and Multi-frame Motion Analysis}

\author{Juanqin Liu, Leonardo Plotegher, Eloy Roura, Cristino de Souza Junior, Shaoming He\textsuperscript{*}
\thanks{This work was supported by the Technology Innovation Institute under Contract No. TII/ARRC/2154/2023.}
\thanks{Juanqin Liu and Shaoming~He are with the School of Aerospace Engineering, Beijing Institute of Technology, Beijing 100081, China.}
\thanks{Leonardo Plotegher, Eloy Roura and Cristino de Souza Junior are with the Autonomous Robotics Research Centre, Technology Innovation Institute, P.O.Box: 9639, Masdar City, Abu Dhabi, United Arab Emirates.}
}

\maketitle

\begin{abstract}Unmanned Aerial Vehicle (UAV) detection technology plays a critical role in mitigating security risks and safeguarding privacy in both military and civilian applications. However, traditional detection methods face significant challenges in identifying UAV targets with extremely small pixels at long distances. To address this issue, we propose the Global-Local YOLO-Motion (GL-YOMO) detection algorithm, which combines You Only Look Once (YOLO) object detection with multi-frame motion detection techniques, markedly enhancing the accuracy and stability of small UAV target detection. The YOLO detection algorithm is optimized through multi-scale feature fusion and attention mechanisms, while the integration of the Ghost module further improves efficiency. Additionally, a motion detection approach based on template matching is being developed to augment detection capabilities for minute UAV targets. The system utilizes a global-local collaborative detection strategy to achieve high precision and efficiency. Experimental results on a self-constructed fixed-wing UAV dataset demonstrate that the GL-YOMO algorithm significantly enhances detection accuracy and stability, underscoring its potential in UAV detection applications.
\end{abstract}

\begin{IEEEkeywords} Small UAV detection, YOLO, Moving object detection, Global-Local collaborative detection
\end{IEEEkeywords}

\section{Introduction}
\begin{figure}[htbp]
	\centering
	\includegraphics[width=\linewidth]{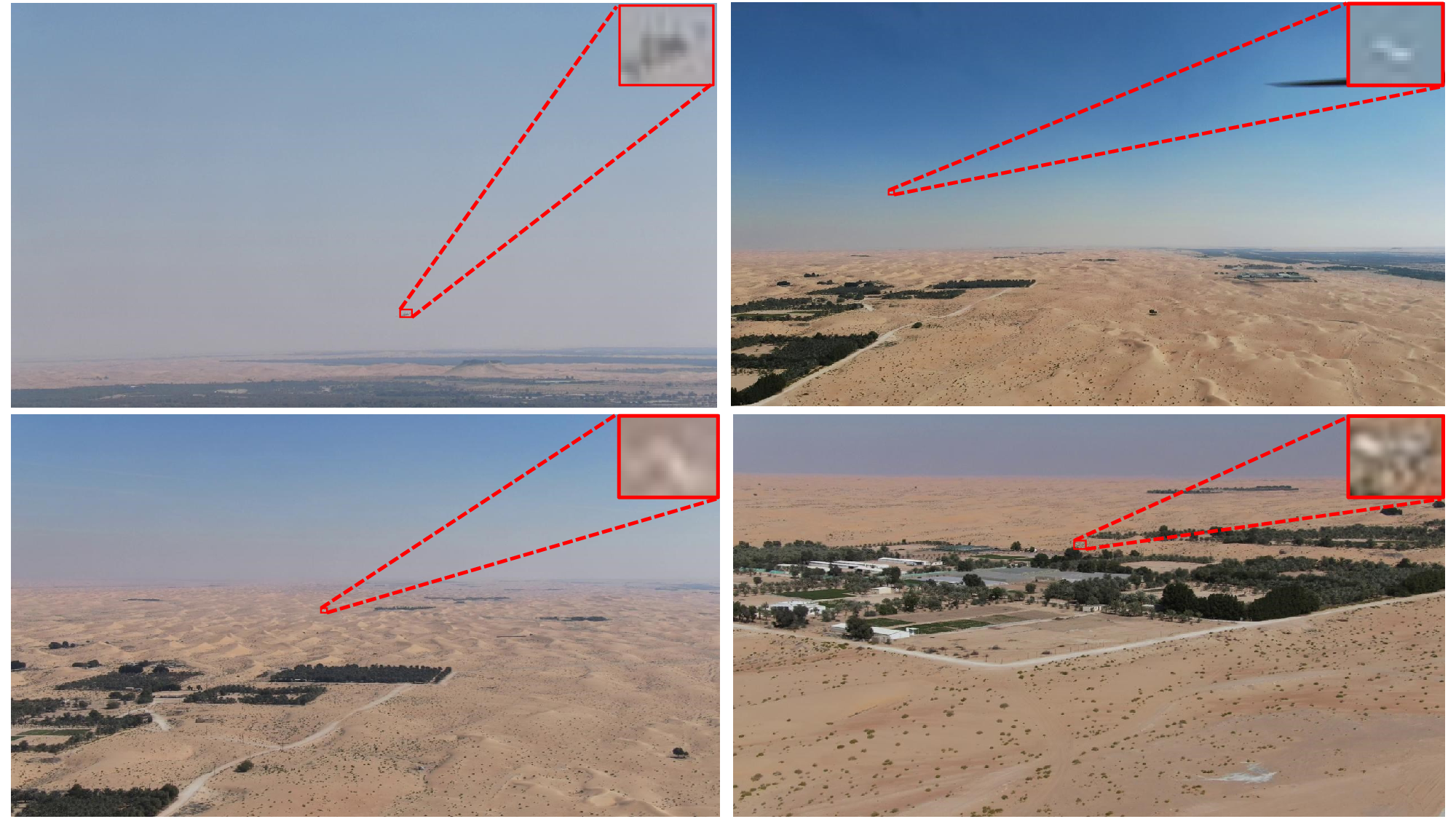}	
	\caption{Challenging conditions examples in UAV detections: 1) Minuscule targets with limited distinctive features; 2) Intricate backgrounds complicating target identification.}	
	\label{Fig1}	
\end{figure}

Since UAV technology emerged, its widespread application across various domains has raised significant safety risks and privacy concerns \cite{UAV1, UAV2, 2}. In response, the development of long-range drone detection technology has become critical, enabling the prompt identification, localization, and intervention of drones to safeguard public safety and personal privacy. However, existing detection technologies face considerable challenges when addressing small-pixel UAV targets at long distances \cite{4,5,6}. Drones typically occupy less than 0.1$\%$ of an image, leading to insufficient feature information. When combined with complex backgrounds, this results in decreased detection accuracy \cite{7,8}. Fig. \ref{Fig1} highlights some of the common challenges associated with detecting drone targets. 

While popular object detection methods like Fast-RCNN, YOLO, and DETR are highly effective for larger targets, they suffer from high false positive and false negative rates i+n long-range, small-object drone detection \cite{9,10,12}. Recent researchers have developed specialized detection methods optimized for drone characteristics \cite{15,17,43}. For example, by integrating appearance and motion features, using techniques like frame differencing or optical flow to extract moving objects, and applying classification methods to distinguish drone targets from other interfering objects. However, due to the diminutive size of drone targets, the amount of useful feature information available for classification is limited, and downsampling often leads to the loss of critical information. Additionally, the presence of diverse noise between drones and complex backgrounds makes it challenging to construct a precisely annotated dataset that can reliably distinguish targets from their surroundings.

In this paper, we propose the GL-YOMO detection algorithm, which effectively integrates appearance and motion features. By enhancing the YOLO model, we aim to improve detection accuracy while reducing computational complexity. The algorithm incorporates multi-frame motion detection for secondary validation, ensuring precise detection of small objects without requiring a manually constructed classification dataset. 

The main contributions of this paper are as follows:

1) \textbf{Development of the GL-YOMO Detection Algorithm}: This algorithm combines YOLO object detection with multi-frame motion detection, leveraging YOLO's efficient detection capabilities while incorporating motion feature capture to significantly enhance detection accuracy and stability.

2) \textbf{Improvement of the YOLO Model}: The enhancements increase detection accuracy and substantially reduce computational complexity and parameter count, resulting in a more efficient and lightweight model.

3) \textbf{Design of a Template Matching-Based Motion Detection Algorithm}: By analyzing pixel changes and displacement variations across three consecutive frames, this algorithm effectively detects extremely small objects, further improving small-object detection accuracy.

4) \textbf{Construction of the Fixed-Wings Dataset}: This dataset includes 13 video sequences and 24,951 frames, encompassing numerous UAV targets with an average image proportion of 0.01\%, providing a robust resource for evaluating UAV detection algorithms.

The paper is structured as follows: Section II reviews small object detection and related methods for UAV detection. Section III details our proposed method. Section IV presents experimental results and analysis. Section V concludes the paper and discusses future research directions.

\section{Related Work}
\subsection{Small Object Detection Methods}

Since the introduction of the YOLO algorithm in 2016, its subsequent versions \cite{26,27, TPH-YOLO} have continually evolved, driving substantial advancements in object detection. To tackle the challenges associated with small object detection, many different useful strategies were proposed in recent works. Among these, multi-scale feature fusion emerged as a critical approach, effectively integrating semantic information across various levels to improve small object detection \cite{c,d,30,31,yang2024fatcnet}. For example, Gold-YOLO \cite{30} enhanced feature fusion by incorporating a Gather-and-Distribute mechanism, achieving a 39.9\% AP on the COCO val2017 dataset, which is a 2.4\% enhancement over the prior state-of-the-art model, YOLOv6., while ASF-YOLO \cite{31} significantly improved small object detection and segmentation through the Scale Sequence Feature Fusion (SSFF) module and the Triple Feature Encoder (TPE) module.

Feature enhancement strategies were also pivotal, with many studies introducing attention mechanisms to amplify target features while suppressing background noise, thereby boosting detection accuracy. Notably, CEAM-YOLOv7 \cite{32} incorporated global attention mechanisms in both the Backbone and Head, improving the mAP by 20.26\% than the original YOLOv7 model. Gong et al. \cite{k} introduced a normalization-based attention module that incrementally improved detection by focusing on channels and spatial dimensions through coefficient penalization, achieving a 7.1\% improvement on the DOTA dataset. The AIE-YOLO \cite{e} employed a context enhancement module that fuses multi-scale receptive fields with attention mechanisms to optimize feature representation. Additionally, super-resolution techniques led to considerable improvements \cite{f,j}. For instance, Yucong et al. \cite{f} combined shallow high-resolution geometric details with deep super-resolution semantic features, augmented by channel attention, to significantly elevate detection precision. Collectively, these innovations represented a substantial leap forward in the field of small object detection.

\begin{figure*}[htbp]
	\centering
	\includegraphics[width=\linewidth]{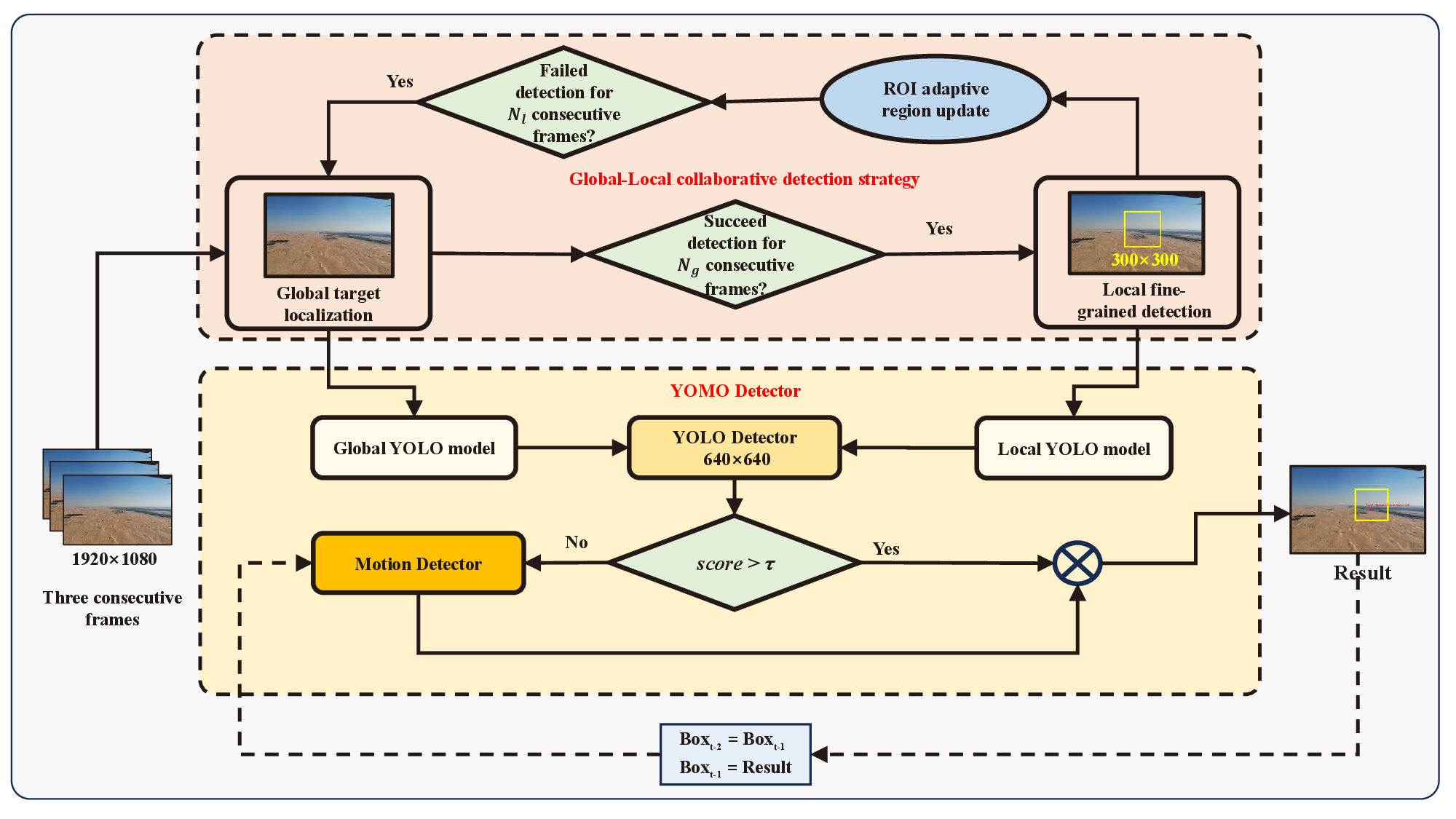}	
	\caption{The GL-YOMO architecture: The YOMO Detector combines YOLO Detector with a multi-frame Motion Detector. A global-local strategy and ROI adaptive update enable continuous detection of small UAV targets.}	
	\label{Fig2}	
\end{figure*}

\subsection{UAV Detection Methods}
Ensuring real-time performance in UAV detection while addressing the challenges posed by small objects and multi-scale variations is crucial. The YOLO series proved to be one of the most effective solutions for these challenges\cite{wang2024low,35,36}. For example, the work in \cite{10} integrated the Spatial Pyramid Pooling module into the YOLOv4 model’s prediction head, combining it with a spatial attention mechanism to achieve real-time drone detection at 14.9 fps. Similarly, the authors in \cite{cai2022lightweight} leveraged the lightweight MobileNet as the backbone, incorporating depthwise separable convolution techniques, and reported a processing speed of 82 fps along with an accuracy of 93.52\% mAP on a custom UAV dataset. Although these methods excelled at extracting prominent features from single frames and performed well in simpler scenarios, they encountered difficulties in more complex environments.

To better handle challenges such as background interference and target occlusion in more intricate environments, researchers explored motion-based detection methods for drone targets. For instance,  a motion detector that employed background subtraction was proposed in \cite{37}, leveraging post-processing to identify moving objects and utilizing the MobileNetV2 classifier for UAV classification, achieving an accuracy metric of 70.1\% on the Drone-vs-Bird dataset \cite{Drone-vs-Bird}. However, this approach was limited in scenarios involving moving cameras. To address this limitation, the work in \cite{38} introduced an optical flow-based motion information extractor that replaced the upsampling component in a standard Feature Pyramid Network, thereby significantly improving UAV detection accuracy, resulting in an impressive average precision(AP) of 67.8 on the Drone-vs-Bird dataset. Additionally, the studies in \cite{39, 40} exploited optical flow estimation to generate candidate points relative to background motion, estimated background dynamics using a global motion model and applied background subtraction in conjunction with a hybrid detector for UAV classification. Low-rank methods, as presented in \cite{1,41}, contributed to the field by generating small motion regions that facilitated drone target classification. Moreover, integrating visual and motion features was shown to enhance UAV detection. For example, the authors in \cite{4} combined YOLO detectors with a motion target classifier through a global-local detection approach, 92\% accuracy and 23.6FPS performance were achieved on the custom ARD-MAV dataset. The approach in \cite{42} achieved precise detection using multi-channel temporal frames and spatiotemporal semantic segmentation with convolutional neural networks, alongside a ResNet classifier, which resulted in an average F1-score of 0.92 across three test videos of the drone-vs-bird dataset. Collectively, these approaches represented significant advancements in UAV target detection within complex environments.

\section{Methods}

We propose a high-precision and robust detection method called GL-YOMO, as depicted in Fig. \ref{Fig2}. This method utilizes a global-local collaborative detection strategy. The process begins with object detection and localization across the entire $1920\times1080$ global frame. Once an object is consistently detected across consecutive frames, the system narrows its focus to a $300\times300$ local region for more detailed detection. The detection is carried out by our customized YOMO detector, which combines the YOLO detector with multi-frame motion detection technology to enable efficient object capture and tracking. To ensure stable and continuous detection of small UAV targets, an Region of Interest (ROI) adaptive update mechanism is integrated, dynamically adjusting the ROI to continuously refresh the detection scope.

\subsection{Global-Local Collaborative Detection Strategy}

While global detection offers broad coverage, it is susceptible to background noise and environmental interference, increasing the likelihood of false negatives and false positives. In contrast, local detection, with its narrower focus, is better suited for accurately capturing target locations. To enhance detection accuracy, this study adopts a global-local collaborative strategy, incorporating two key components: dynamic switching between global and local detection modes and adaptive updating of the ROI.

\subsubsection{Dynamic Switching}
The dynamic switching strategy between global and local detection modes is based on an analysis of target detection frame rates. Initially, the system performs global target localization. Once a target is consistently detected in ${N_g}$ consecutive frames, the system automatically transitions to local mode for more refined detection. If no targets are detected within ${N_l}$ consecutive frames in local mode, the system reverts to global mode for re-localization. By establishing appropriate frame thresholds (${N_g}$ and ${N_l}$), this approach enables seamless switching between global and local detection modes, thereby enhancing the system's stability and reliability for UAV target detection in complex environments.

\subsubsection{ROI Adaptive Region Update}
The ROI adaptive updating is crucial for defining the local detection range and directly impacts target detection effectiveness. When transitioning from global to local detection modes, the system crops an ROI, focusing the detection range on an initial $300\times300$ pixel area for more refined local detection. To accommodate UAV target movement, a dynamic strategy is employed based on the target's proximity to the ROI edge. If the target remains within ${R_s}$ of the ROI's radius, the ROI remains unchanged; if the target moves beyond this range, the ROI is automatically updated to center on the target. This approach improves fault tolerance and reduces errors caused by frequent ROI updates. Additionally, to mitigate potential missed detections, the system expands the ROI based on missed frames, ensuring effective detection while preserving the advantages of local detection and minimizing the loss of global detection accuracy, as outlined in (1).
\begin{equation}
{R_{size}} = 300 + {k_1} \cdot {F_{lost}}
\label{eq1}
\end{equation}
where ROI is defined by an area of $R_{size} \times R_{size}$, $k_1$ is a proportional coefficient used to control the rate of ROI expansion, and $F_{lost}$ represents the number of consecutive frames lost.

\begin{figure*}[htbp]
	\centering
	\includegraphics[width=\linewidth]{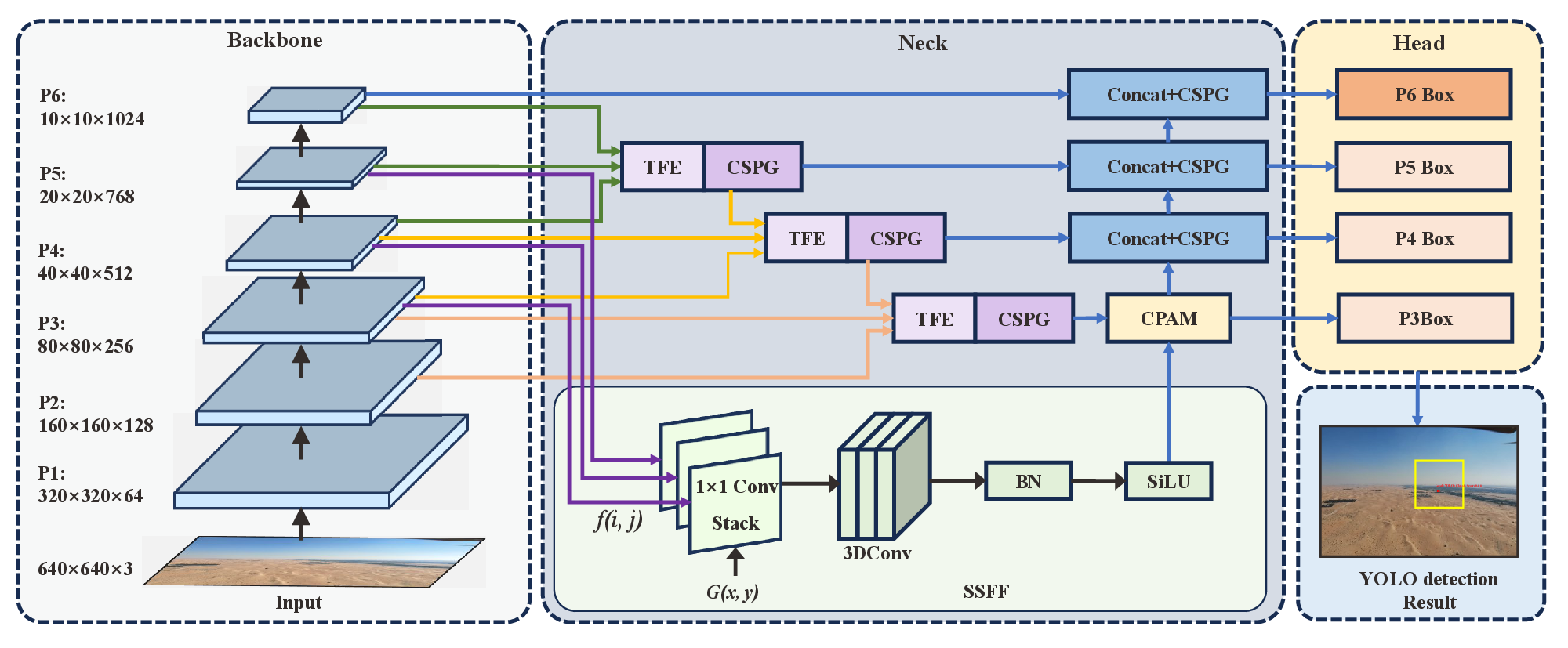}	
	\caption{YOLO network model architecture diagram: The model integrates SSFF, TFE, and CPAM modules from ASF-YOLO and replaces the C3 structure in the CSP module with C3Chost, forming the CSPG module. The model utilizes four detection heads to accommodate multi-scale object detection tasks.}	
	\label{Fig3}	
\end{figure*}

\subsection{YOMO Detector}

The YOMO Detector, a central method for detecting small UAV targets, comprises two key components: the YOLO Detector and the Motion Detector. The YOLO Detector is known for its efficiency in target detection, while the Motion Detector captures the dynamic characteristics of targets. Together, these components work synergistically to significantly improve detection accuracy.

\subsubsection{YOLO Detector}
Although the YOLO algorithm is highly effective in a range of object detection tasks, it faces challenges when dealing with very small objects due to feature loss during downsampling. To overcome this limitation and inspired by ASF-YOLO \cite{31}, we introduce a new multi-scale feature fusion approach with its architecture illustrated in Fig. \ref{Fig3}.

In the YOLO Detector, both the global detection and local detection phases involve resizing the input to $640\times640$ pixels, this adjustment was made to balance inspection accuracy with processing speed, ensuring high performance and reducing the computational burden, and multi-scale feature extraction is leveraged through successive downsampling in the Backbone. The backbone network is based on the YOLOv5 CSPDarknet53 architecture, utilizing stacked C3 modules and the Conv-BN-SiLU structure for feature extraction. This Backbone module generates four crucial feature layers: P3, P4, P5, and P6, each capturing features at different scales. These feature layers are then fed into the Neck for further multi-scale feature fusion. In the Neck, We have integrated the TFE module, as proposed by ASF-YOLO, which ingeniously fuses feature maps of large, medium, and small scales, concatenating them along the channel dimension. This approach preserves the fine-grained information crucial for detecting small objects. Additionally, the SSFF module employed Gaussian smoothing techniques, progressively increasing the standard deviation to process feature maps. It then utilizes 3D convolution technology for stacking and processing, enabling the model to better handle objects of varying sizes, orientations, and aspect ratios while capturing scale-space relationships. Subsequently, we combine the Channel and Position Attention Mechanism (CPAM) with the SSFF and TFE modules, CPAM captures cross-channel interactions through its channel attention mechanism without dimension reduction, while its position attention mechanism processes feature maps along horizontal and vertical axes separately. This allows the model to adaptively focus on small objects across different scales. The incorporation of multi-scale feature fusion and attention mechanisms allows the model to more accurately detect extremely small objects in drone imagery. The Head is a critical part of the YOLO module for object detection, and four detection heads are used, corresponding to the feature maps P3, P4, P5, and P6 at different scales. The Head section’s multi-scale feature fusion allows the model to detect objects of various sizes, enhancing its performance in complex scenes and multi-scale target scenarios.

In terms of model efficiency, we have integrated the efficient Ghost module \cite{44} into various layers of the YOLO model's Backbone and Neck. This includes replacing traditional convolution operations with GhostConv and substituting the original C3 module with the C3Ghost module. These modifications significantly reduce the model's computational load and parameter count without compromising performance.

Based on this enhanced YOLO model architecture, the dataset is used for training to produce the required YOLO models. It is important to emphasize that the training for the YOLO models in the global and local detection stages is distinct in focus. During the global detection stage, the model is trained on full images to create the Global YOLO Model, enabling comprehensive detection of targets across the entire image. Conversely, in the local detection stage, the model is trained on cropped images, resulting in the Local YOLO Model, which is specifically tailored for more precise target recognition within the ROI.

\subsubsection{Motion Detector}

\begin{figure*}[htbp] 
	\centering
	\includegraphics[width=\linewidth]{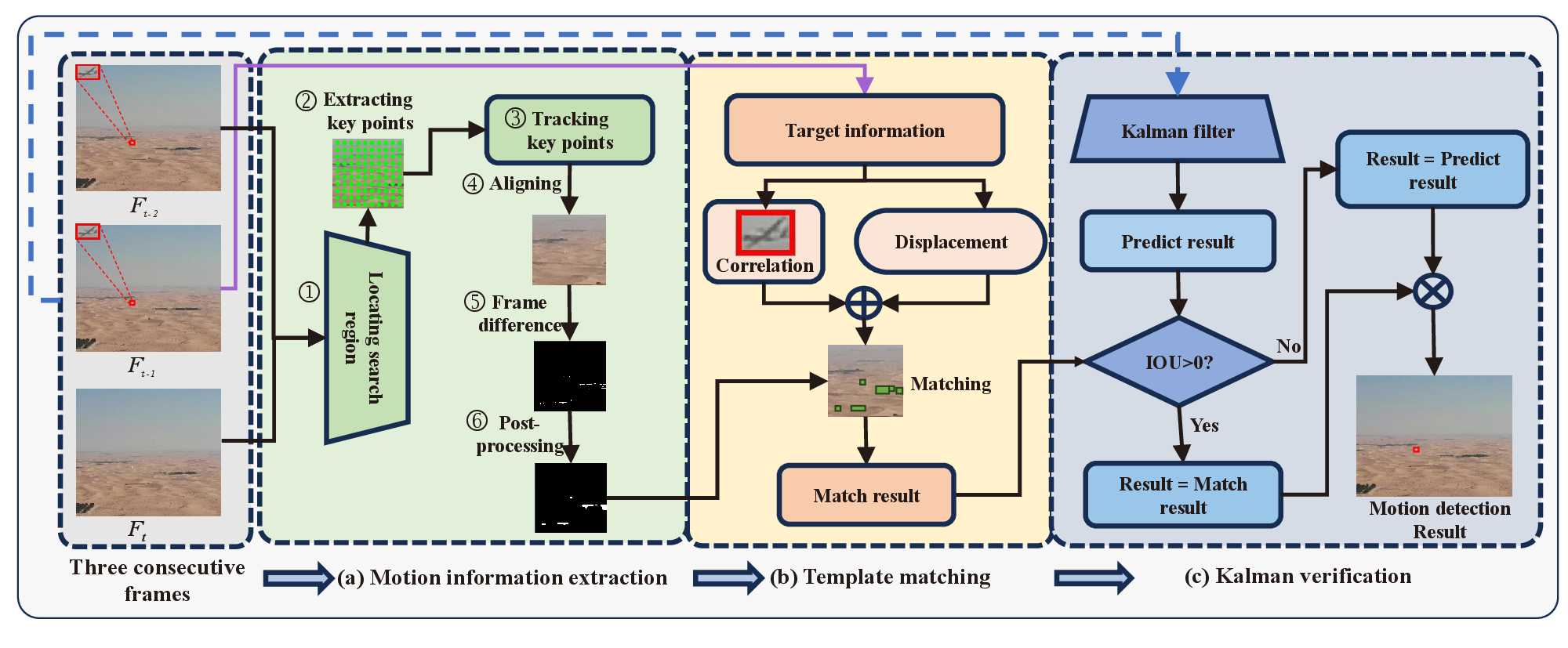}	
	\caption{Motion detection algorithm framework diagram: Three consecutive frames undergo three key stages: motion information extraction, template matching, and Kalman verification, to obtain the final motion detection result.}	
	\label{Fig4}	
\end{figure*}

When the confidence score of the YOLO detector falls below a predefined threshold, the system turns to the Motion Detector for further analysis. The reliability of the YOLO detection results is determined by the confidence threshold $\tau$. In global detection, if the YOLO confidence score exceeds the threshold $\tau_g$, the detection result is considered valid; otherwise, it is deemed invalid, prompting the system to activate the motion detector for supplementary analysis. Similarly, in local detection, if the confidence score exceeds the threshold $\tau_l$, the detection result is considered valid; if not, the system will initiate the motion detector for further detection. The motion detector integrates optical flow methods, template matching techniques, and Kalman filtering algorithms to comprehensively analyze the motion information between frames. The implementation steps are detailed in Fig. \ref{Fig4}, where ${F_{t}}$ denotes the frame at time instant $t$.

\begin{figure}[htbp]
	\centering
	\includegraphics[width=\linewidth]{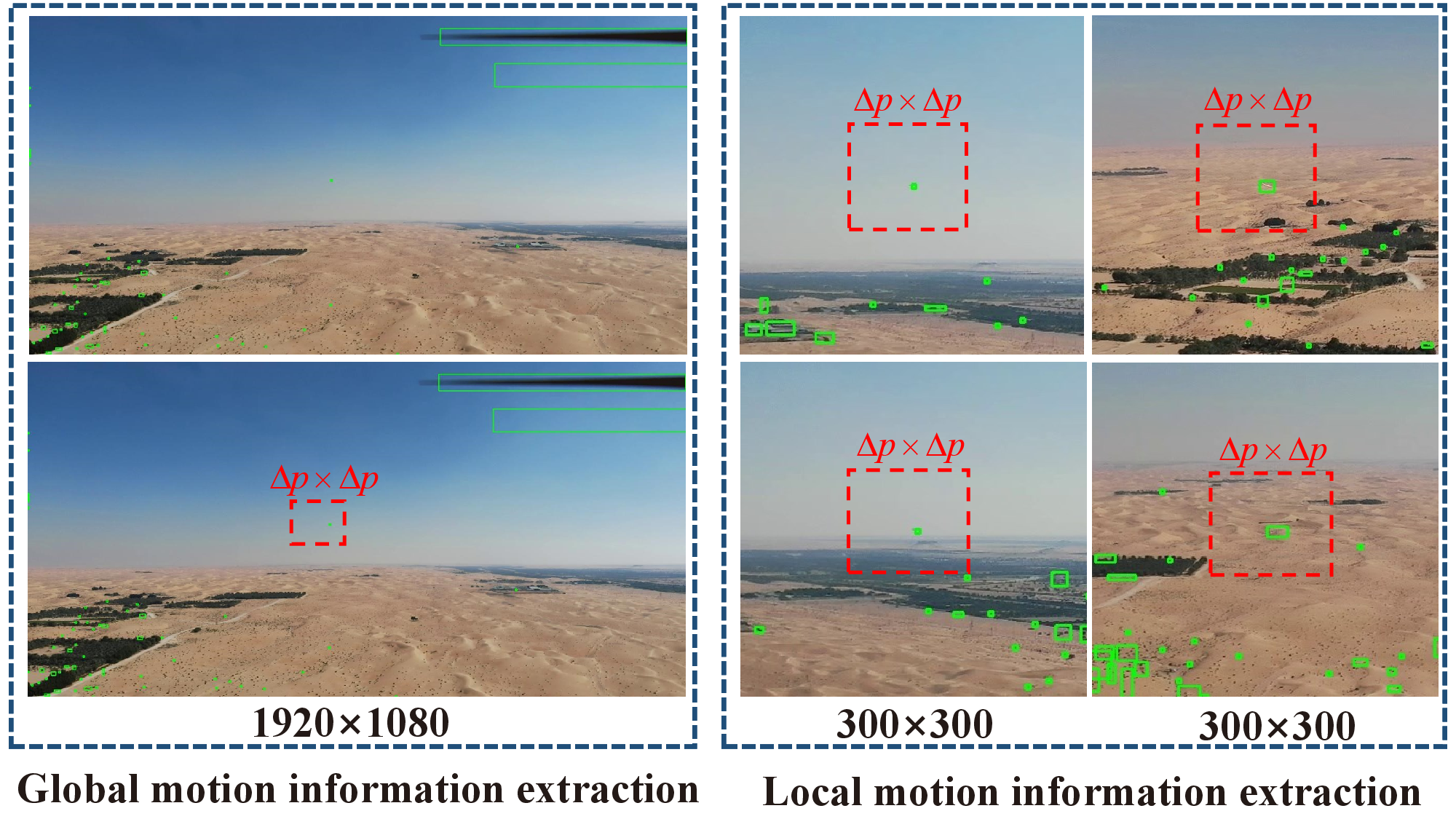}	
	\caption{Examples of motion information extraction at global and local scales: The green boxes highlight regions containing moving targets or noise, where distinguishing between small UAV targets and noise is challenging. The red boxes denote the motion information extraction areas defined by our method, which effectively minimizes interference from extraneous noise.}	
	\label{Fig5}	
\end{figure}

\textbf{(a) Motion Information Extraction.} To accurately extract motion information, it is crucial to isolate moving objects from dynamic backgrounds, forming the foundation for subsequent template-matching algorithms. We employ a frame interval-based strategy, utilizing the interval between ${F_{t-2}}$ and ${F_{t}}$ to capture motion information. This approach reveals motion changes more distinctly than using consecutive frames, particularly for small UAV targets. According to Fig. \ref{Fig4}(a), the process of extracting motion information is as follows:

To retain small-sized UAV targets while minimizing noise, motion information is extracted from a localized region around the target position rather than the entire image. Using the target coordinates from ${F_{t-2}}$ as the center, crop a $\Delta p \times \Delta p$ pixel region (e.g., $\Delta p=50$) for optical flow extraction. This region size, determined through extensive experimentation, adequately encompasses the target's motion range over three frames, ensuring accurate and effective motion capture while avoiding unnecessary noise and enhancing processing efficiency. Partial results of motion feature extraction are shown in Fig. \ref{Fig5}.

The extraction method leverages optical flow to track motion. Key points within the extraction region are first identified using a grid-based approach, which divides the area into a grid and selects key points at the intersections to capture critical motion features. The pyramidal Lucas-Kanade method is then employed for multi-scale analysis of optical flow, allowing precise tracking of key point trajectories and subtle motion changes \cite{tran2018pyramidal}. To account for camera movement or perspective shifts, a 2D perspective transformation is applied to the previous frame ${F_{t-2}}$, which maps points from one plane to another, aligning the content of ${F_{t-2}}$ geometrically with ${F_t}$. This alignment compensates for motion unrelated to scene dynamics, ensuring that static elements in the scene are closely matched. After alignment, frame difference is performed to isolate genuine scene changes, following a process similar to that outlined in \cite{4}. Significant differences between frames are identified by computing pixel-wise differences between the aligned frame and the current frame. To further enhance the results, corrections for lighting variations and background motion are applied, followed by binarization to emphasize regions of motion. The frame difference identifies pixels corresponding to moving objects, but the output often contains noise and lacks smoothness. therefore post-processing techniques are necessary for noise suppression and shape optimization. First, morphological operations such as erosion and dilation are used to refine the contours of detected objects, improving their structure. Next, median filtering and Gaussian blurring are applied to further reduce noise and smooth the object boundaries. After noise reduction, a thresholding operation is performed to create a clean binary image.  Finally, connected component analysis is used to segment individual objects, enabling the extraction of relevant attributes including their location and size.

\textbf{(b)	Template Matching}. Given the high consistency of the target across consecutive frames, we adopt a template-matching approach to solve the target tracking problem. The target detected in the previous frame is used as a template and matched with the motion region extracted from optical flow analysis in the current frame. To enhance matching accuracy, we introduce a weighted matching mechanism that combines correlation metrics with displacement information from three consecutive frames, providing a more comprehensive assessment of the target’s matching probability.

Specifically, the normalized cross-correlation coefficient is employed to measure the visual similarity between the target region and the template as
\begin{equation}
{NCC_{(x,y)}} = \frac{{\sum\limits_{u,v} {({T_{(u,v)}}} - \bar T)({I_{(x + u,y + v)}} - \overline {{I_{(x,y)}}} )}}{{\sqrt {\sum\limits_{u,v} {({T_{(u,v)}} - } \bar T{)^2}\sum\limits_{u,v} {({I_{_{(x + u,y + v)}}}} - \overline {{I_{(x,y)}}} {)^2}} }}
\label{eq2}
\end{equation}
which is then normalized to the range $[0,1]$, as 
\begin{equation}
N_c = \frac{{NCC_{(x,y)}} + 1}{2}
\label{eq3}
\end{equation}
where $T_{(u,v)}$ denotes the pixel value of the template image at coordinate $(u,v)$, $\bar{T}$ represents the mean value of the template image, $I_{(x+u, y+v)}$ denotes the pixel value of the original image at location $(x+u, y+v)$, and $\overline{I_{(x,y)}}$ represents the mean pixel value of the region in the original image with $(x,y)$ as the top-left corner and the same size as the template.

To account for potential variations in target size, a multi-scale visual similarity evaluation is performed for each candidate region. The candidate region is scaled up or down according to three different scale factors, and the visual similarity between the target and the previous frame is computed at each of these scales. The final similarity $C_c$ is defined as the maximum normalized cross-correlation $N_c$ across the different scales as
\begin{equation}
C_c = \max\left\{ N_c^{\text{scale}=0.7}, N_c^{\text{scale}=1.0}, N_c^{\text{scale}=1.3} \right\}
\label{eq4}
\end{equation}
where $N_c^{\text{scale}}$ represent the normalized cross-correlation value at each respective scale.

Define ${C_{d}}$ as the similarity of the displacement between the current target and the target in the previous frame, and ${C_{d}}$ is determined by
\begin{equation}
C_d = 1 - (\Delta {d_{norm}}+ \Delta {\theta _{norm}})/2
\label{eq5}
\end{equation}
where $\Delta {d_{norm}}$ and $\Delta {\theta _{norm}}$, respectively, represent the normalized distance difference and direction difference, which are determined by
\begin{equation}
\Delta {d_{norm}} = \frac{|d_{F_t} - d_{F_{t-1}}|}{\sqrt{w_t^2 + h_t^2}}
\label{eq6}
\end{equation}
\begin{equation}
\Delta {\theta_{norm}} = \frac{|\theta_{F_t} - \theta_{F_{t-1}}|}{\pi}
\label{eq7}
\end{equation}
where $d_{F_t}$ is the pixel Euclidean distance between the current frame and its preceding frame, i.e., the pixel distance between two bounding box center points. We use the image dimensions \(w_t\) and \(h_t\) of the current frame to normalize $\Delta {d_{norm}}$. Likewise, the normalized direction difference $\Delta {\theta _{norm}}$ is derived by normalizing the directional change between two consecutive frames. Define $(x_{c,t}, y_{c,t})$ as the pixel position of the center of the bounding box at time instant $t$, the Euclidean distance $d_{F_t}$ and directional change $\theta_{F_t}$ are given by
\begin{equation}
d_{F_t} = \sqrt{(x_{c,t}-x_{c,t-1})^2 + (y_{c,t}-y_{c,t-1})^2}
\label{eq8}
\end{equation}
\begin{equation}
\tan\theta_{F_t} = \frac{y_{c,t}-y_{c,t-1}}{x_{c,t}-x_{c,t-1}}
\label{eq9}
\end{equation}

The final weighted matching cost, denoted by $C_w$, can then be given by
\begin{equation}
{C_{w}} = {k_2}\cdot C_c + {k_3}\cdot C_d
\label{eq10}
\end{equation}
where the constants \(k_2\) and \(k_3\) are weighting coefficients used to balance the impact of $C_c$ and $C_d$ in the final matching result.

\textbf{(c)	Kalman Filter Verification}. To enhance detection accuracy and robustness, we introduce the Kalman filter as a verification mechanism for state estimation. Throughout the target detection process, we use the 8-state Kalman filter to predict the target state in the next frame. The state vector $\mathbf{x}_t = \left[x, y, w, h, v_x, v_y, v_w, v_h \right]^T$ includes the target position, size, and rate of change. Here, $(x, y)$ represent the top-left coordinates of the target bounding box, $(w, h)$ its width and height, $(v_x, v_y)$ the velocities, and $(v_w, v_h)$ the rates of change in width and height. We leverage the output of the YOLO detection as the observation vector $\mathbf{z}_t $, i.e.,  $\mathbf{z}_t = \left[x, y, w, h\right]^T$. With this in mind, the motion of the bounding box can modeled by a linear dynamics as
\begin{equation}
\left\{ \begin{array}{l}
{\bf x}_t = {\bf F}{\bf x}_{t-1} \\
{\bf z}_t = {\bf H}{\bf x}_t
\end{array} \right.
\label{eq11}
\end{equation}
where ${\bf{F}}$ is the state transition matrix and ${\bf{H}}$ denotes the obervation matrix.

The Kalman filter operates in two key stages: prediction and update. The prediction stage can be expressed mathematically as
\begin{equation}
\left\{ \begin{array}{l}
{{{\bf{\hat x}}}_{t|t - 1}} = {\bf{F}}{{{\bf{\hat x}}}_{t - 1|t - 1}} \\
{{\bf{P}}_{t|t - 1}} = {\bf{F}}{{\bf{P}}_{t - 1|t - 1}}{{\bf{F}}^T} + {\bf{Q}}
\end{array} \right.
\label{eq12}
\end{equation}
where ${{\bf{\hat x}}_{t|t - 1}}$ represents the predicted state at time $t$, ${{\bf{P}}_{t|t - 1}}$ is the covariance matrix of the predicted state, and ${\bf{Q}}$ is the process noise covariance matrix.

In the update phase, the detection results provided by the YOLO detector are used to correct the predicted state as
\begin{equation}
\left\{ \begin{array}{l}
{{\bf{y}}_t} = {{\bf{z}}_t} - {\bf{H}}{{{\bf{\hat x}}}_{t|t - 1}}\\
{{\bf{S}}_t} = {\bf{H}}{{\bf{P}}_{t|t - 1}}{{\bf{H}}^T} + {\bf{R}}\\
{{\bf{K}}_t} = {{\bf{P}}_{t|t- 1}}{{\bf{H}}^T}{\bf{S}}_t^{ - 1}\\
{{{\bf{\hat x}}}_{t|t}} = {{{\bf{\hat x}}}_{t|t - 1}} + {{\bf{K}}_t}{{\bf{y}}_t}\\
{{\bf{P}}_{t|t}} = ({\bf{I}} - {{\bf{K}}_t}{\bf{H}}){{\bf{P}}_{t|t - 1}}
\end{array} \right.
\label{eq13}
\end{equation}
where ${{\bf{y}}_t}$ is the residual vector, ${{\bf{S}}_t}$ is the covariance matrix of the residual, ${{\bf{K}}_t}$ is the Kalman gain, ${\bf{R}}$ is the measurement noise covariance matrix, and ${{\bf{\hat x}}_{t|t}}$ and ${{\bf{P}}_{t|t}}$ are the updated state and state covariance matrix, respectively.

During motion detection, we combine the Kalman filter's predicted output with the template matching results, verifying accuracy through the Intersection over Union (IOU) ratio. Since the target UAV occupies a minimal area in the image, the template matching is confirmed to be accurate if the IOU is positive. In contrast, zero or negative IOU indicates a potential mismatch and the detection output updates the target position based on the Kalman filter prediction.

\begin{algorithm}
\caption{GL-YOMO Algorithm}
\begin{algorithmic}[1]
\Require Previous two frames and current frame, i.e., $F_{t-2}, F_{t-1}, F_{t}$
\Statex $Y_{\text{det}}$: YOLO Detector
\Statex $M_{\text{det}}$: Motion Detector
\Ensure Detection results, i.e., $boxes, scores, class\_ids$
\State Initialize $N_g \gets 0$, $N_l \gets 0$, $\texttt{detector} \gets \text{`global'}$, $ROI \gets \text{None}$
\For{each frame $F_{t}$}
    \If{$\texttt{detector} = \text{`global'}$}
        \State $boxes, scores, class\_ids \gets Y_{\text{det}}(F_{t})$
    \Else
        \State $boxes, scores, class\_ids \gets Y_{\text{det}}(ROI(F_{t}))$
    \EndIf

    \If{$boxes = \emptyset$}
        \State $boxes, scores, class\_ids \gets M_{\text{det}}(F_{t-2}, F_{t-1}, F_{t})$ \Comment{Switch to motion detection}
    \EndIf
    
    \If{$boxes \neq \emptyset$}
        \State Update $ROI$ based on $boxes$
        \If{$\texttt{detector}= \text{`global'}$}
            \State $N_g \gets N_g + 1$, $N_l \gets 0$
            \If{$N_g \geq 30$}
                \State $\texttt{detector} \gets \text{`local'}$
            \EndIf
        \Else
            \State $N_l \gets 0$
        \EndIf
    \Else
        \If{$\texttt{detector} = \text{`local'}$}
            \State $N_l \gets N_l + 1$
            \State Enlarge $ROI$
            \If{$N_l \geq 60$}
                \State $\texttt{detector}\gets \text{`global'}$
                \State $N_g \gets 0$
                \State Reset $ROI$ to full frame
            \EndIf
        \EndIf
    \EndIf

    \State Update frame history: $F_{t-2} \gets F_{t-1}$, $F_{t-1} \gets F_{t}$
\EndFor
\end{algorithmic}
\end{algorithm}

\subsection{Summary}
Algorithm 1 provides a detailed description of the core logic and procedural flow of the proposed GL-YOMO method. The detection outputs are characterized by bounding box $boxes$, detection scores $scores$ and target ID $class\_ids$.

\section{Experiments}
\subsection{Dataset}

In evaluating the performance of our proposed GL-YOMO algorithm, we selected two challenging video datasets to ensure a comprehensive and accurate assessment.

\subsubsection{Drone-vs-Bird Dataset}

\begin{figure}[htbp]
	\centering
	\includegraphics[width=\linewidth]{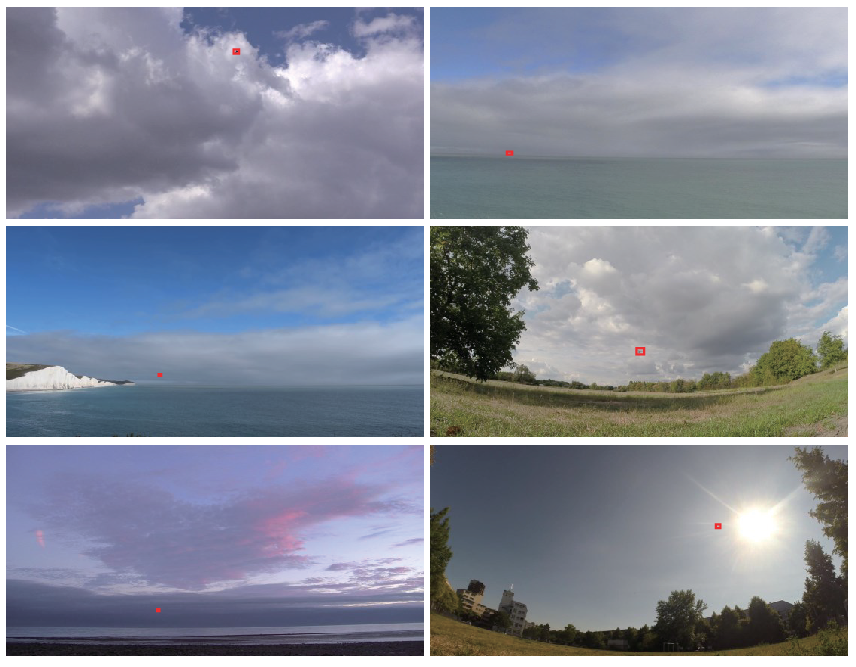}	
	\caption{Sample Images from the Drone-vs-Bird dataset. The red boxes highlight actual drone targets.}	
	\label{Fig6}	
\end{figure}

The Drone-vs-Bird dataset consists of 77 videos, encompassing a total of 104,760 frames. Many of the videos feature small drones filmed from considerable distances, frequently accompanied by birds and insects, adding an extra layer of complexity to the surveillance tasks. Captured with both static and dynamic cameras, these videos effectively simulate a wide range of outdoor scenarios. The average object size is 34×23 pixels, constituting 0.1\% of the image, as shown in Fig. \ref{Fig6}. To ensure reliable evaluation results, we used 60 videos for training and validation, while the remaining 15 were reserved for testing.

\subsubsection{Fixed-Wings Dataset}
\begin{figure}[htbp]
	\centering
	\includegraphics[width=\linewidth]{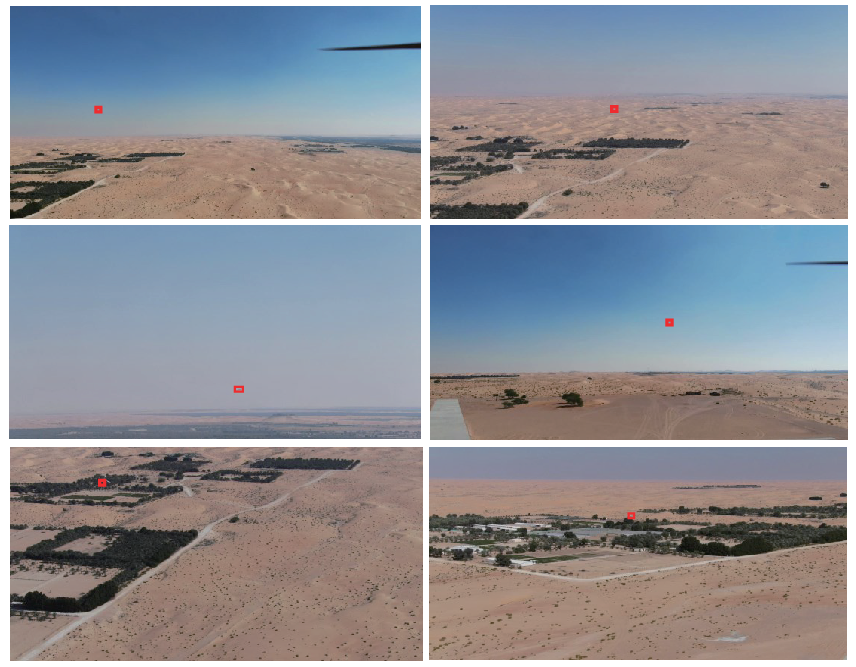}	
	\caption{Sample Images from the Fixed-Wings dataset. The red boxes highlight actual drone targets.}	
	\label{Fig7}	
\end{figure}

\begin{figure}[htbp]
	\centering
	\includegraphics[width=\linewidth]{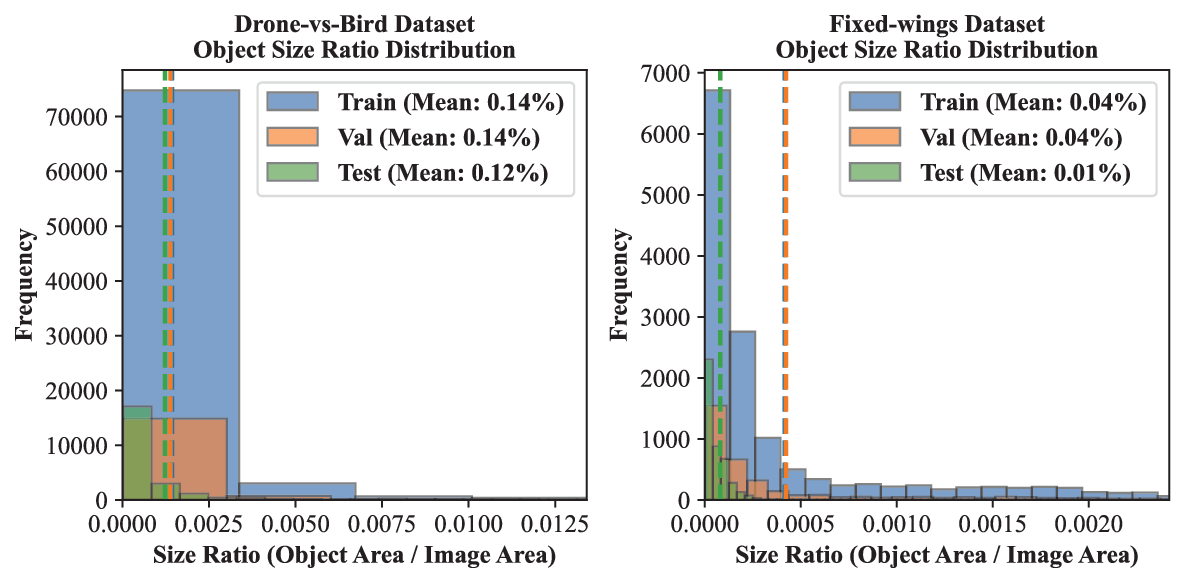}	
	\caption{Object size ratio distribution statistics of Drone-vs-Bird dataset and Fixed-wings dataset.}	
	\label{Fig8}	
\end{figure}

We have developed a challenging video dataset specifically for fixed-wing UAV targets. Comprising 13 video sequences with a total of 24,951 frames, all recorded at 30 FPS and 1920×1080 resolution, the dataset features numerous tiny targets against complex backgrounds, as shown in Fig. \ref{Fig7}. Many targets closely resemble background features, making them difficult to distinguish visually. To validate our algorithm, we used a test set comprising 4,673 frames with tiny targets ranging from 1×1 to 146×95 pixels, as illustrated in Fig. \ref{Fig8}, the average size of the test set occupies only 0.01\% of the image area. The remaining 12 sequences were randomly split at 8:2 for training and validation. To our knowledge, this is one of the smallest fixed-wings UAV target datasets available.

\subsection{Evaluation Metrics and Implementation Details}

In this study, we employed a standard evaluation metric system to quantify the performance of the detection algorithm, including Precision, Recall, AP, and two variants of mean average precision: mAP50 and mAP50-95. The experiments were conducted on a high-performance computing system equipped with two NVIDIA GeForce RTX 3090 GPUs. The image size was adjusted to $640\times640$ during the training phases of the model, with SGD as the optimizer, a momentum of 0.937, an initial learning rate of 0.01, 200 epochs, a batch size of 64, and an IOU threshold of 0.1 for evaluation.  Table \ref{tab1} outlines the threshold parameters in this paper, which have been validated to ensure precision and reliability.

\begin{table}[ht]
\centering
\caption{Threshold parameter configuration}
\begin{tabular}{c|c||c|c||c|c}
\hline
\textbf{Notation} & \textbf{Value} & \textbf{Notation} & \textbf{Value} & \textbf{Notation} & \textbf{Value} \\
\hline
$N_g$ & 30 & $\tau_g$ & 0.3 & $k_1$  & 1 \\

$N_l$  & 60 & $\tau_l$ & 0.1 & $k_2$ & 0.6 \\

$R_s$  & 4/5 & $\Delta p$ & 50 & $k_3$ & 0.4 \\
\hline
\end{tabular}
\label{tab1}
\end{table}

\subsection{Comparison with Existing Works}	

\begin{figure*}[htbp]
    \centering
    \setlength{\tabcolsep}{2pt}  
    \renewcommand{\arraystretch}{1.2}  
    \begin{tabular}{>{\centering\arraybackslash}m{0.1\textwidth}
                    >{\centering\arraybackslash}m{0.22\textwidth}
                    >{\centering\arraybackslash}m{0.22\textwidth}
                    >{\centering\arraybackslash}m{0.22\textwidth}
                    >{\centering\arraybackslash}m{0.22\textwidth}}
        
        YOLOv5s & 
        \includegraphics[width=\linewidth]{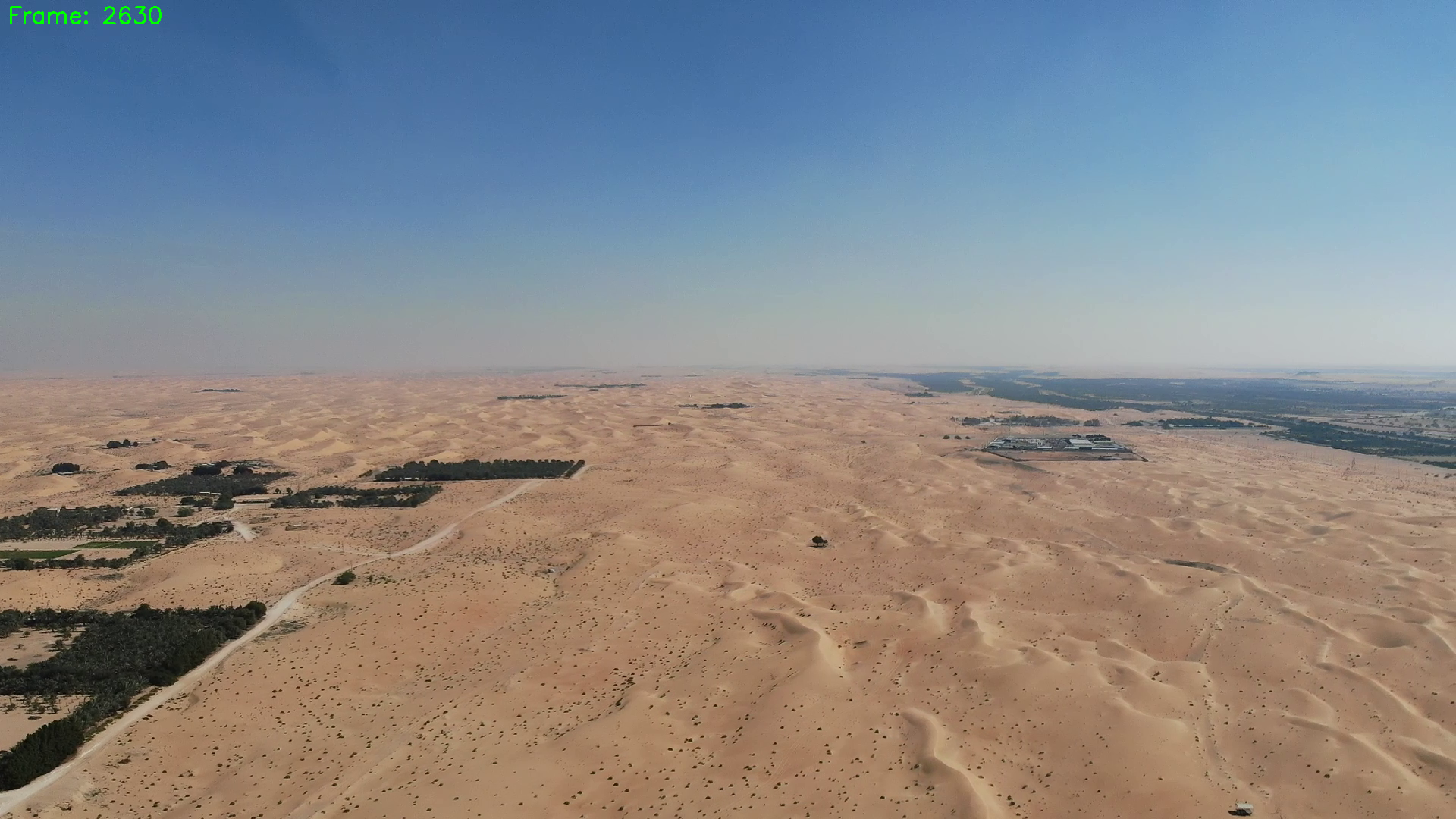} &
        \includegraphics[width=\linewidth]{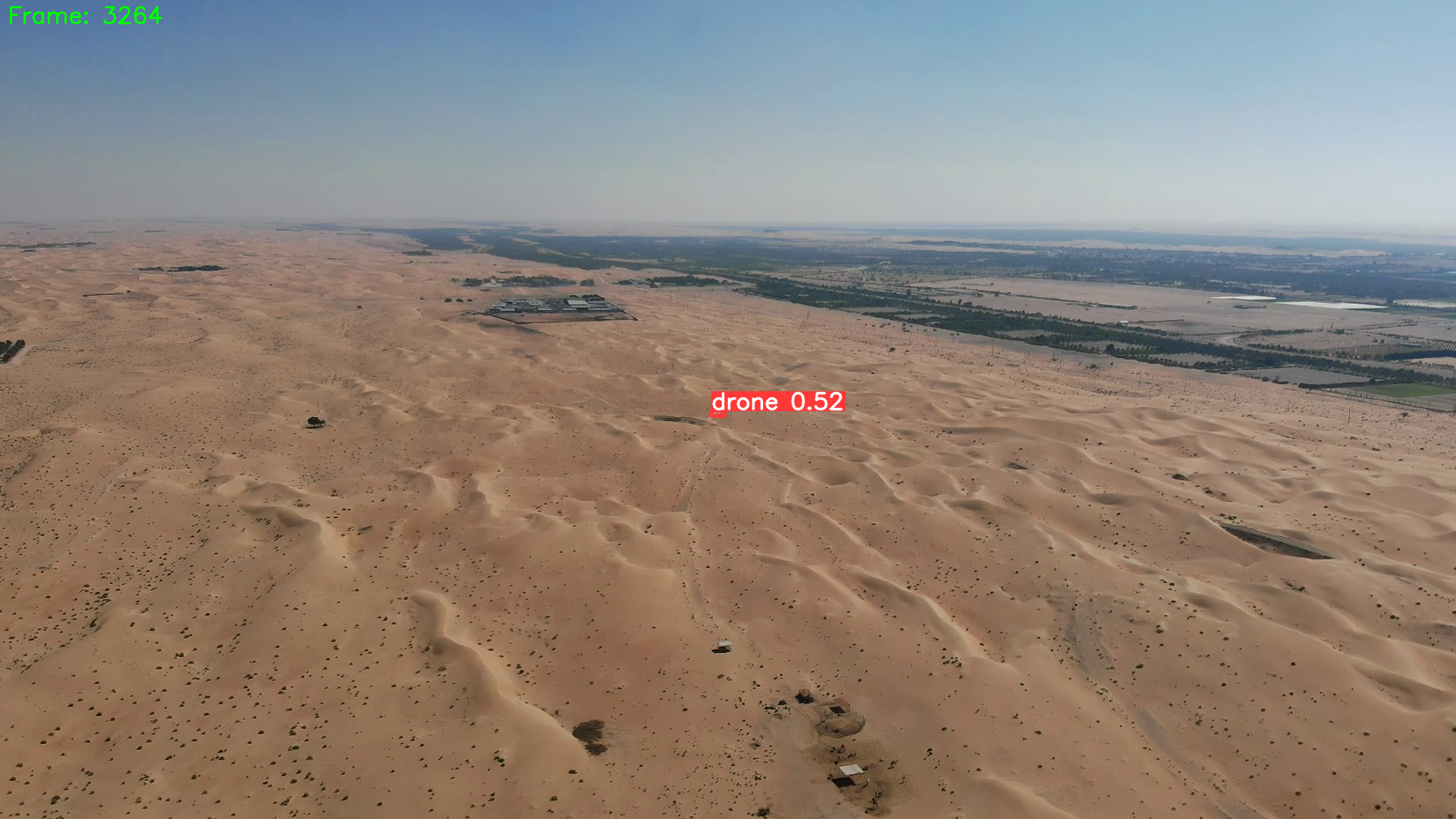} &
        \includegraphics[width=\linewidth]{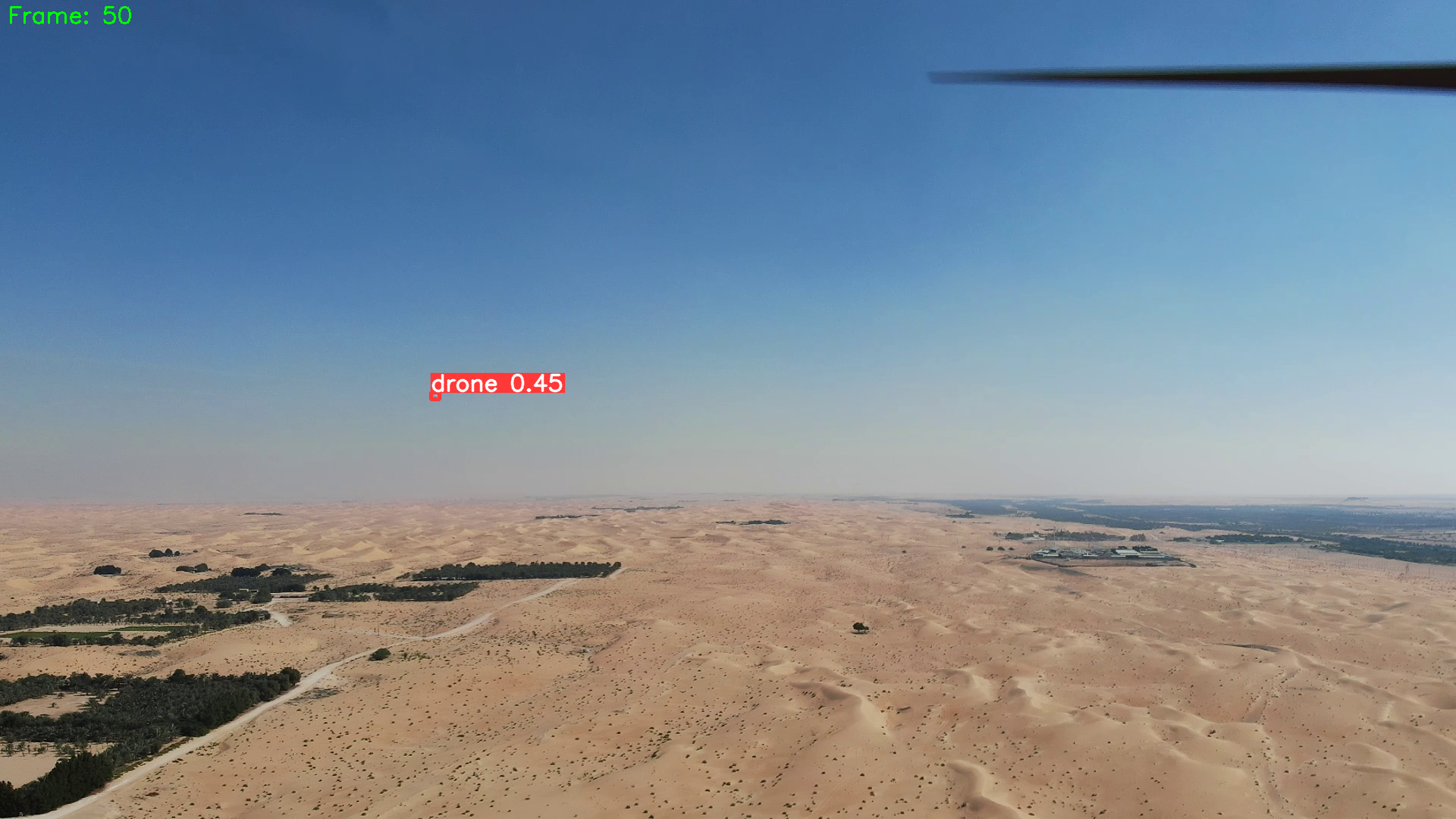} &
        \includegraphics[width=\linewidth]{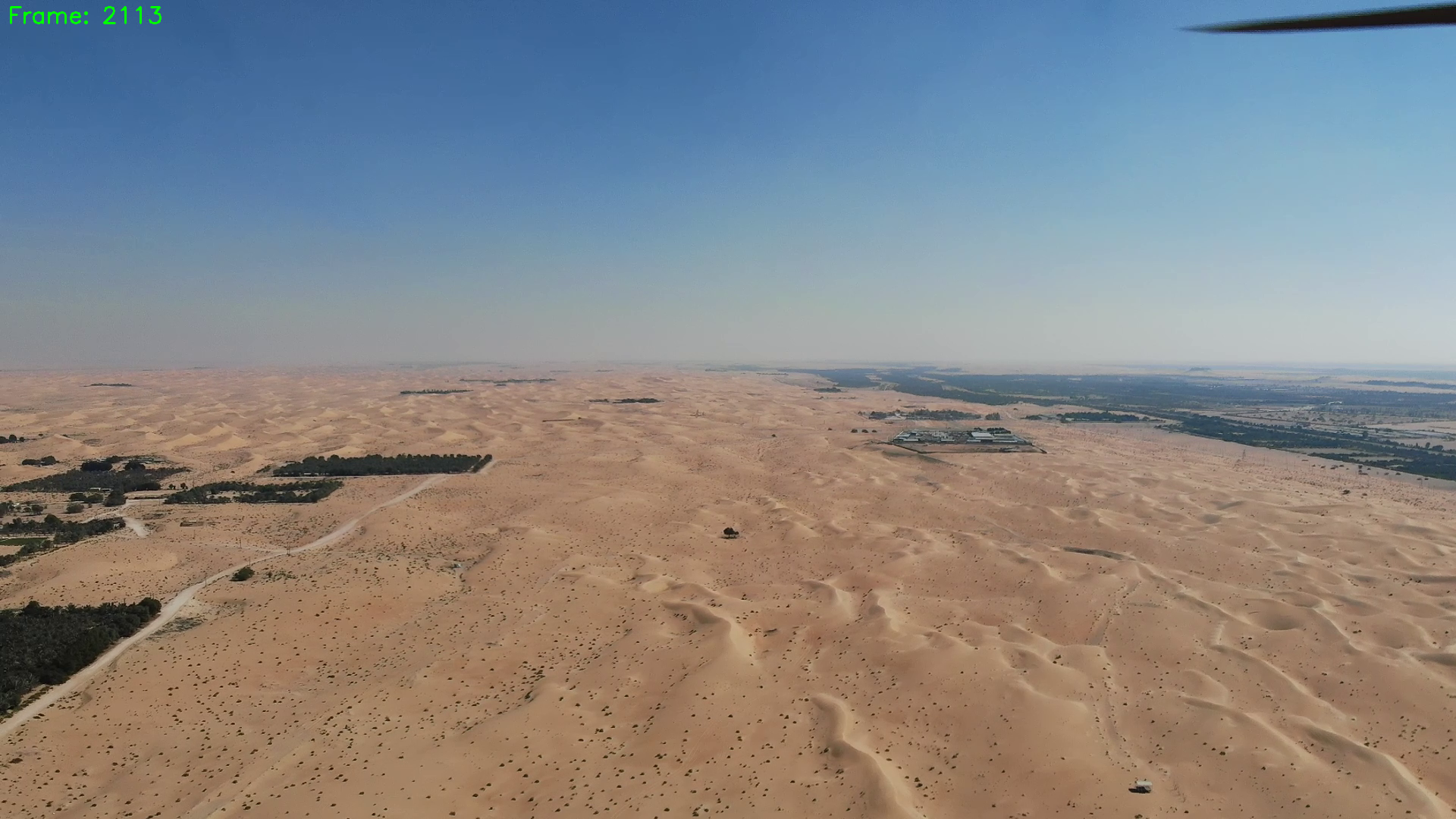} \\
        YOLOv8s & 
        \includegraphics[width=\linewidth]{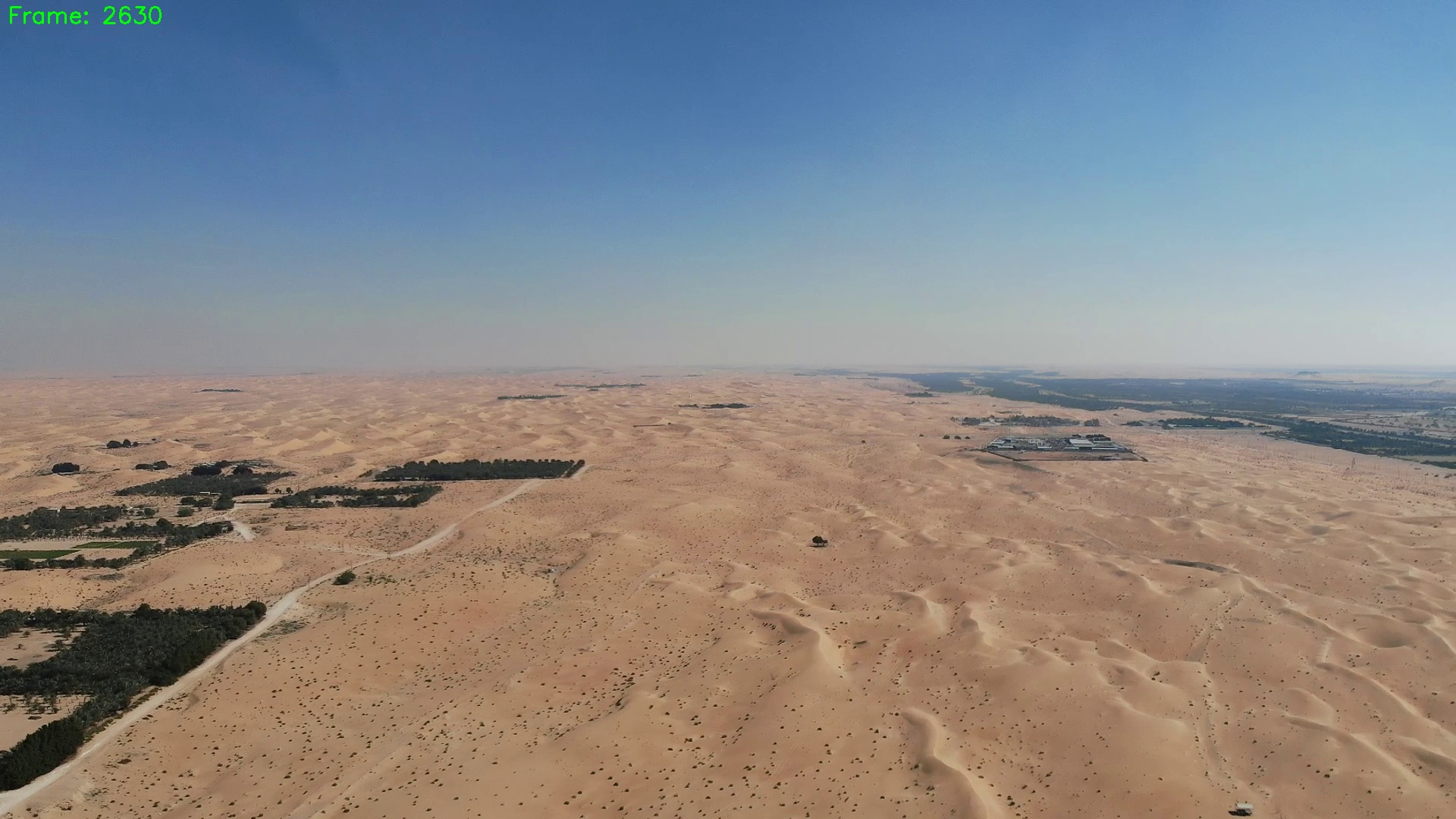} &
        \includegraphics[width=\linewidth]{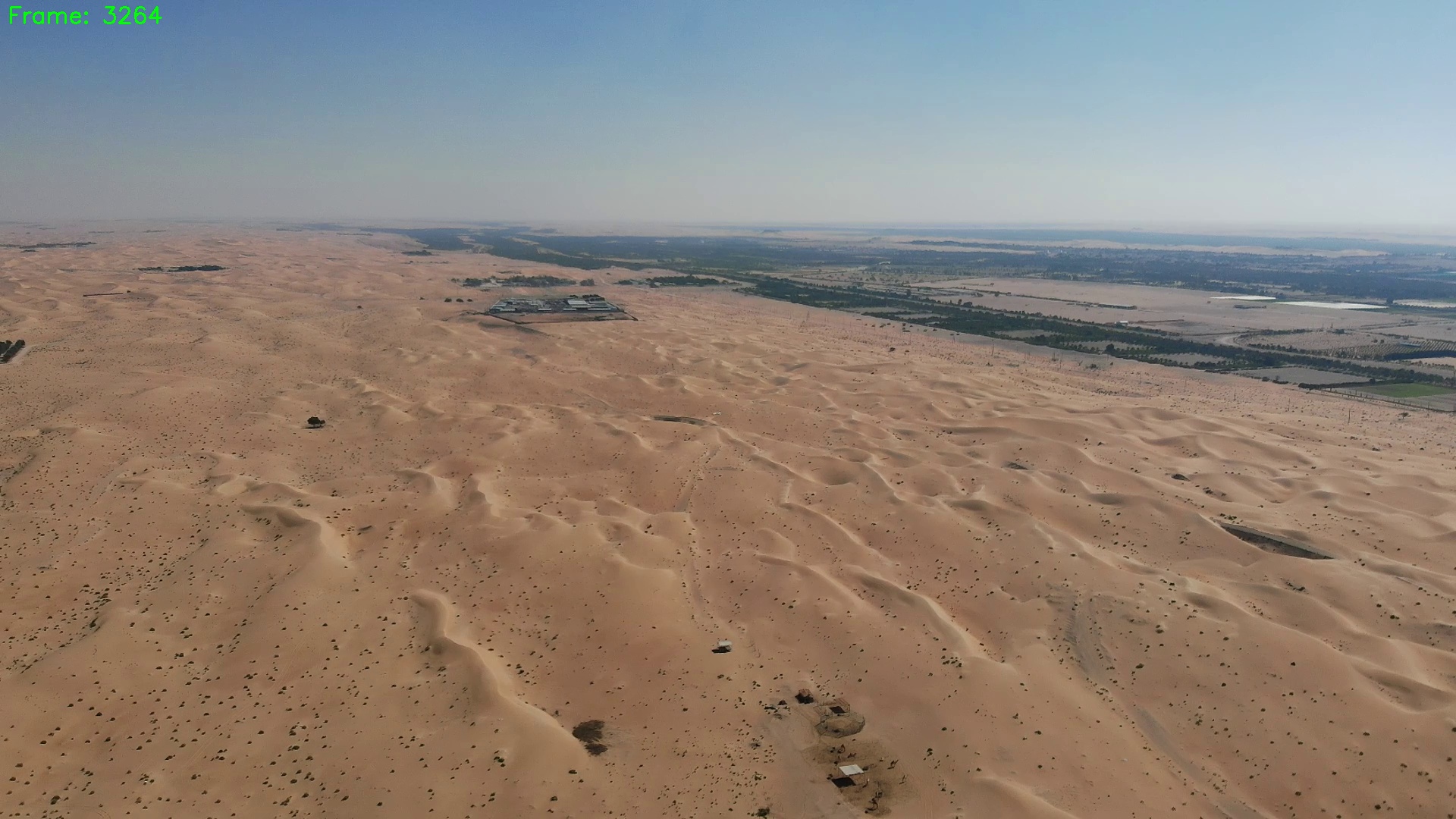} &
        \includegraphics[width=\linewidth]{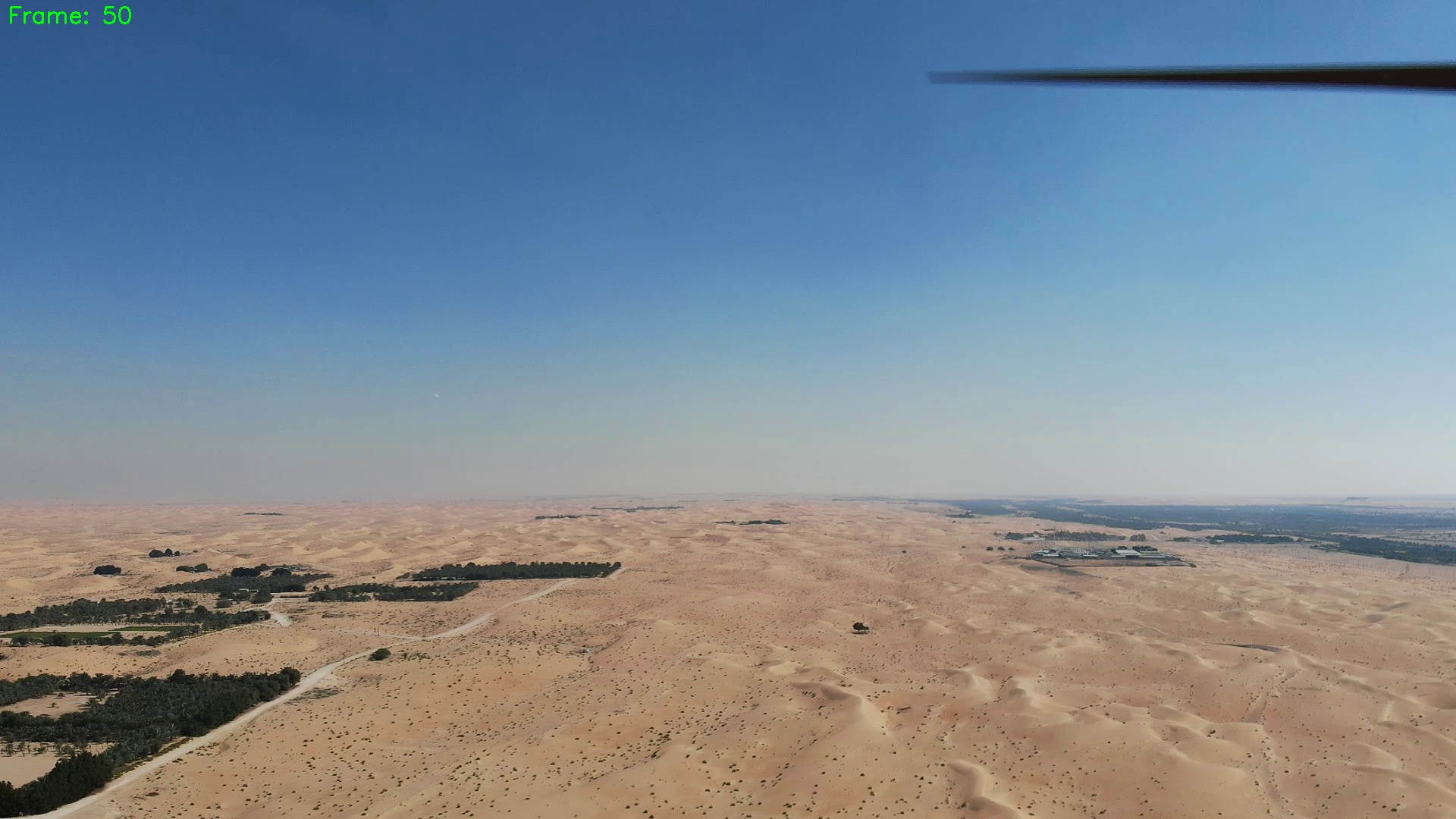} &
        \includegraphics[width=\linewidth]{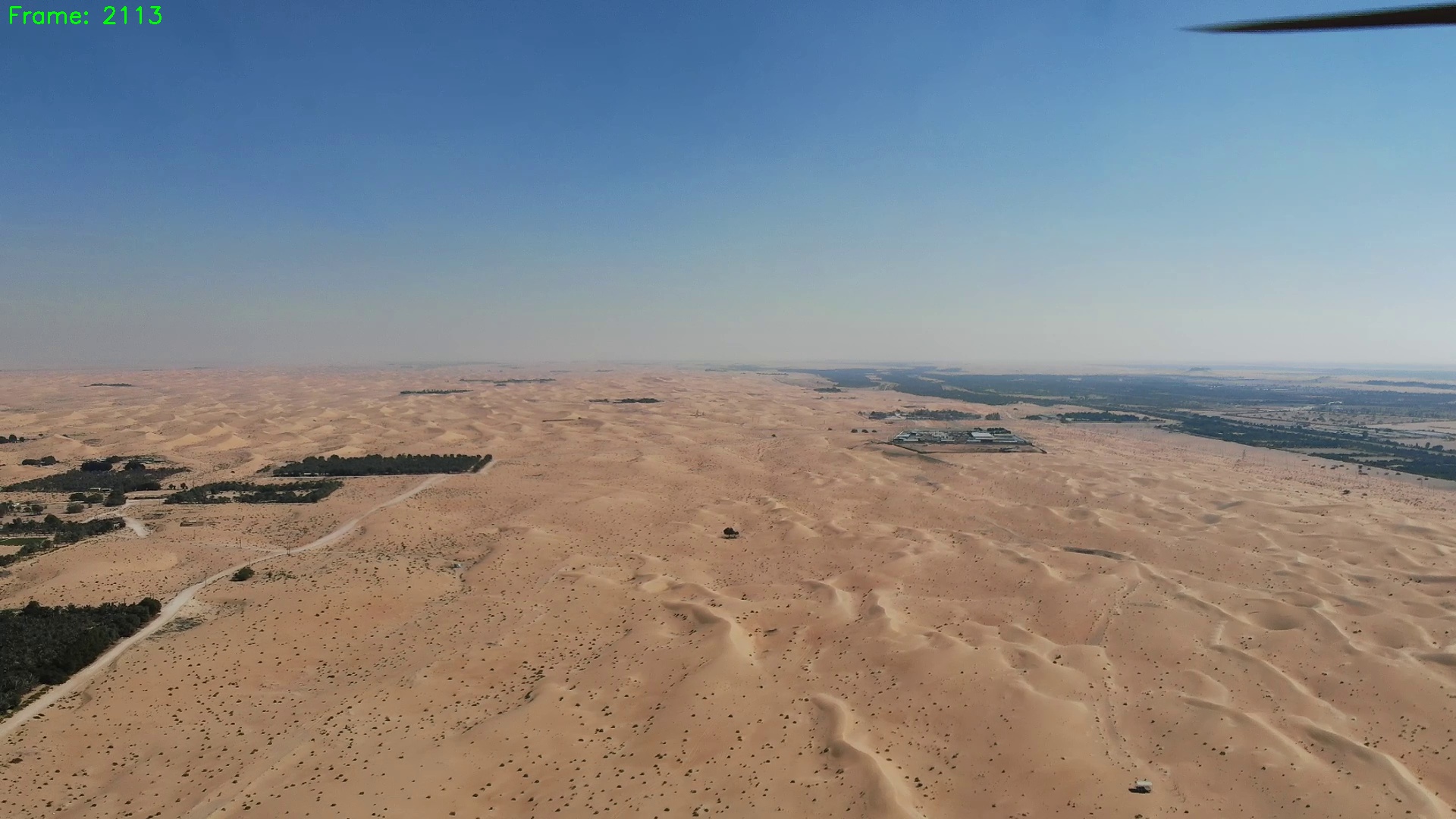} \\
        YOLOv10s & 
        \includegraphics[width=\linewidth]{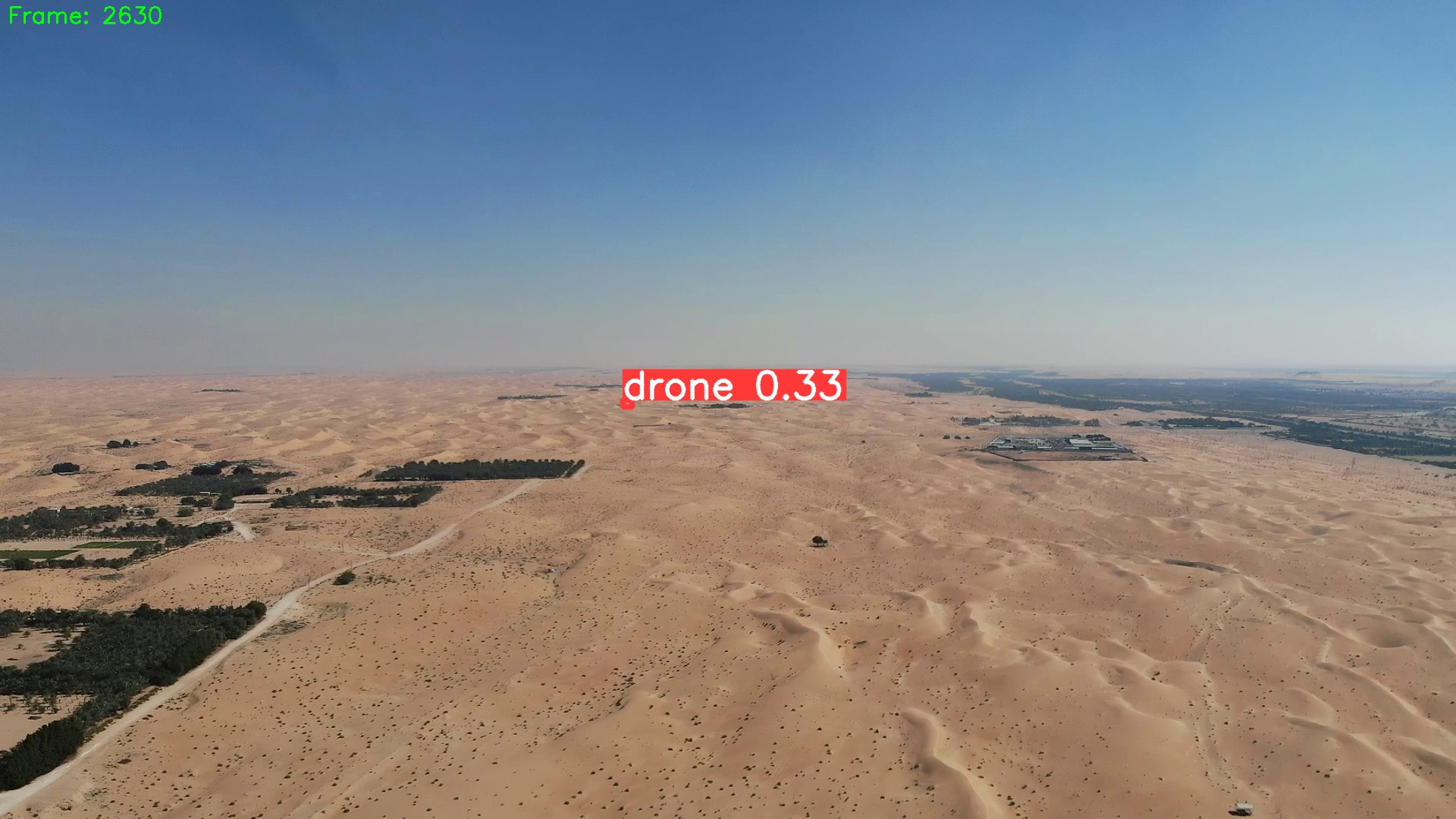} &
        \includegraphics[width=\linewidth]{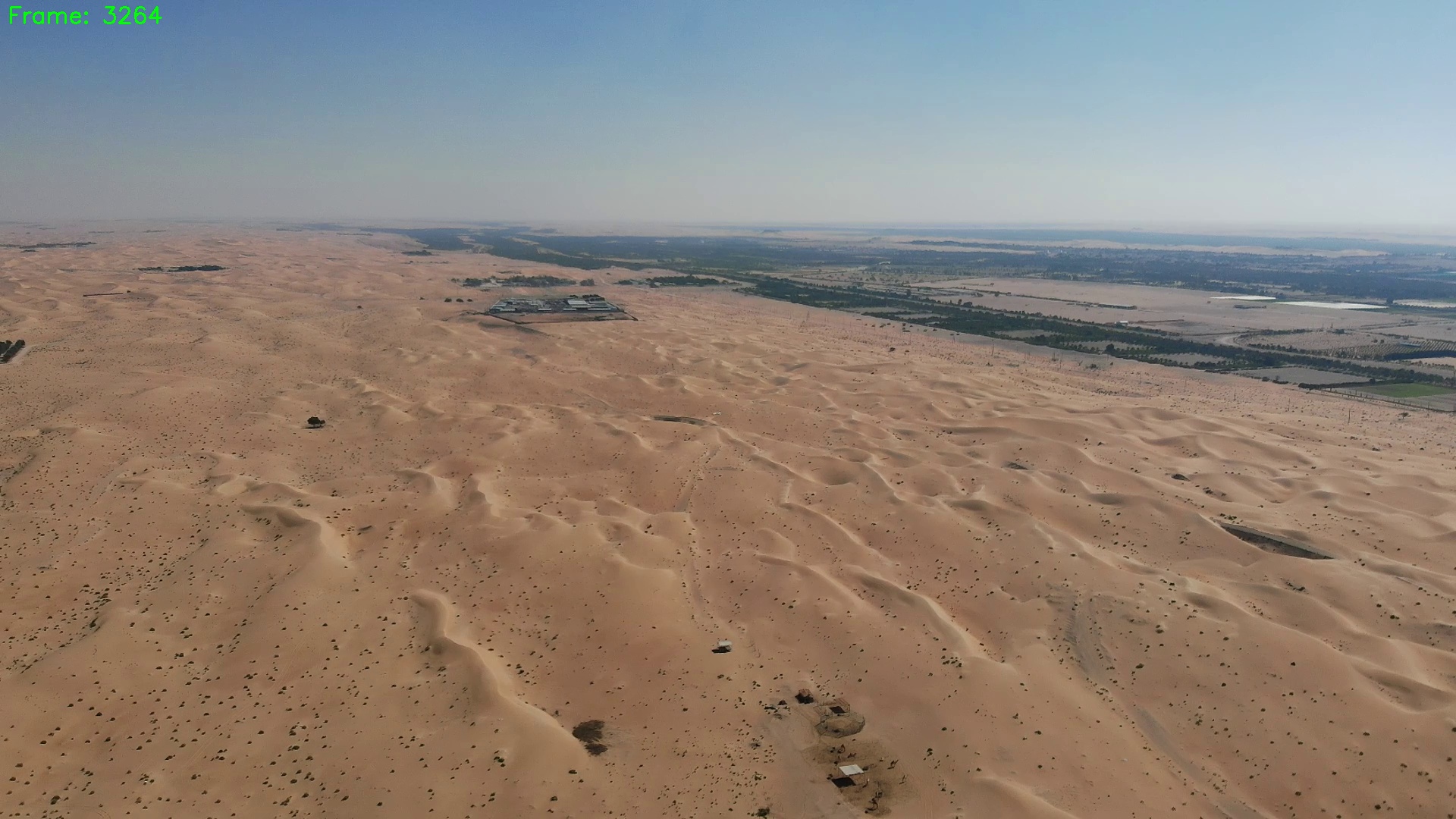} &
        \includegraphics[width=\linewidth]{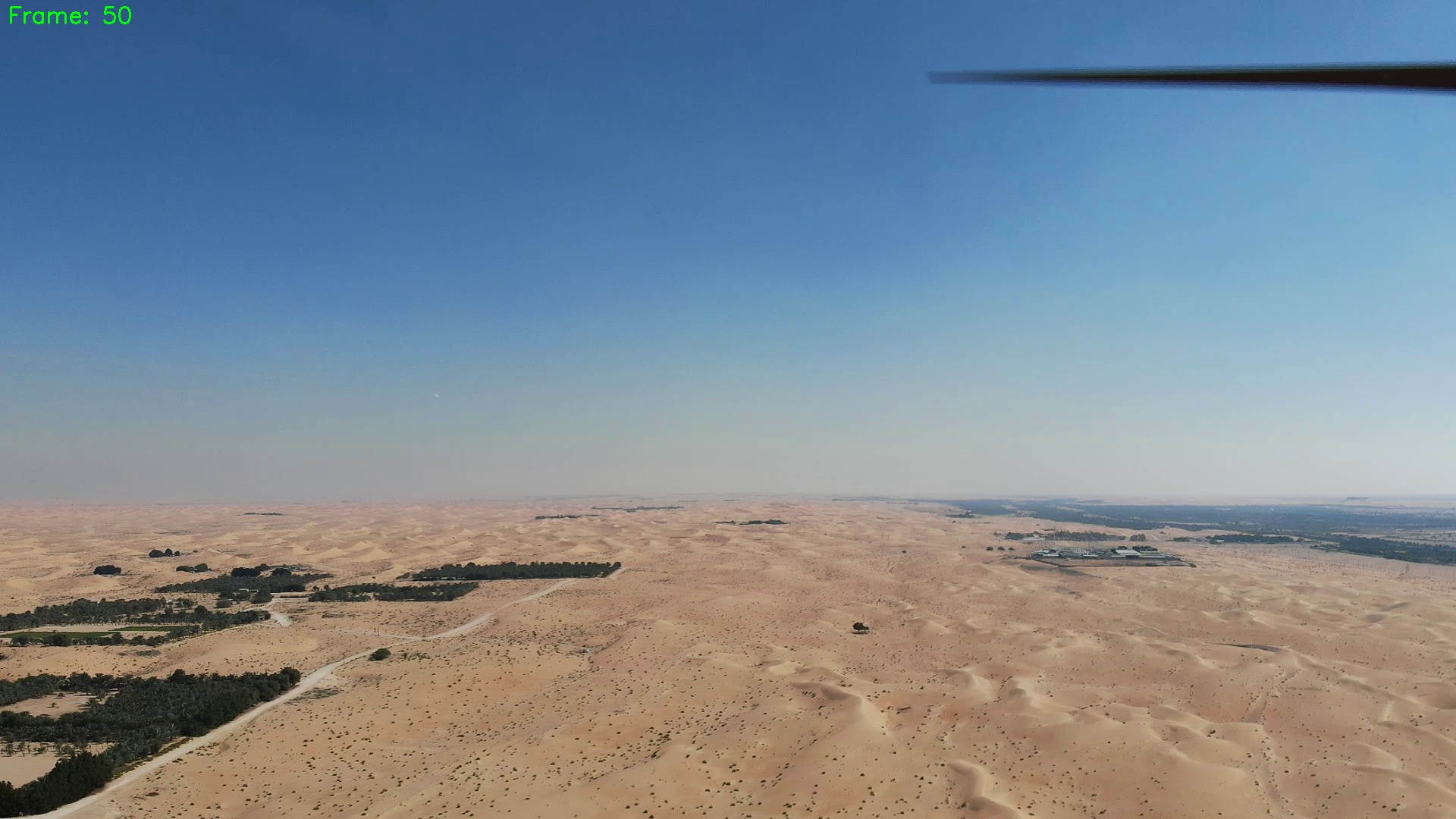} &
        \includegraphics[width=\linewidth]{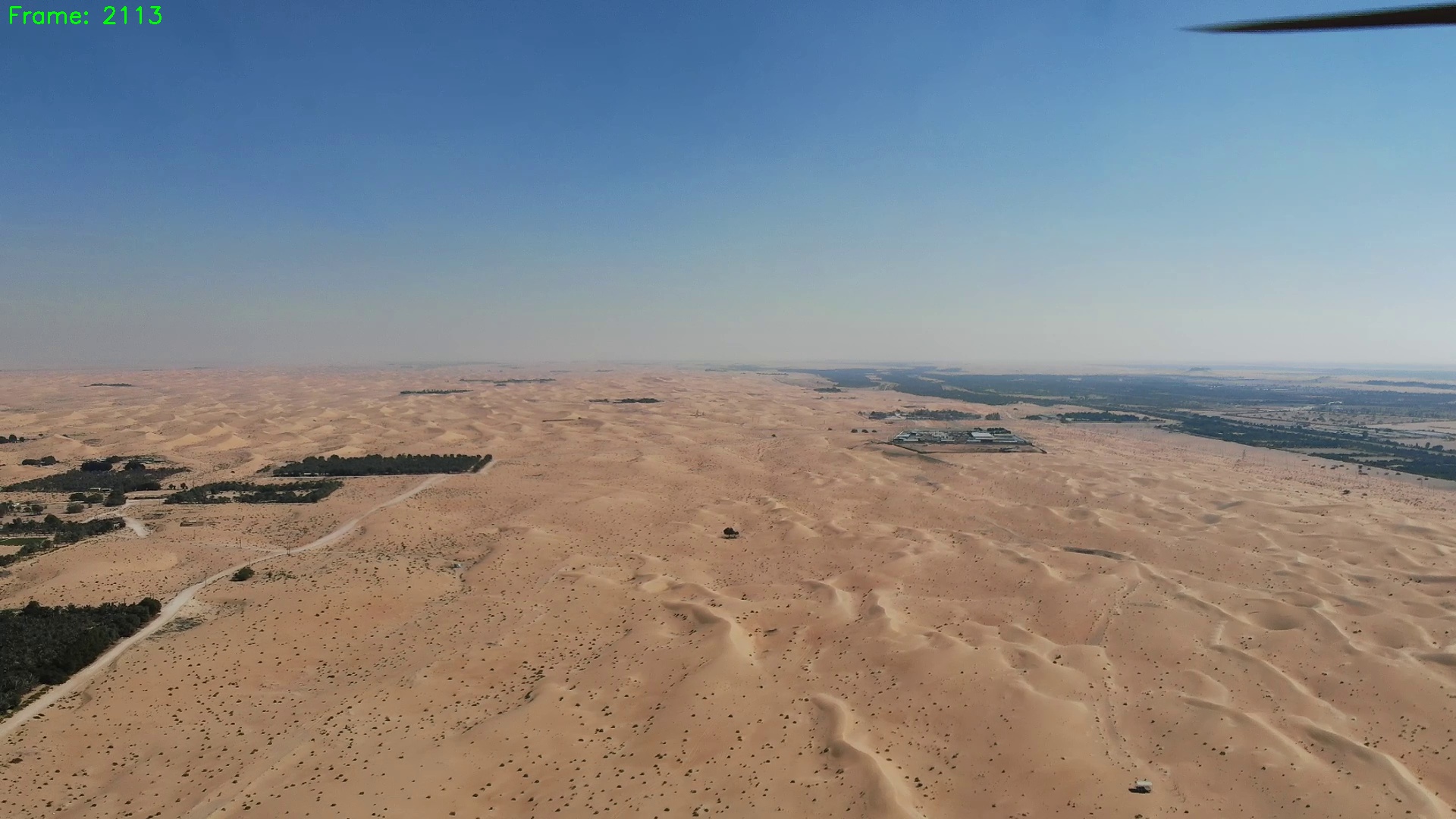} \\
        THP-YOLOv5 & 
        \includegraphics[width=\linewidth]{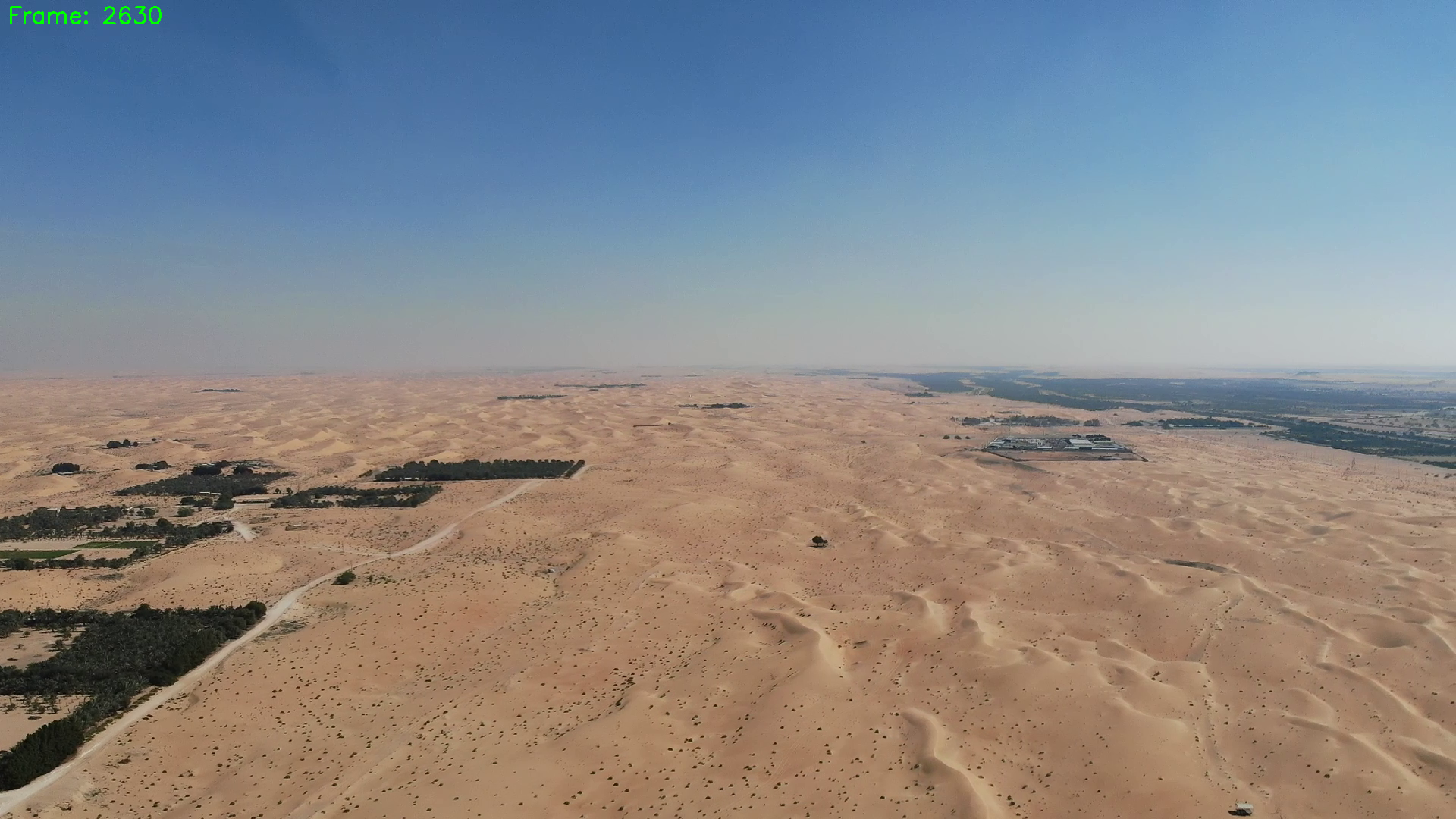} &
        \includegraphics[width=\linewidth]{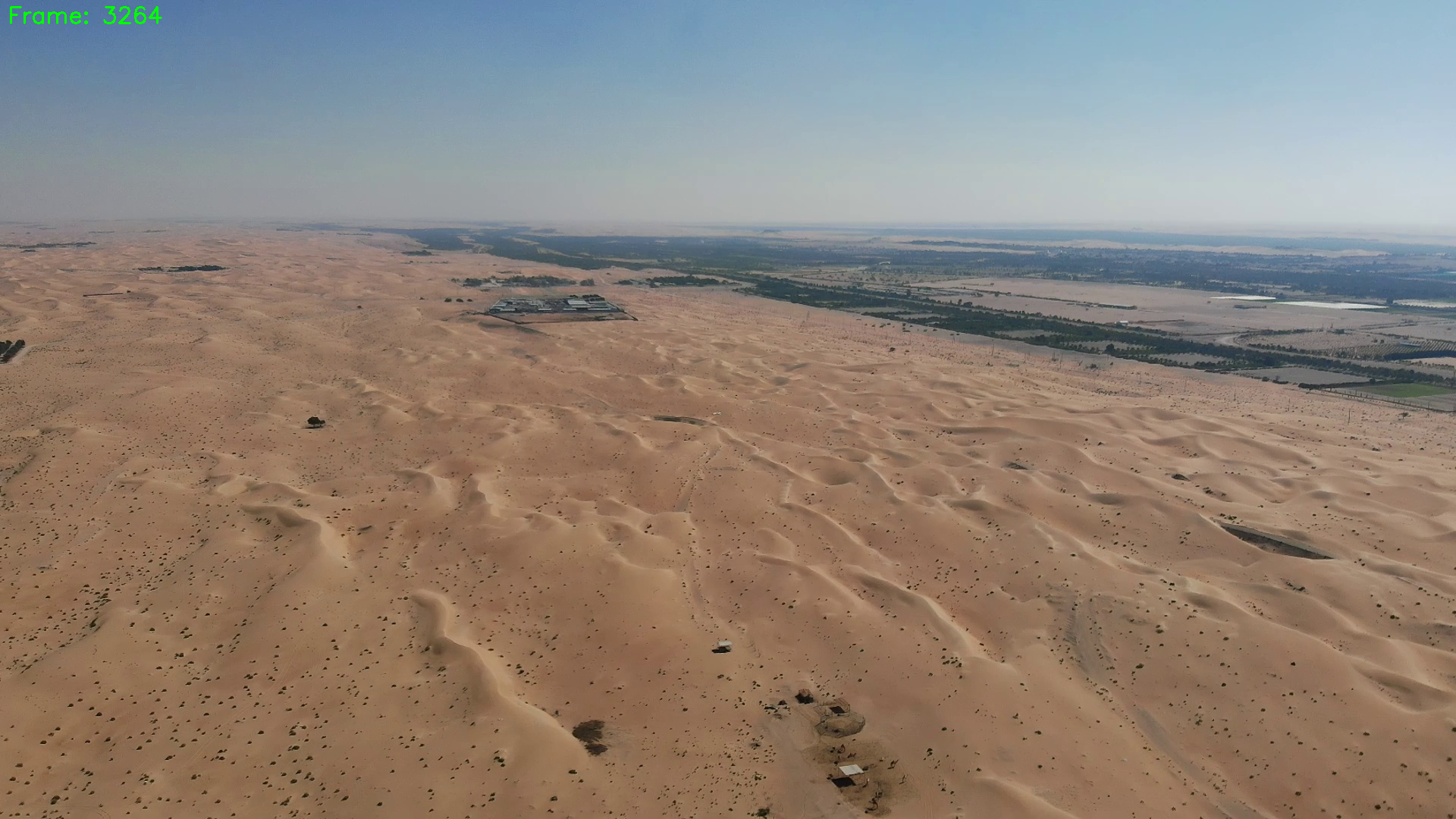} &
        \includegraphics[width=\linewidth]{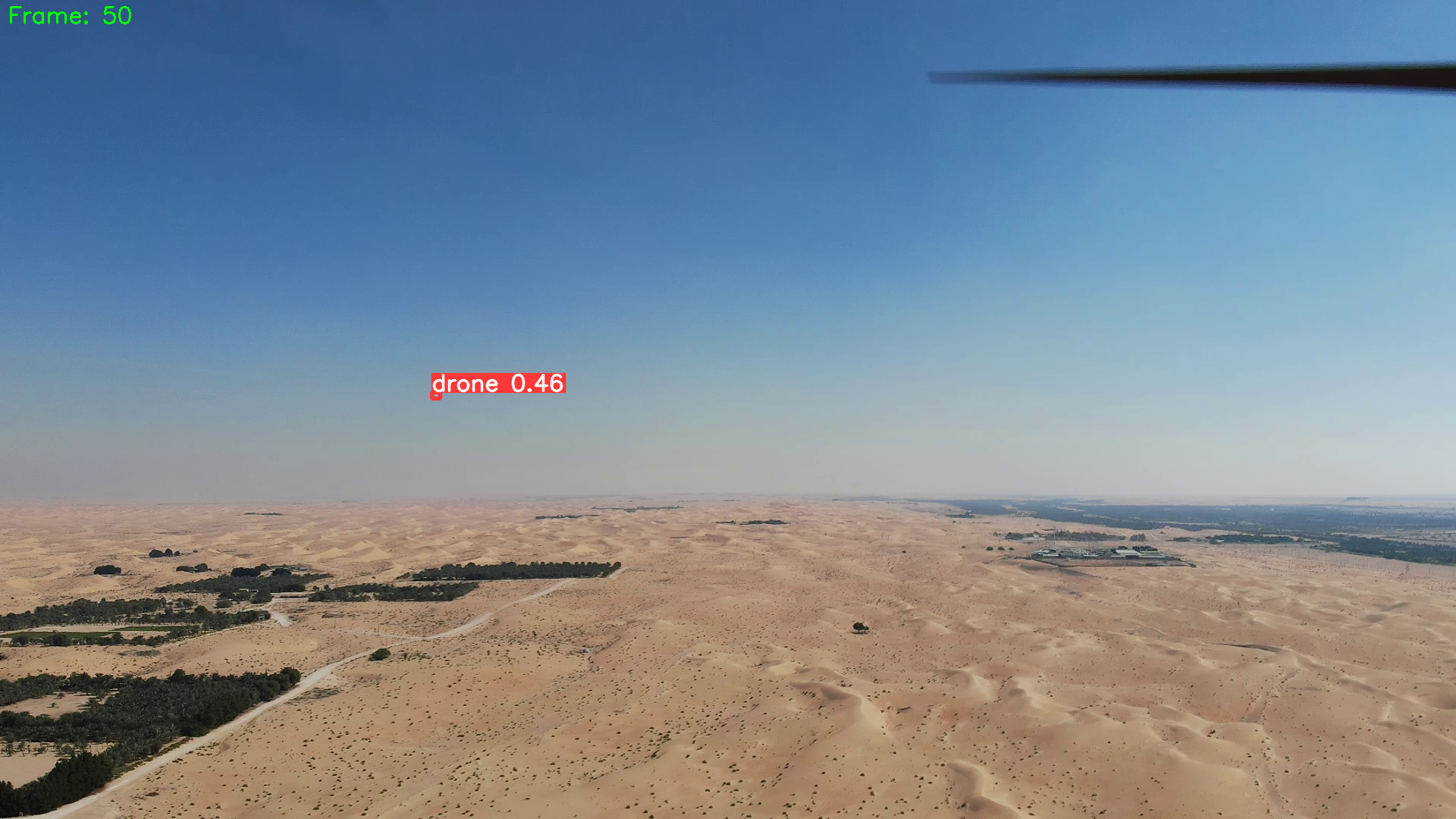} &
        \includegraphics[width=\linewidth]{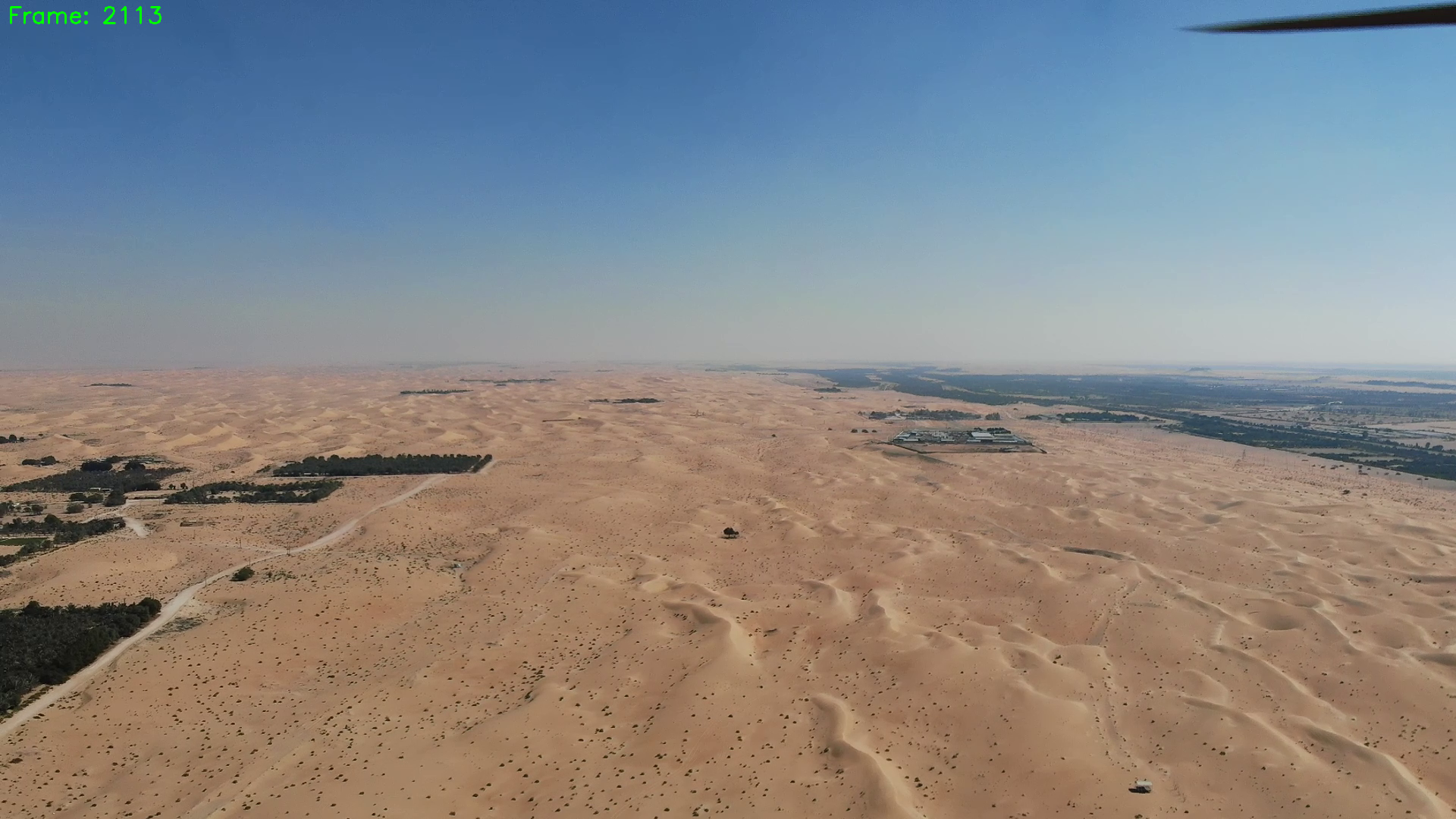} \\
        RT-DETR & 
        \includegraphics[width=\linewidth]{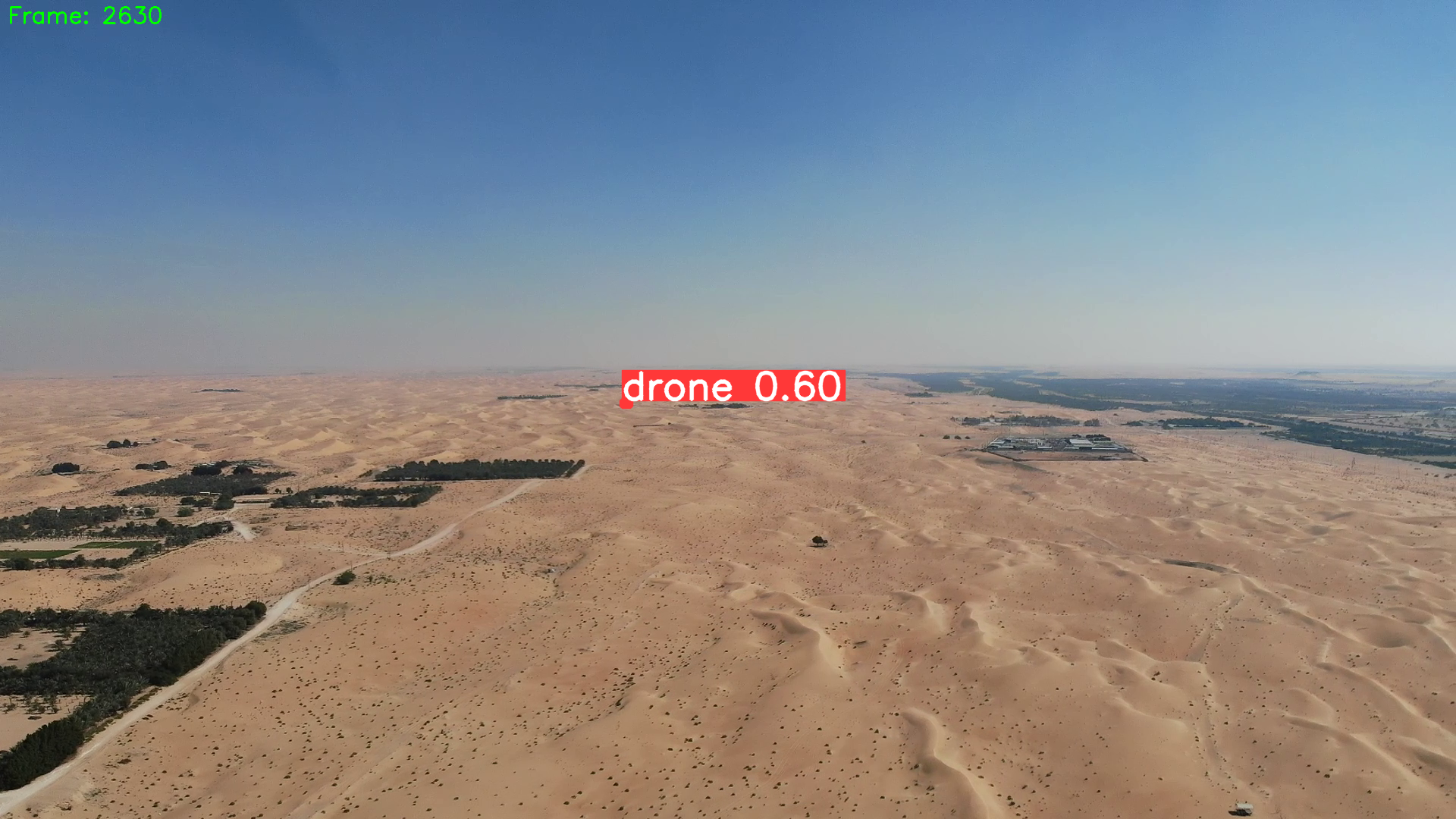} &
        \includegraphics[width=\linewidth]{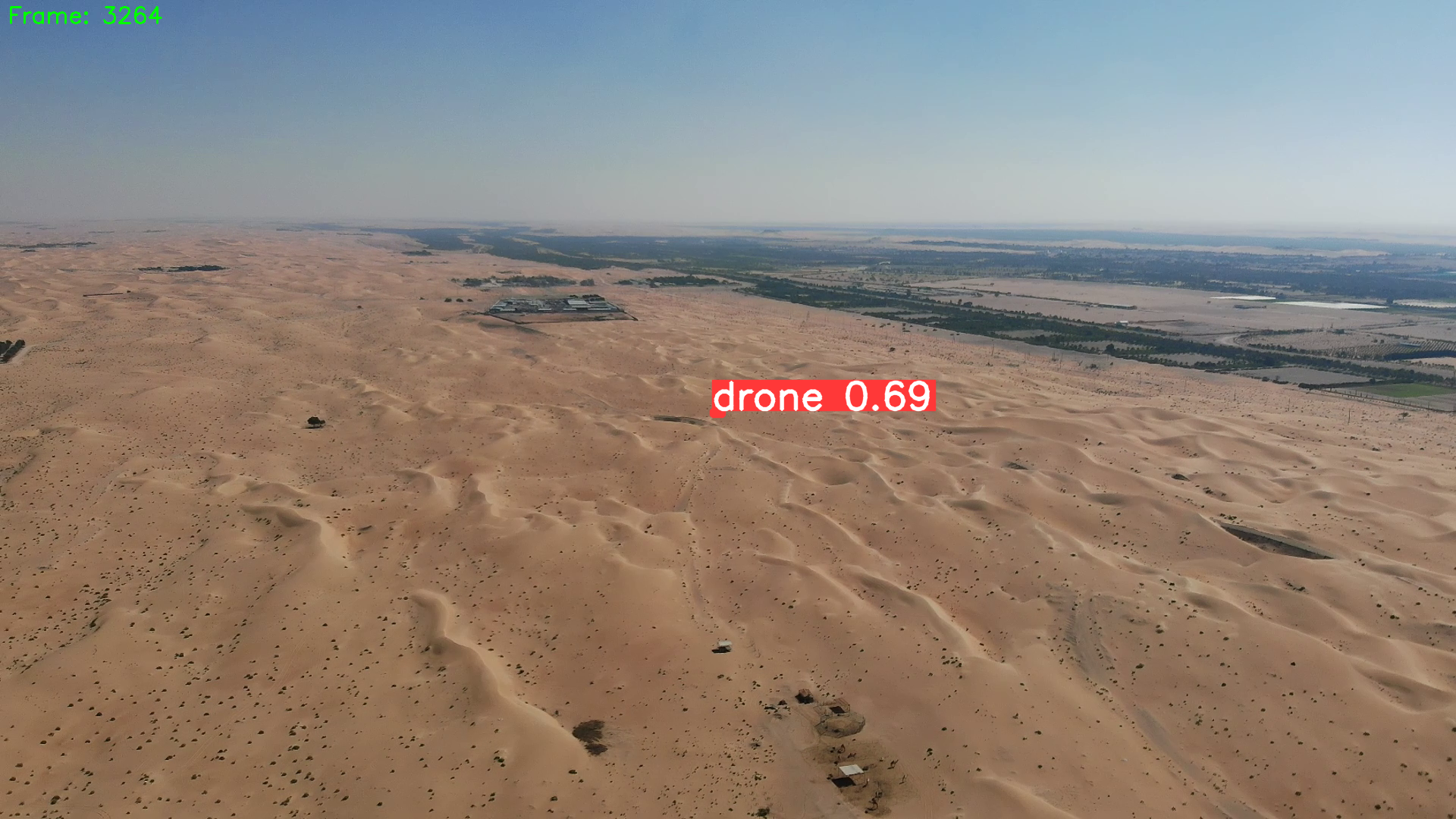} &
        \includegraphics[width=\linewidth]{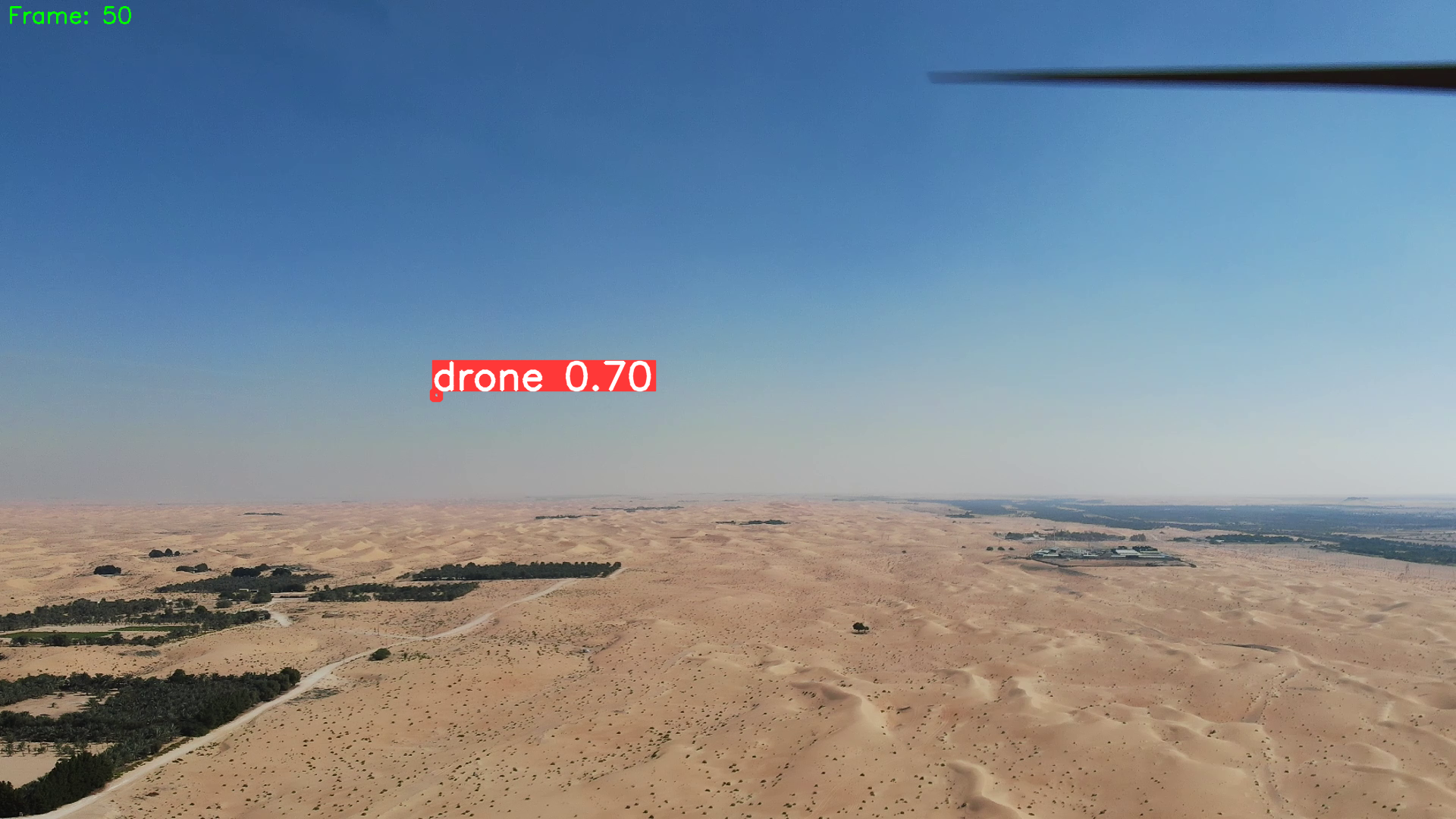} &
        \includegraphics[width=\linewidth]{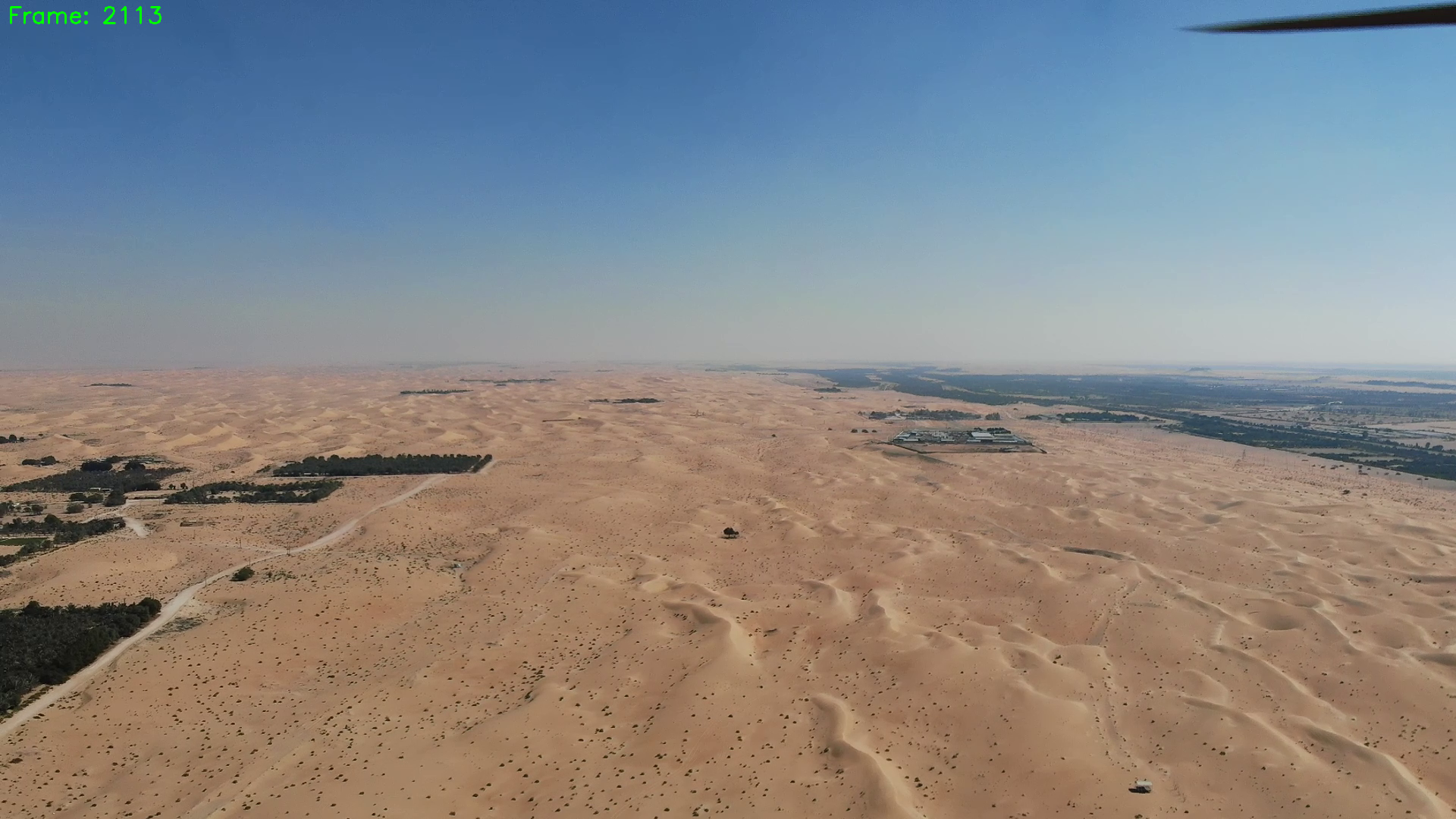} \\
        GL-YOMO & 
        \includegraphics[width=\linewidth]{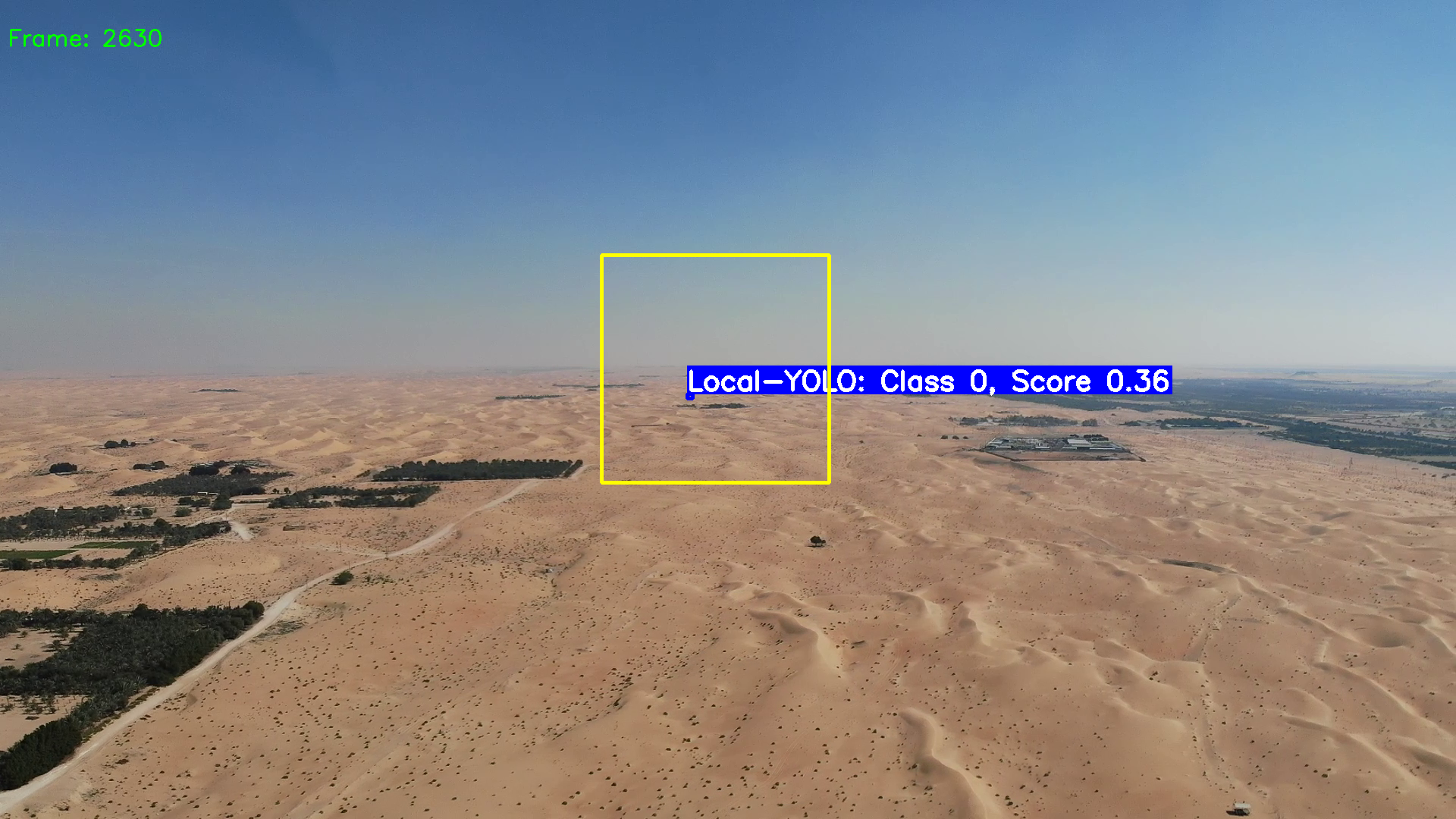} &
        \includegraphics[width=\linewidth]{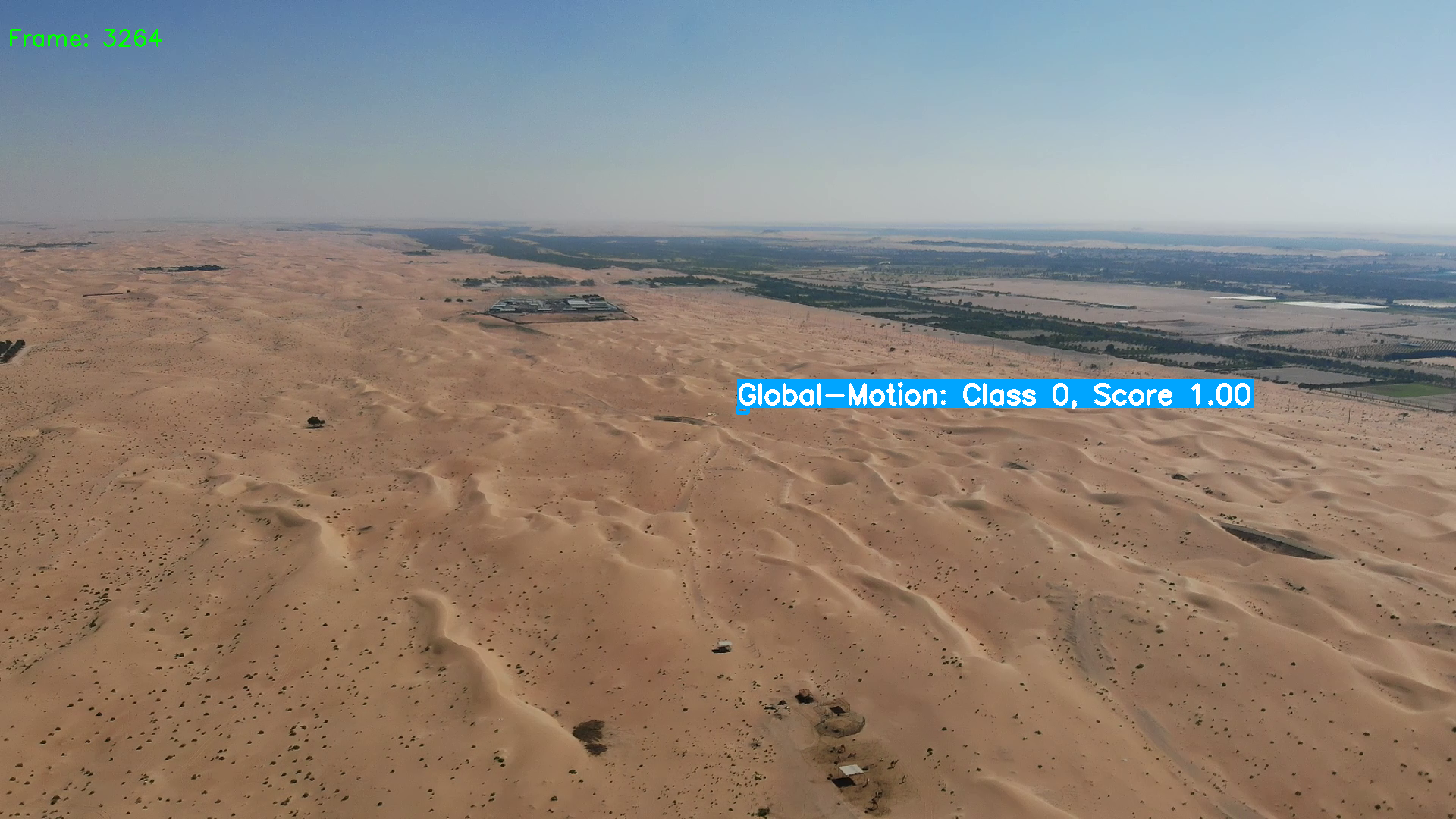} &
        \includegraphics[width=\linewidth]{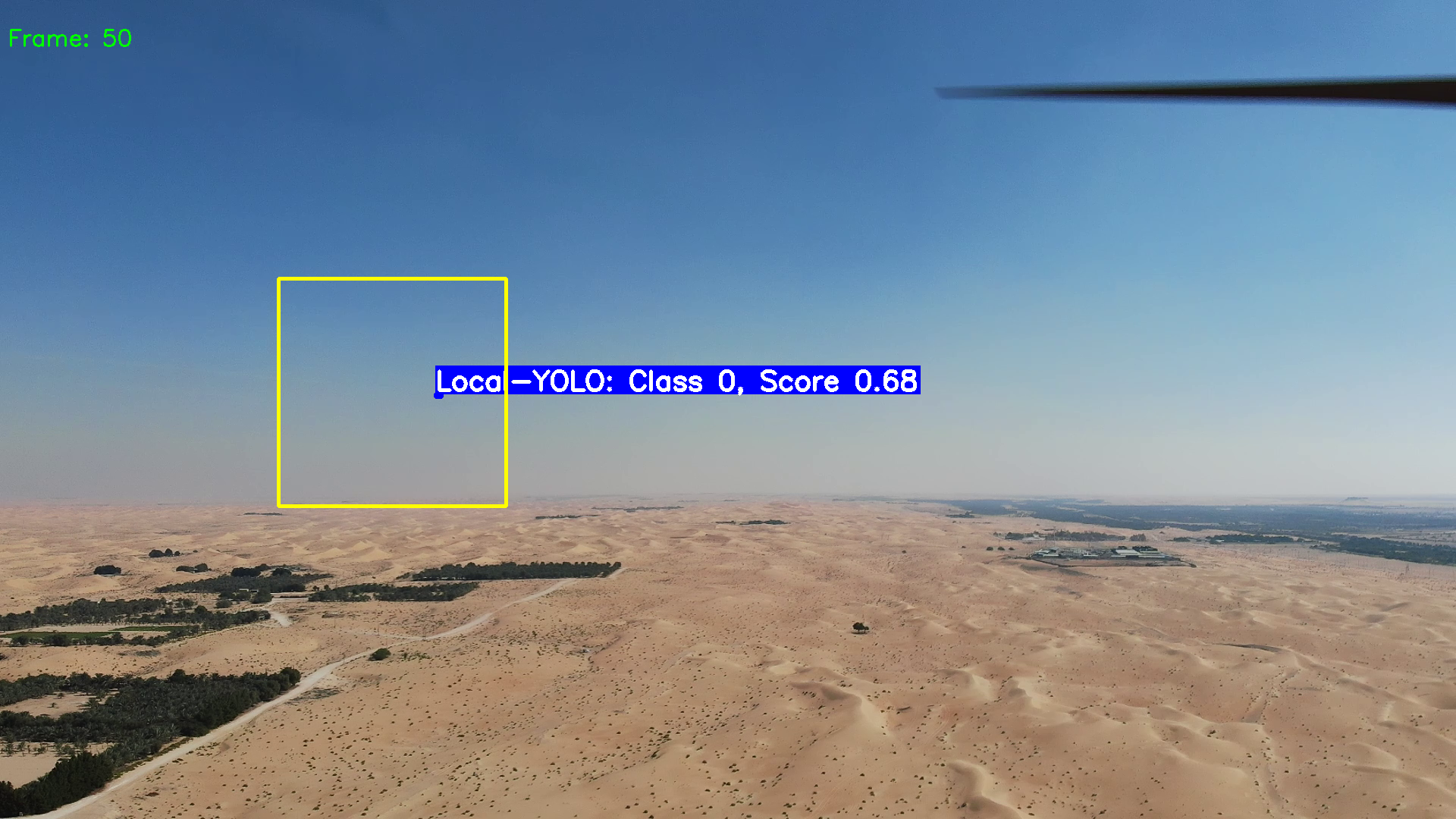} &
        \includegraphics[width=\linewidth]{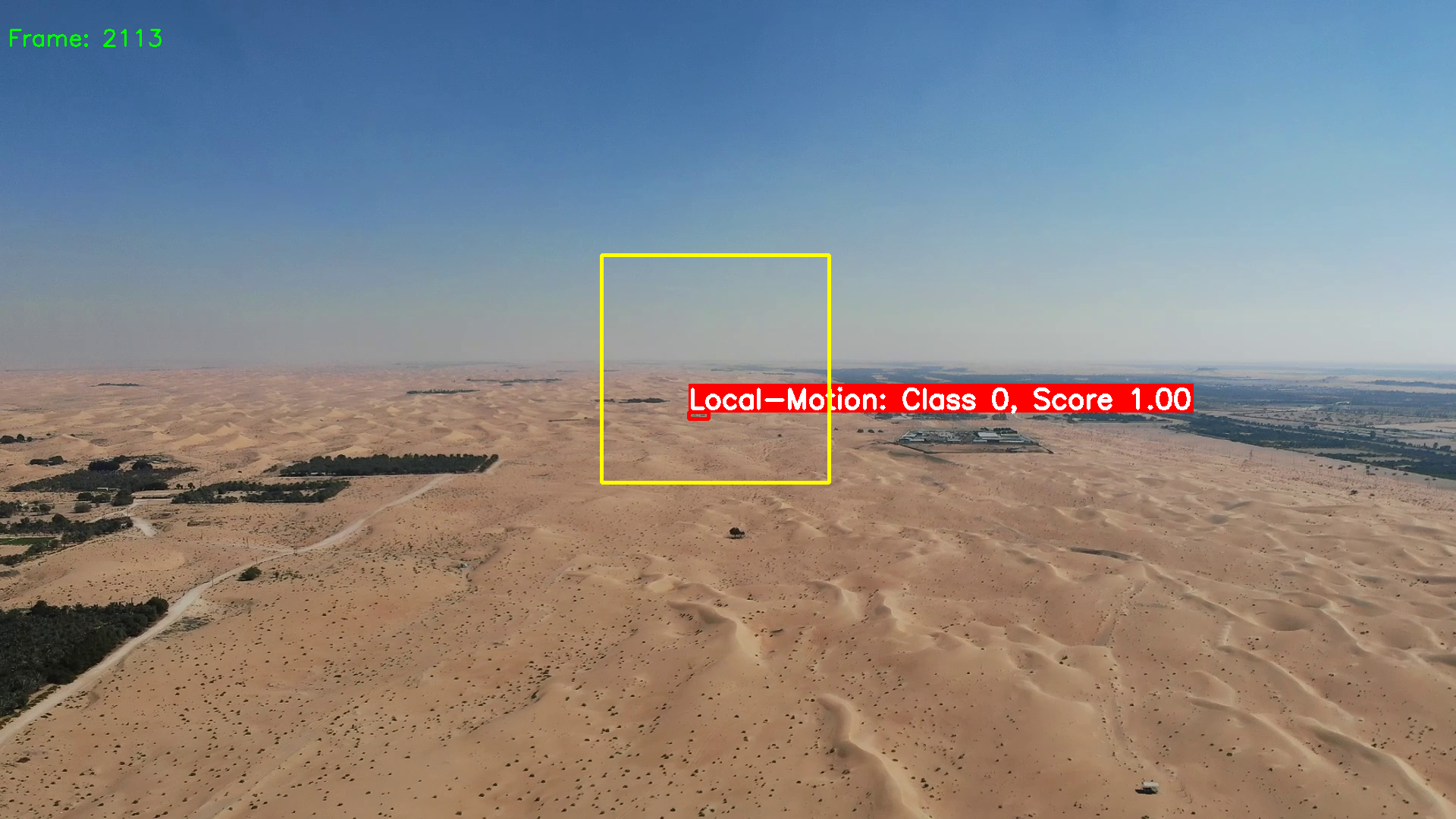} \\
    \end{tabular}
    
    \caption{Comparison of typical detection result samples achieved by various methods, with each row presenting the output of a different detection method.}
    \label{Fig9}
\end{figure*}

Our study quantitatively compares the performance of the proposed GL-YOMO algorithm with several state-of-the-art methods across various datasets, with results detailed in Table \ref{tab2} and Table \ref{tab3}. As shown in Table \ref{tab2}, our approach excels across multiple metrics on the Drone-vs-Bird dataset,  with a Recall 4.8\% higher than the second-best performer, RT-DETR \cite{RT-DETR}. Other metrics also demonstrate superior performance compared to other methods. In Table \ref{tab3}, the performance of our method is even more pronounced on the Fixed-Wing dataset, with a 23.7\% improvement in F1 score and a 25.1\% increase in AP over RT-DETR.

\begin{table}[htbp!]
\centering
\caption{Comparison of the GL-YOMO with state-of-the-art methods on Drone-vs-Bird dateset}
\label{tab2}
\begin{tabular}{c|c|c|c|c}
\hline
\textbf{Method} & \textbf{Precision} & \textbf{Recall} & \textbf{F1-score} & \textbf{AP} \\
\hline
YOLOv5s         & 0.824              & 0.756           & 0.789             & 0.761       \\

YOLOv8s         & 0.837              & 0.682           & 0.752             & 0.744       \\

YOLOv10s        & 0.822              & 0.662           & 0.734             & 0.702       \\

TPH-YOLOv5     & 0.821              & 0.671           & 0.739             & 0.718       \\

RT-DETR         & 0.859              & 0.761           & 0.807             & \textbf{0.773}       \\

GL-YOMO         & \textbf{0.886}     & \textbf{0.809}  & \textbf{0.846}    & 0.763  \\
\hline
\end{tabular}
\end{table}

\begin{table}[h!]
\centering
\caption{Comparison of the GL-YOMO with state-of-the-art methods on Fixes-wings dateset}
\label{tab3}
\begin{tabular}{c|c|c|c|c}
\hline
\textbf{Method} & \textbf{Precision} & \textbf{Recall} & \textbf{F1-score} & \textbf{AP} \\
\hline
YOLOv5s         & 0.640              & 0.405           & 0.496             & 0.382  \\

YOLOv8s         & 0.663              & 0.177           & 0.280             & 0.207  \\

YOLOv10s        & 0.706              & 0.175           & 0.280             & 0.196  \\

TPH-YOLOv5        & 0.613              & 0.353           & 0.448             & 0.335  \\

RT-DETR         & 0.826     & 0.577           & 0.680             & 0.571  \\

GL-YOMO         & \textbf{0.981}         & \textbf{0.861}     & \textbf{0.917}     & \textbf{0.822}       \\
\hline
\end{tabular}
\end{table}

\begin{table*}[h]
\centering
\caption{Performance Comparison of Different YOLO Models on Global Images}
\label{tab4}
\begin{tabular}{c|c|c|c|c|c|c|c|c|c}
\hline
Model          & p & g & a & Model Size & GFLOPs & Precision & Recall & mAP50  & mAP95 \\
\hline
YOLOv5s        &   &   &   & 14.4M      & 15.8   & 0.834    & 0.637  & 0.768  & 0.423 \\

YOLOv8s        &   &   &   & 21.5M      & 28.4   & 0.84     & 0.377  & 0.448  & 0.241 \\

YOLOv10s       &   &   &   & 16.5M      & 24.4   & 0.839    & 0.379  & 0.432  & 0.227 \\

YOLOv5s-p       & \checkmark &   &   & 25.0M      & 16.1   & 0.842    & 0.671  & 0.788  & 0.425 \\

YOLOv5s-pa     & \checkmark &   & \checkmark & 26.0M      & 18.6   & 0.827    & 0.650  & 0.774  & 0.431 \\

YOLOv8s-pg     & \checkmark & \checkmark &   & 10.6M      & 22.0   & 0.862    & 0.592  & 0.649  & 0.327 \\

YOLOv10-pg     & \checkmark & \checkmark &   & 13.0M      & 25.1   & 0.835    & 0.569  & 0.628  & 0.317 \\

YOLOv8s-pga    & \checkmark & \checkmark & \checkmark & \textbf{9.4M}       & 22.8   & \textbf{0.863}    & 0.597  & 0.664  & 0.343 \\

YOLO Detector    & \checkmark & \checkmark & \checkmark & 14.3M      & \textbf{10.6}   & 0.844    & \textbf{0.681}  & \textbf{0.794}  & \textbf{0.433} \\
\hline
\end{tabular}
\end{table*}

Analysis indicates that other methods exhibit significant shortcomings in detecting small objects, leading to a marked decline in recall, F1-score, and AP. This issue primarily arises from information loss caused by downsampling techniques, which adversely affect detection accuracy.  In contrast, the GL-YOMO algorithm markedly enhances small object detection capabilities by innovatively integrating appearance and motion features.  Additionally, the global-local synergistic strategy of GL-YOMO effectively preserves critical visual information of small objects and significantly reduces interference from complex backgrounds, thereby substantially improving detection performance.  

As shown in Fig. \ref{Fig9}, the GL-YOMO methods demonstrate exceptional performance in detecting extremely small targets, with the majority of targets being effectively identified through local detection strategies.  Moreover, it proves highly effective in detecting particularly challenging targets.  For instance, as shown in the fourth column of images, other methods fail to detect the target, while our approach successfully identifies the correct target with precision. Overall, the experimental results from both datasets underscore the effectiveness of the GL-YOMO algorithm in small object detection tasks.

\begin{table}[h!]
\centering
\caption{Performance Comparison of Different Methods on Local Images}
\label{tab5}
\begin{tabular}{c|c|c|c|c}
\hline
Method          & Precision & Recall & mAP50 & mAP50-95 \\
\hline
YOLOv5s          & \textbf{0.912}     & 0.804  & 0.873  & 0.491 \\

YOLOv8s           & 0.898     & 0.804  & 0.826  & 0.442 \\

YOLOv8s-pga   & 0.894 & \textbf{0.816} & 0.836 & 0.450 \\

YOLO Detector     & 0.910 & 0.811 & \textbf{0.880} & \textbf{0.503} \\
\hline
\end{tabular}
\end{table}

\begin{figure*}[htbp]
    \centering
    \setlength{\tabcolsep}{2pt}  
    \renewcommand{\arraystretch}{1.5}  
    \begin{tabular}{cccc}  
        \includegraphics[width=0.24\textwidth]{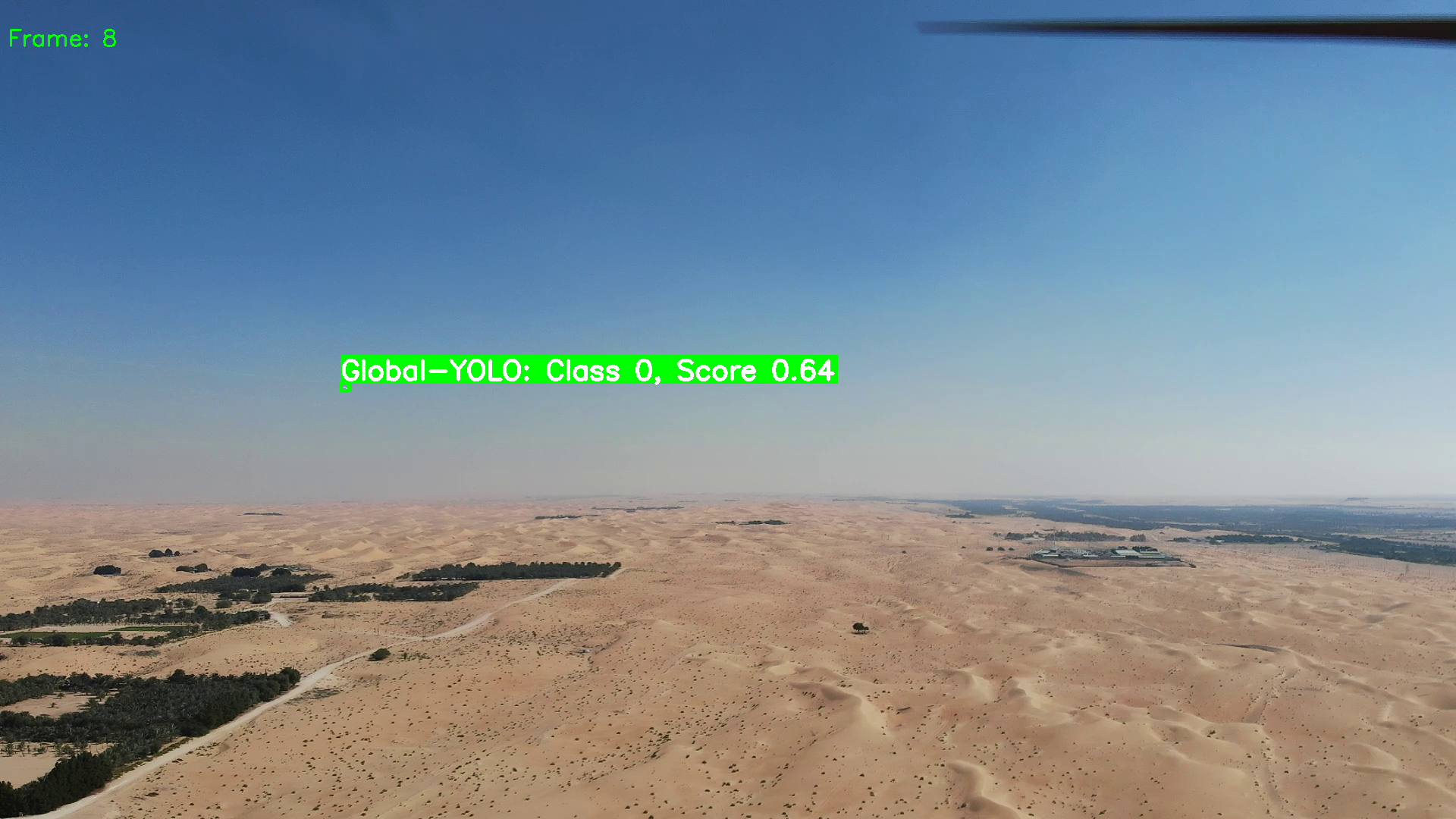} &
        \includegraphics[width=0.24\textwidth]{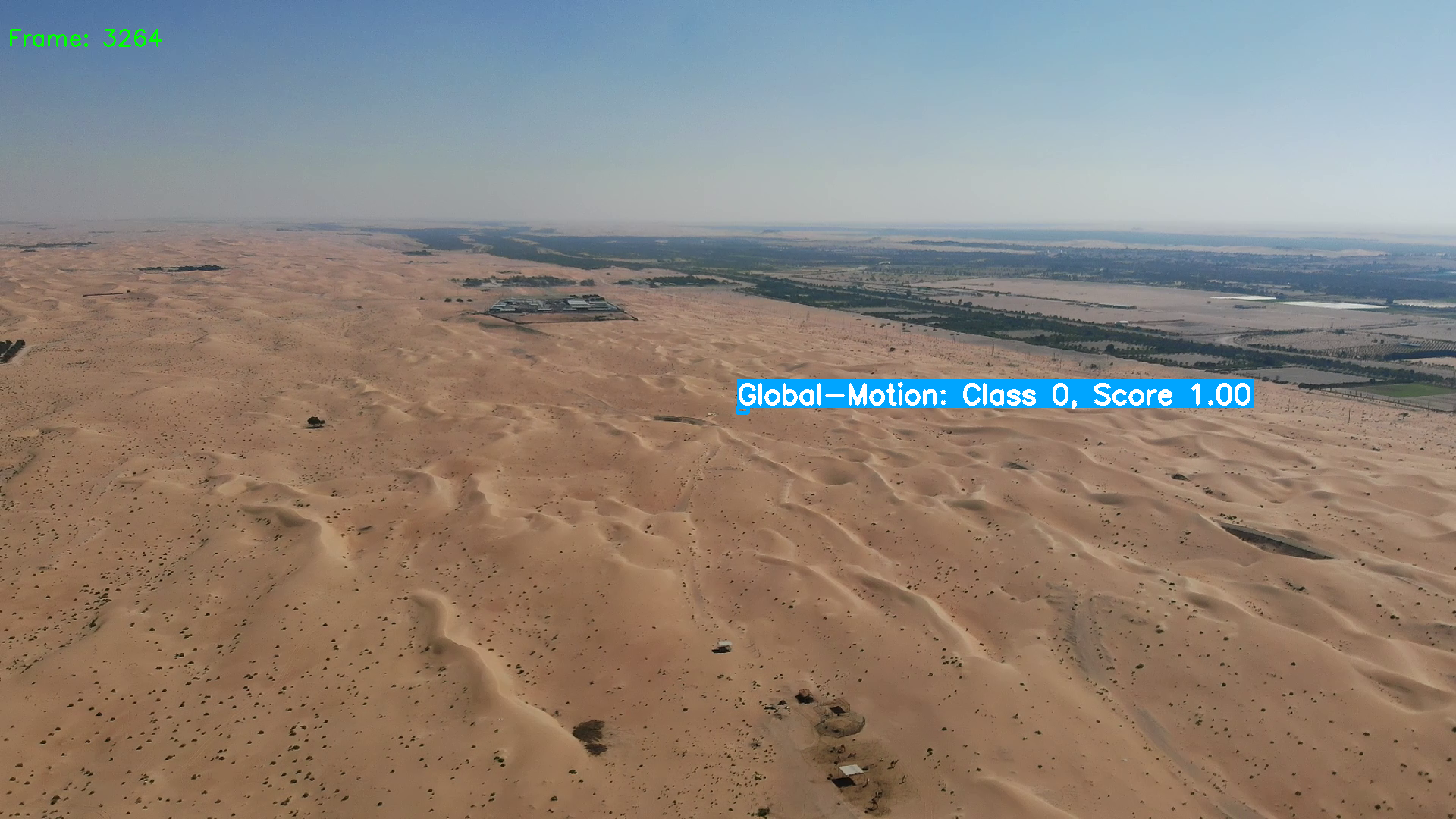} &
        \includegraphics[width=0.24\textwidth]{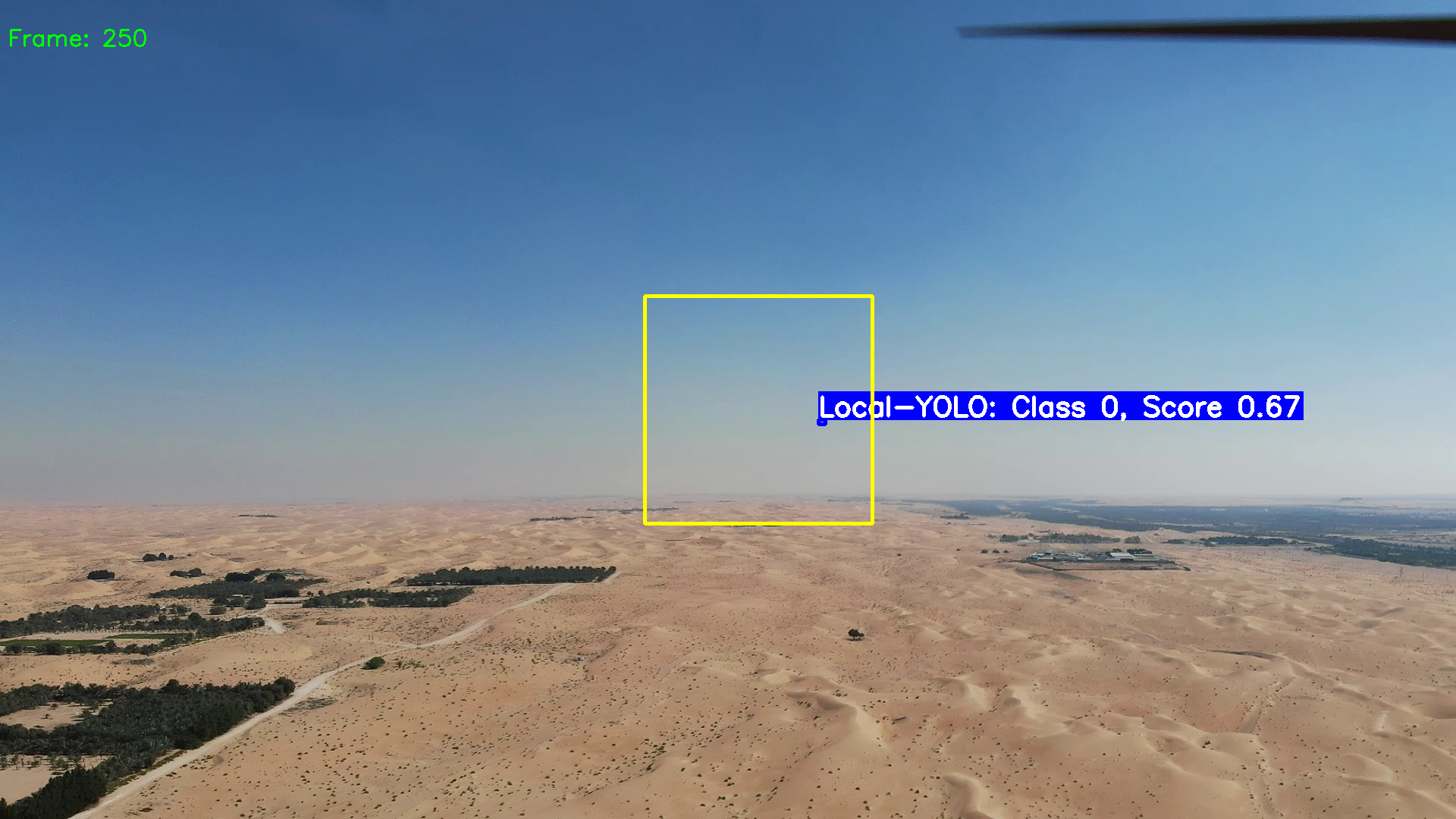} &
        \includegraphics[width=0.24\textwidth]{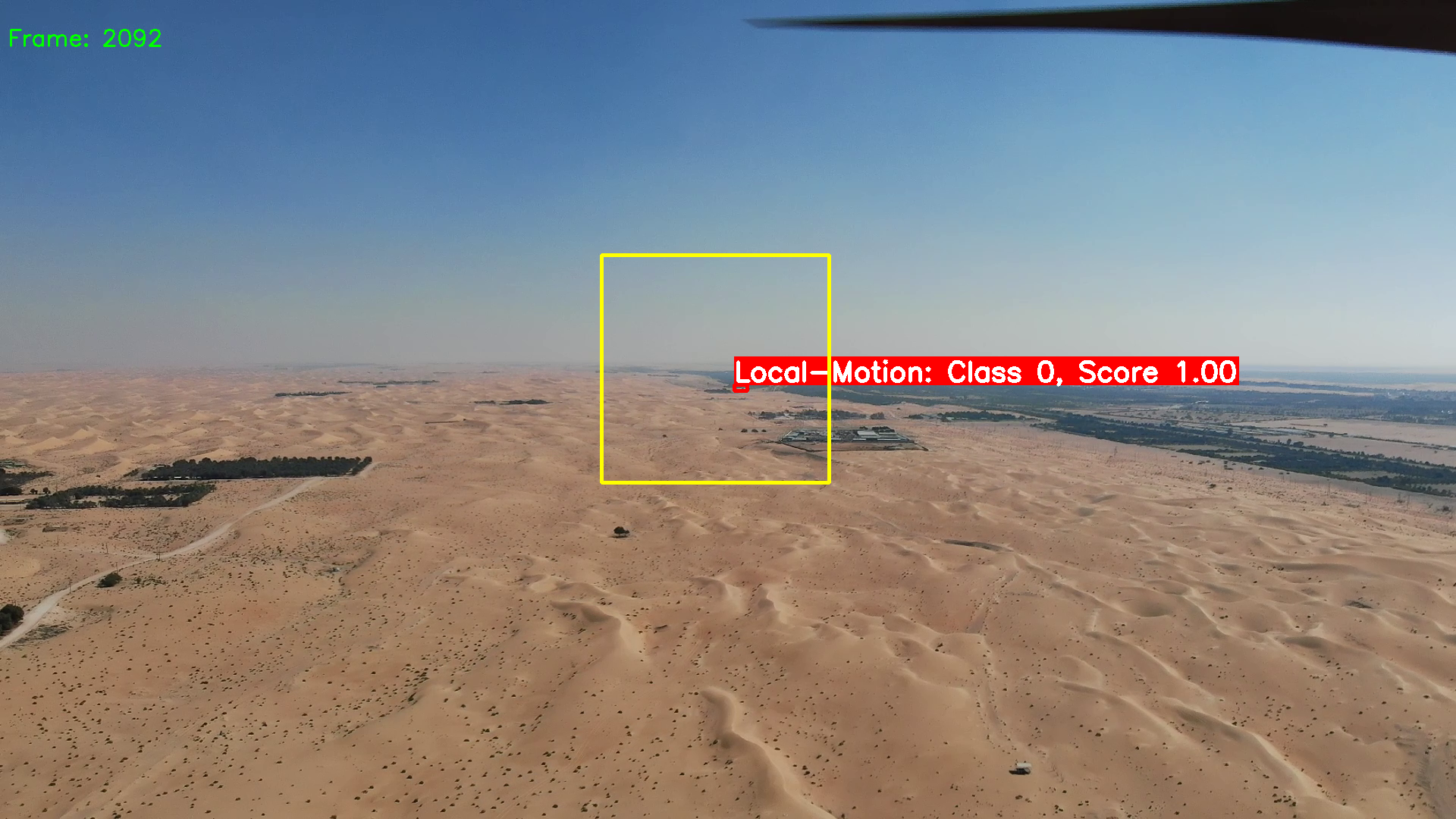} \\
        
        \includegraphics[width=0.24\textwidth]{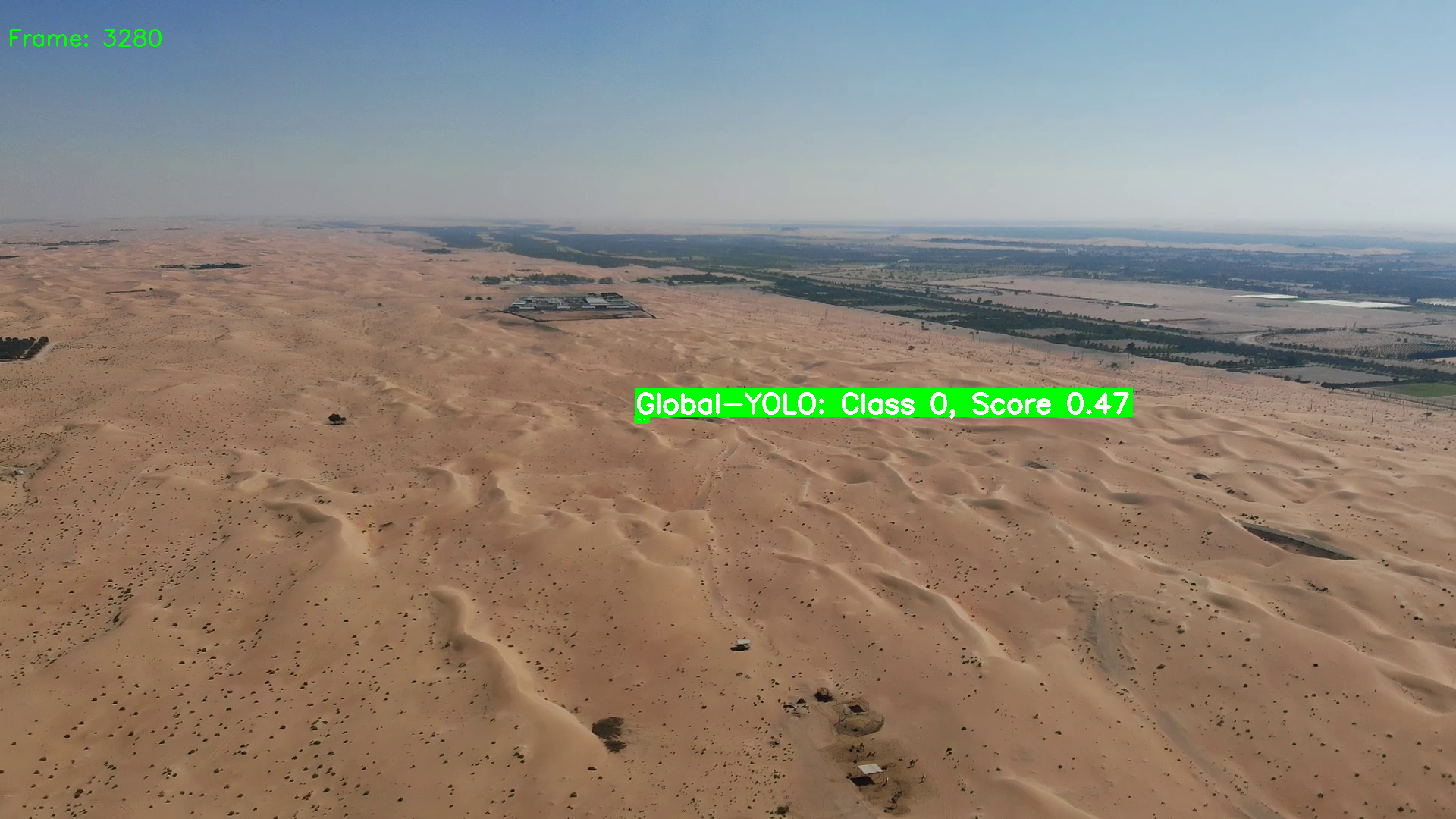} &
        \includegraphics[width=0.24\textwidth]{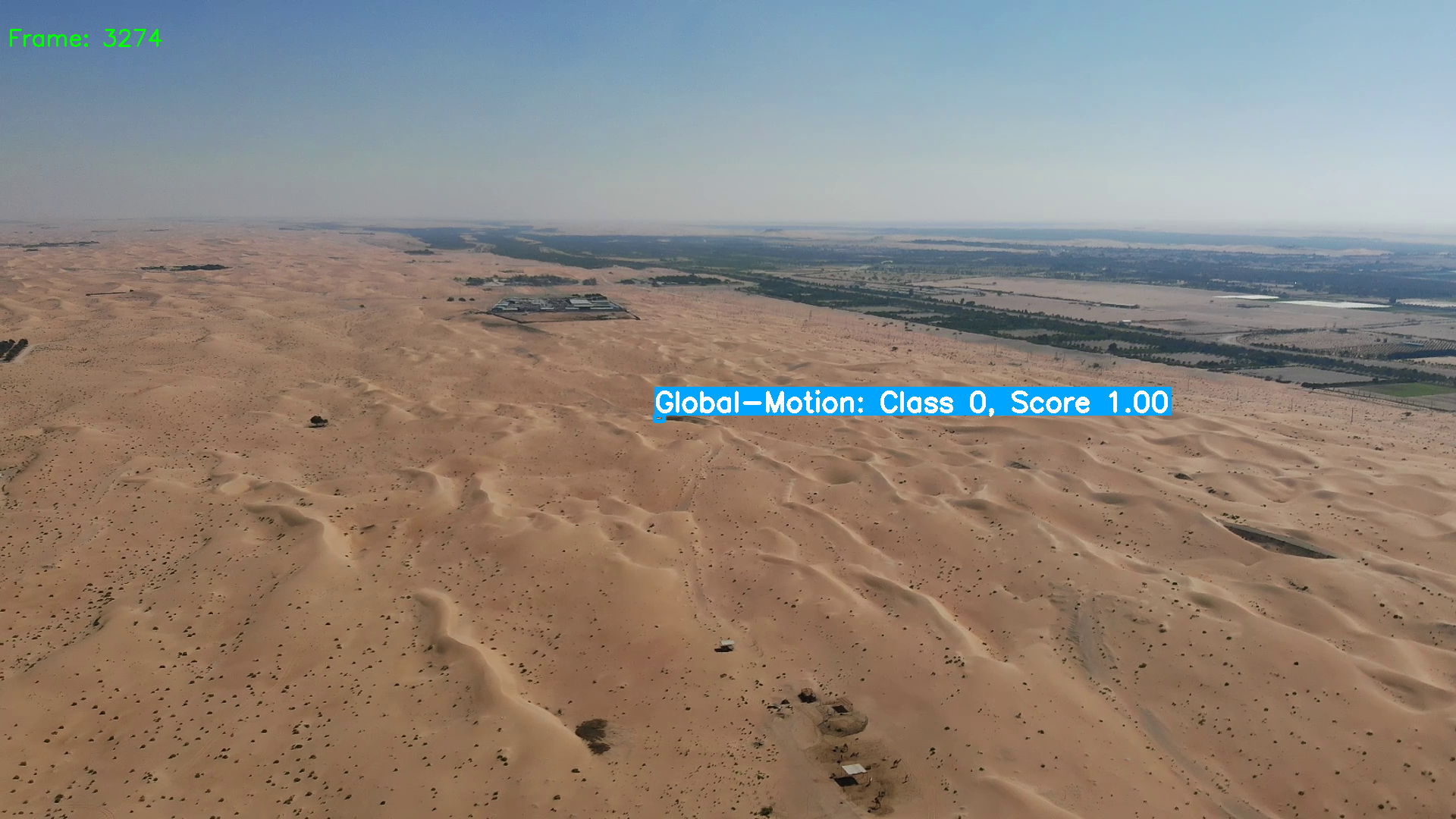} &
        \includegraphics[width=0.24\textwidth]{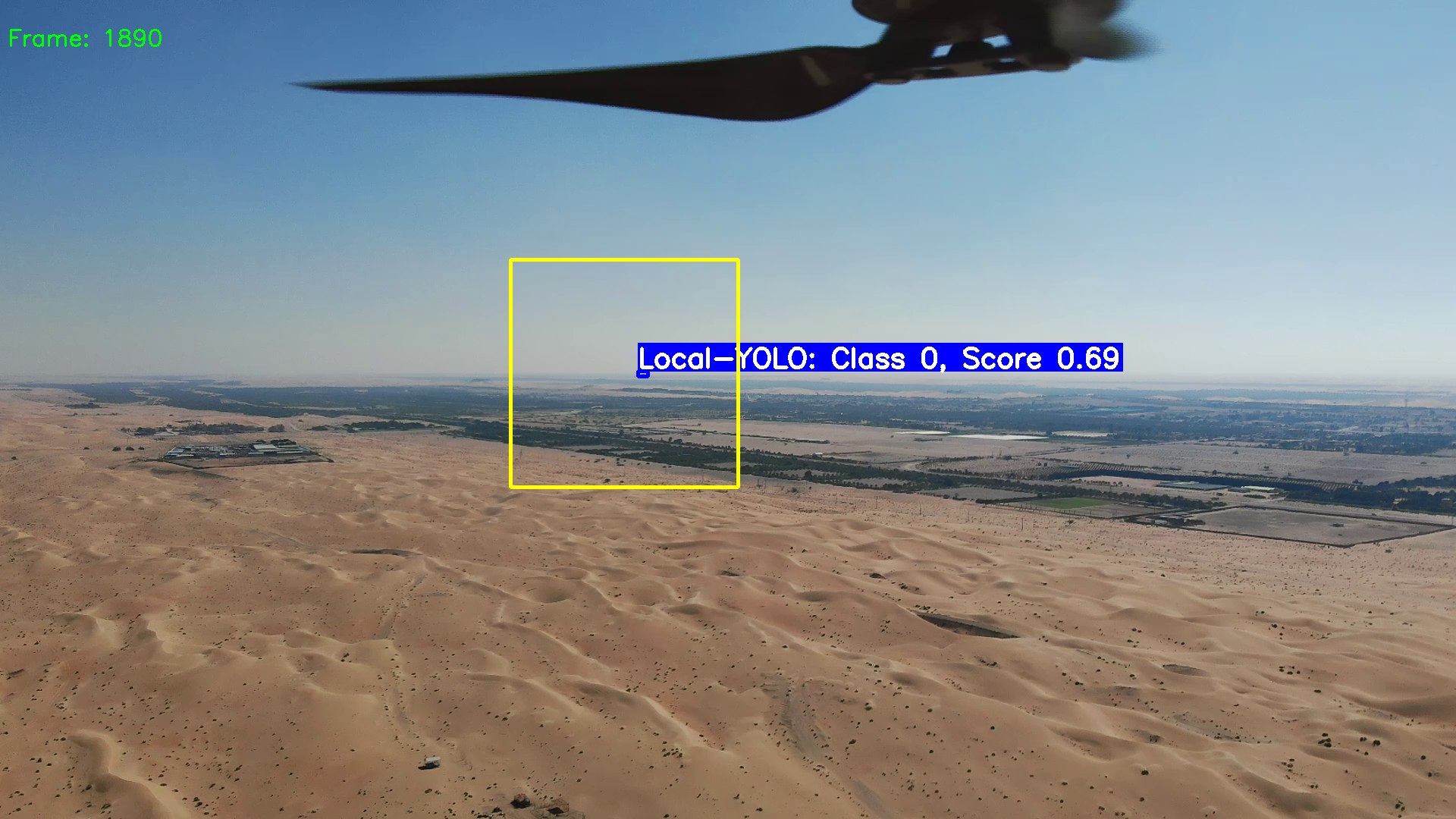} &
        \includegraphics[width=0.24\textwidth]{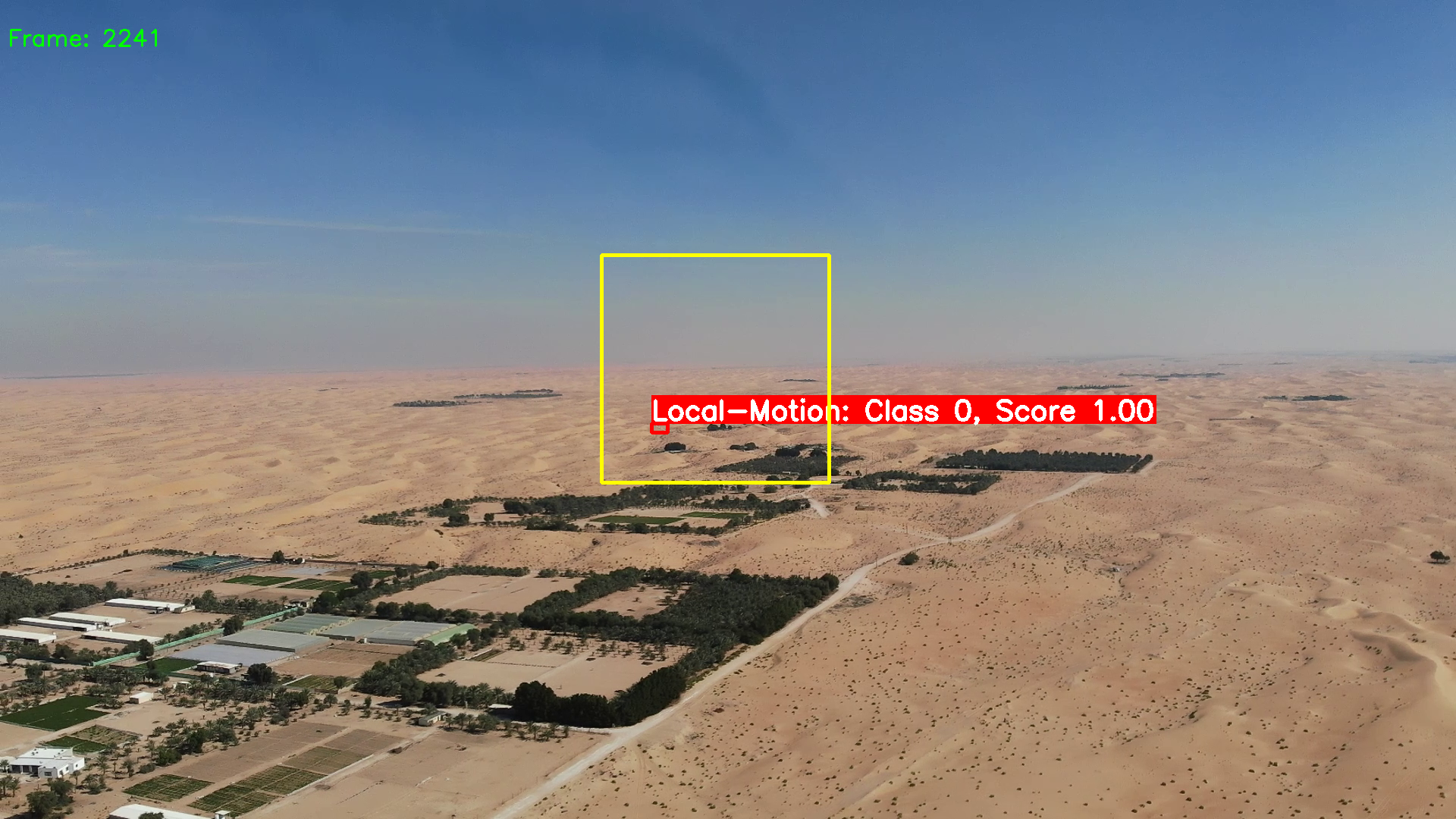} \\
        
        \includegraphics[width=0.24\textwidth]{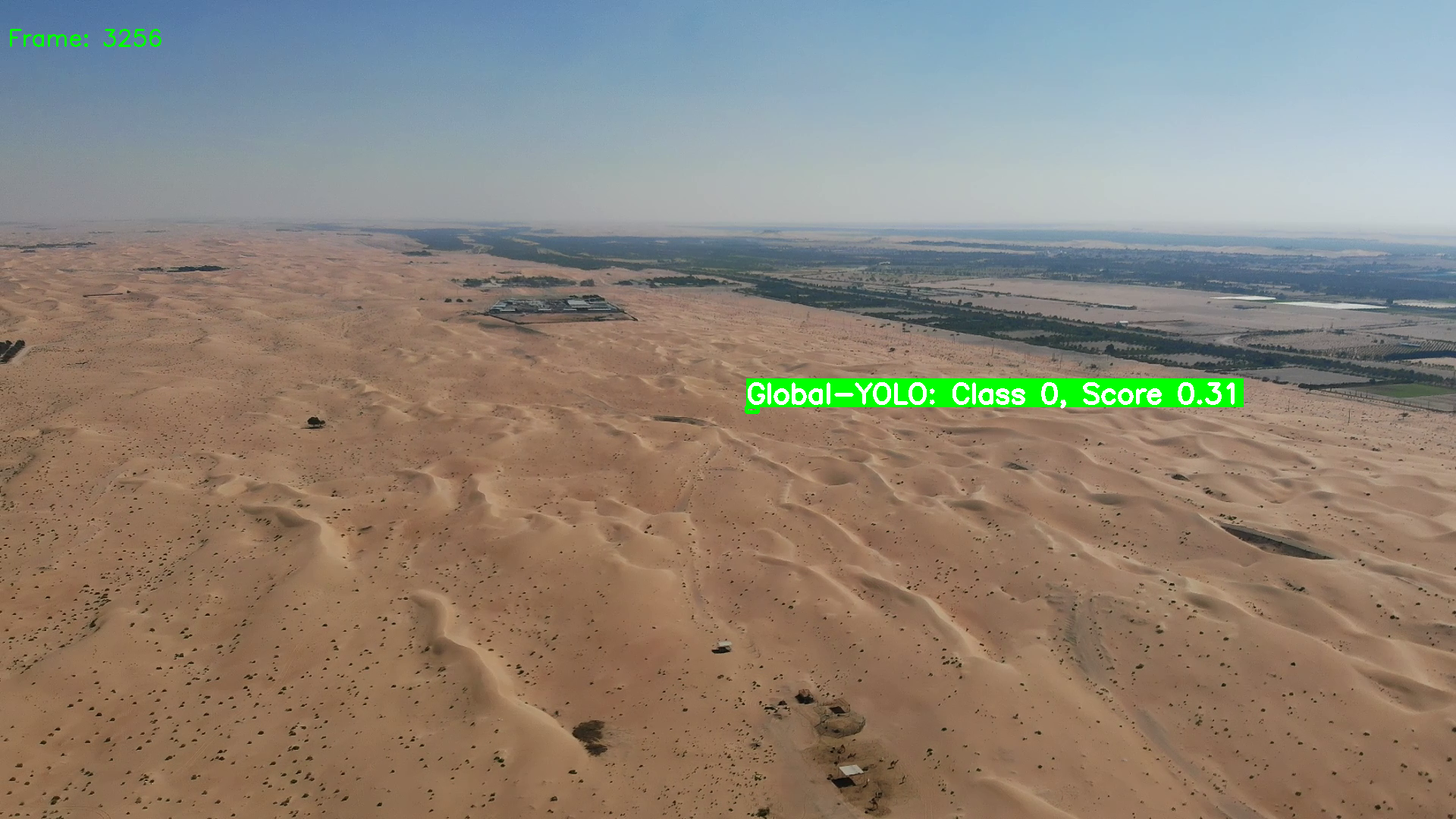} &
        \includegraphics[width=0.24\textwidth]{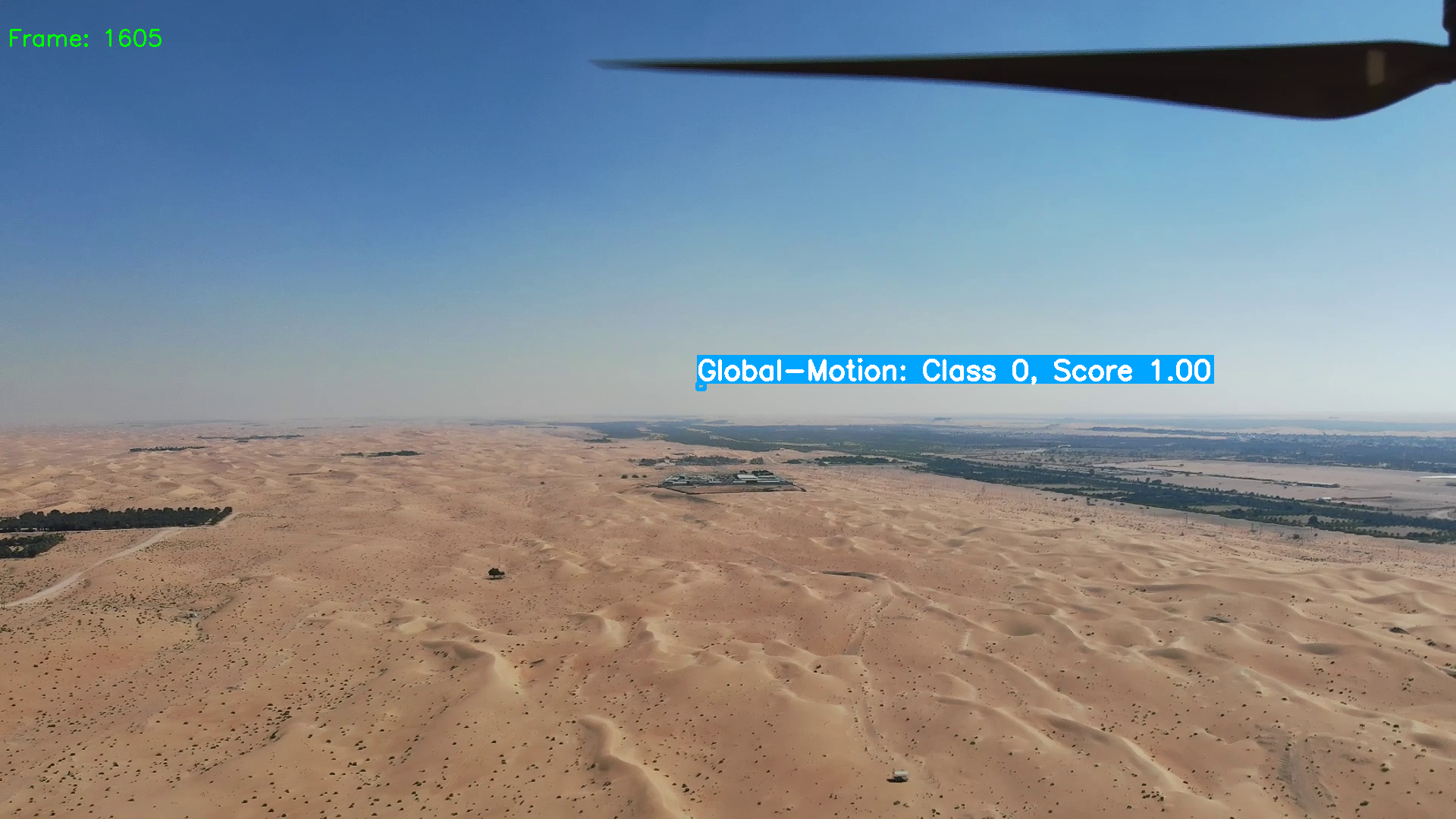} &
        \includegraphics[width=0.24\textwidth]{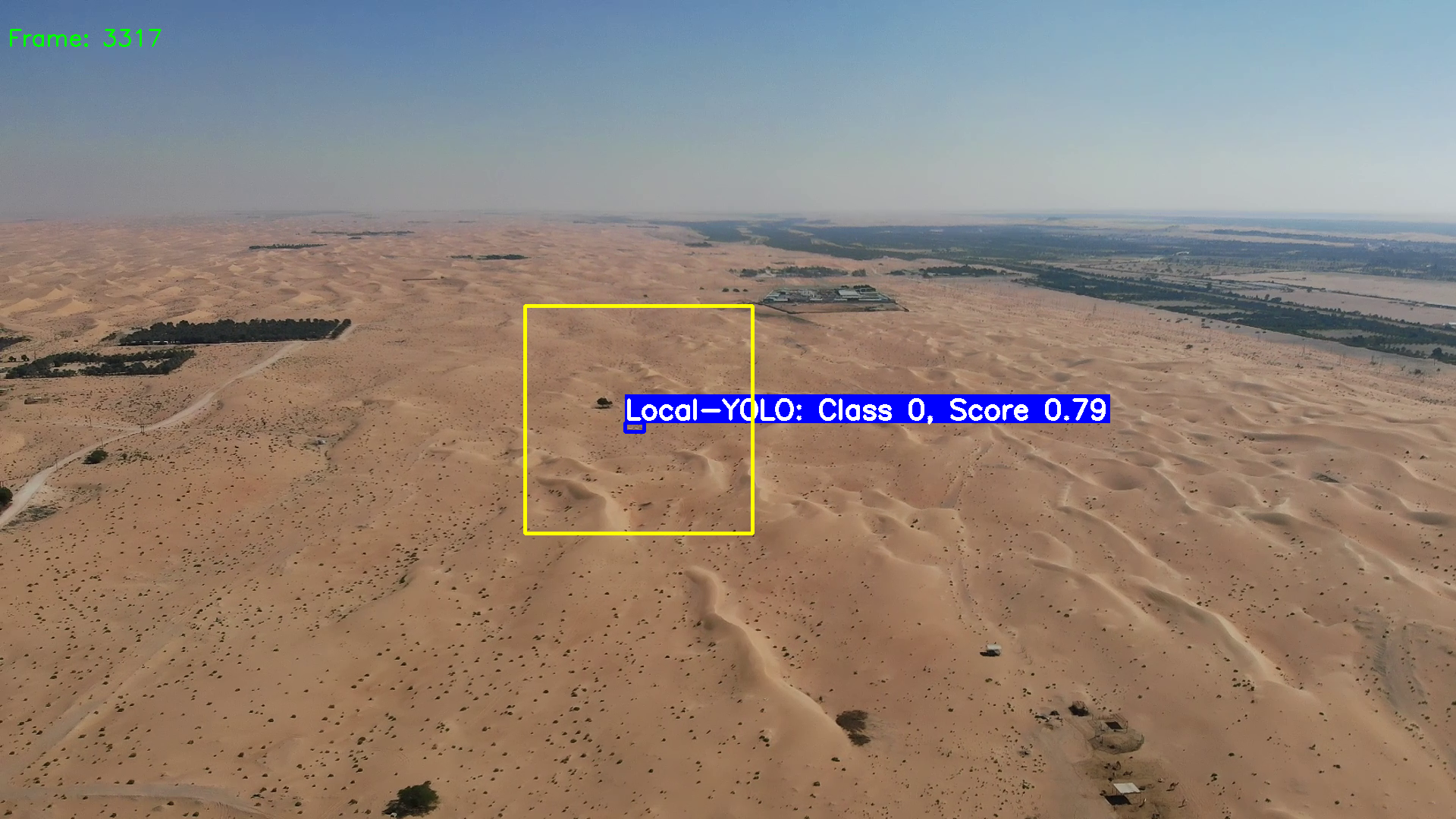} &
        \includegraphics[width=0.24\textwidth]{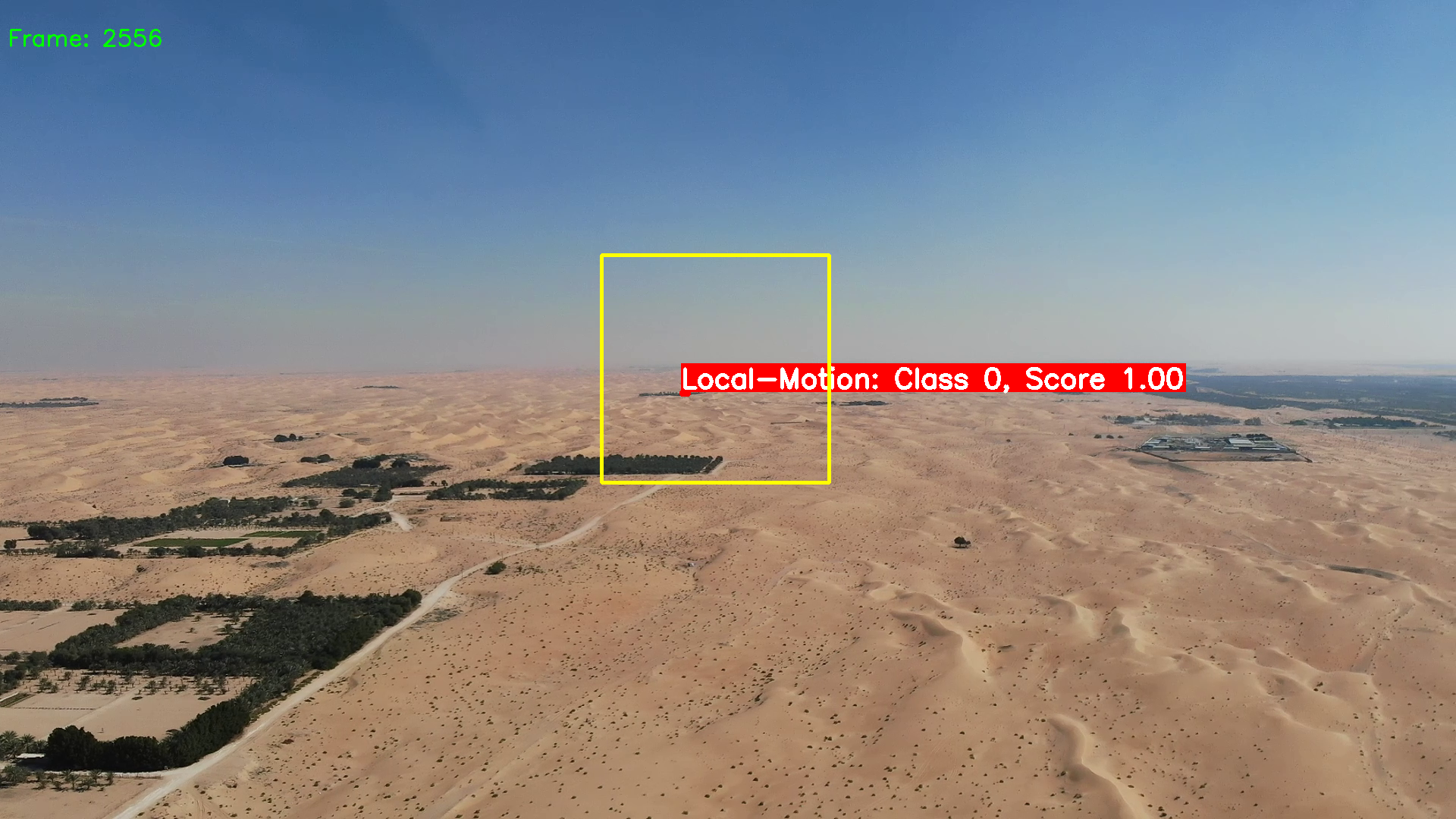} \\
        
        \includegraphics[width=0.24\textwidth]{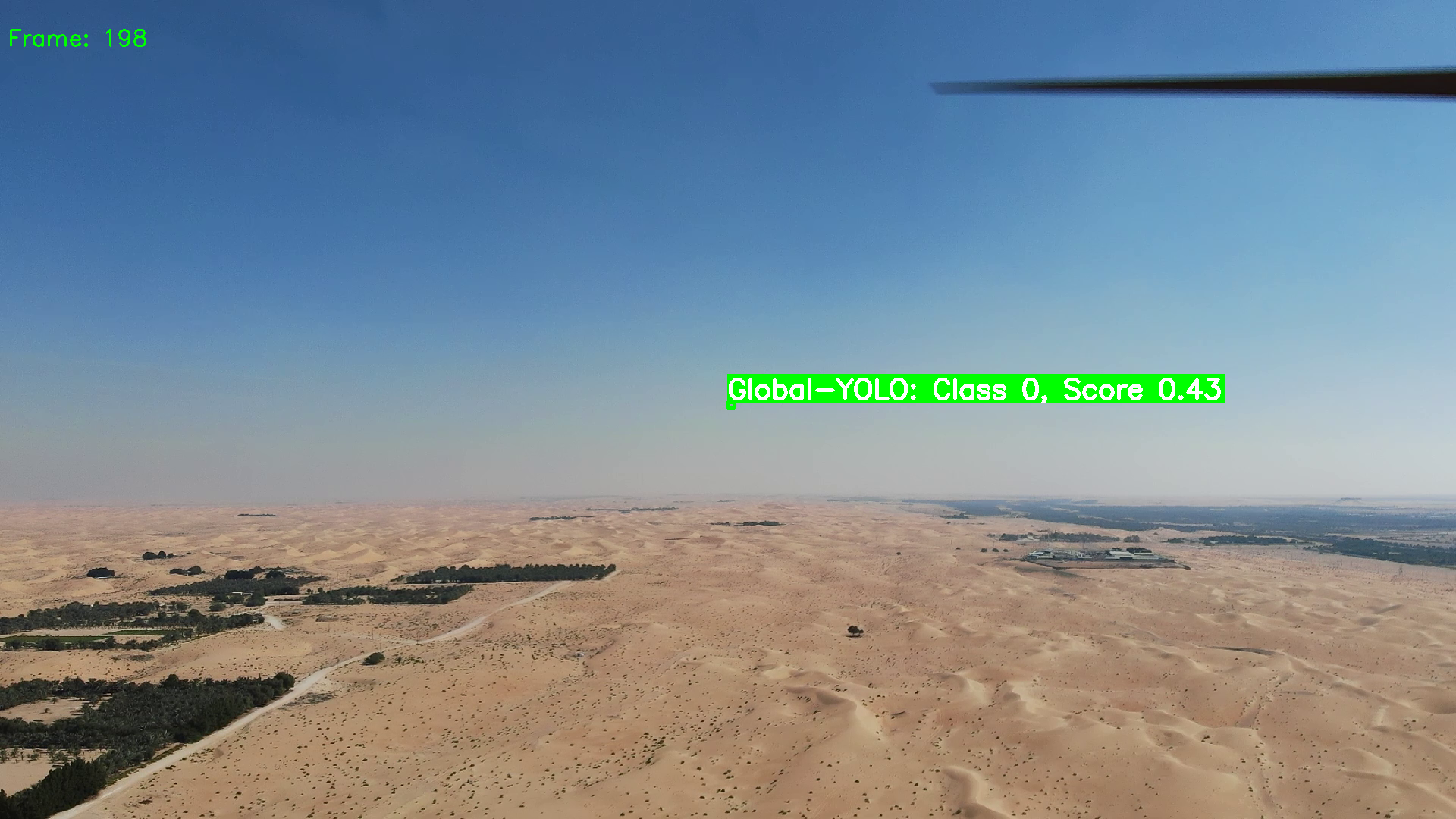} &
        \includegraphics[width=0.24\textwidth]{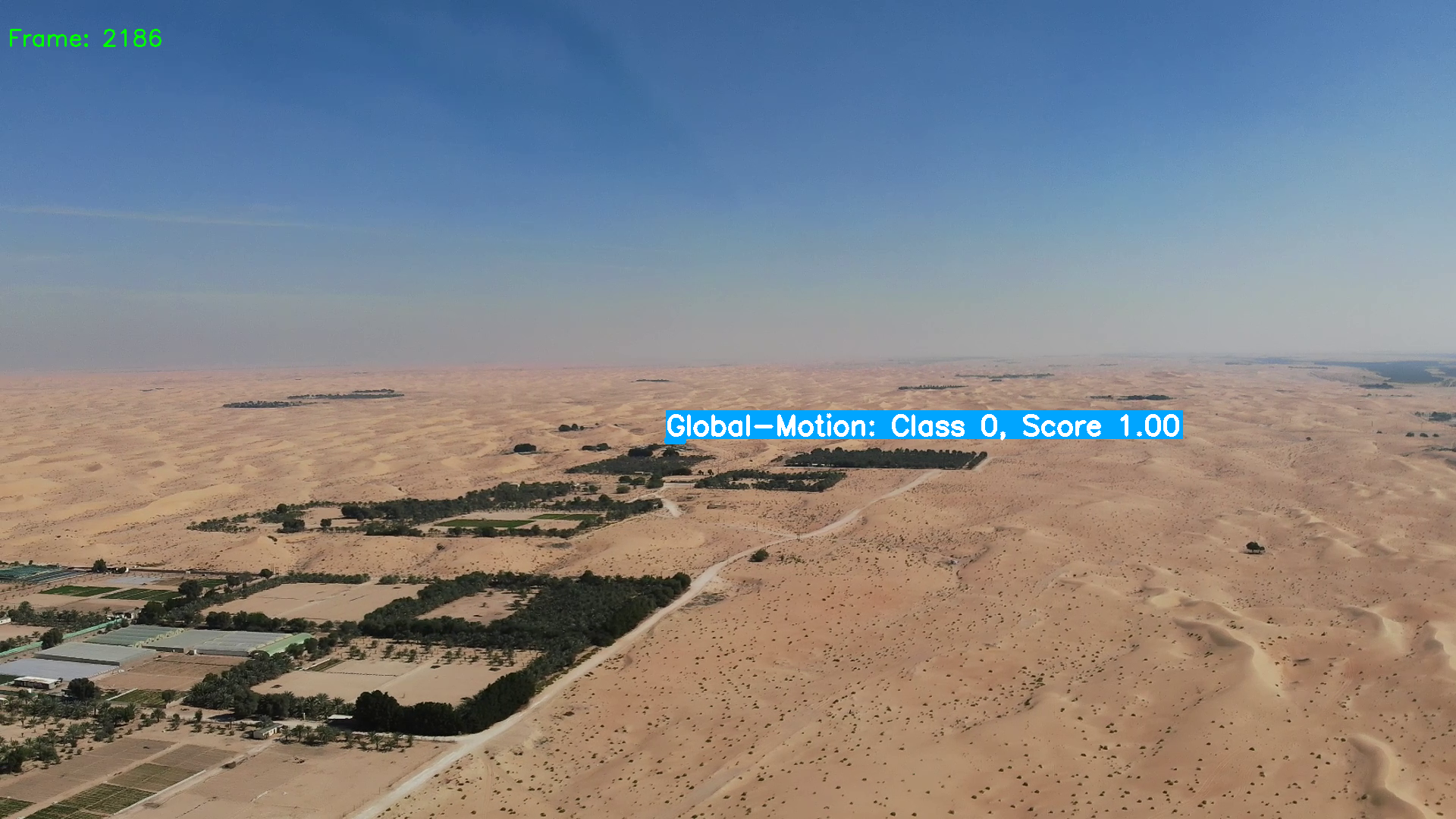} &
        \includegraphics[width=0.24\textwidth]{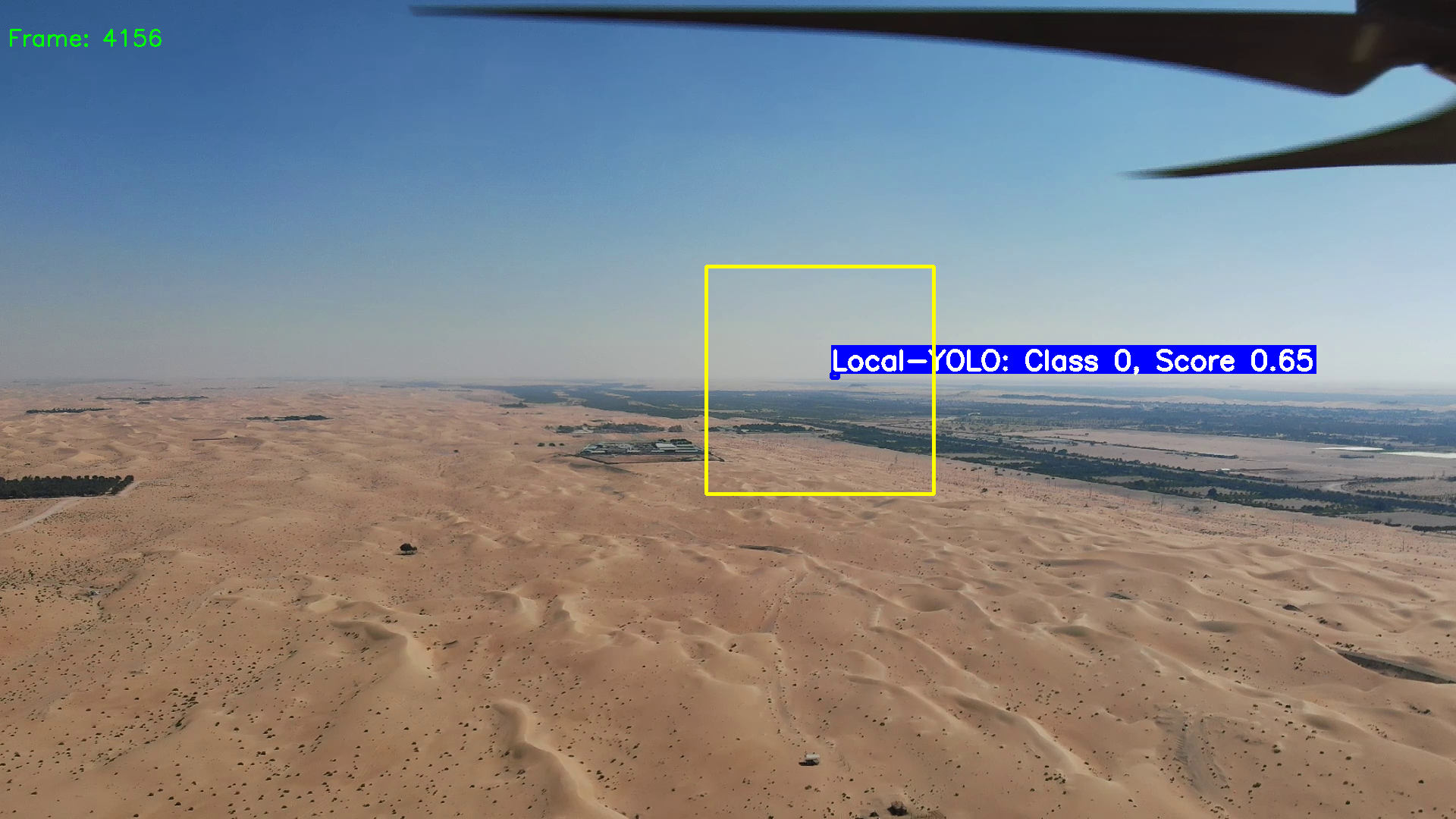} &
        \includegraphics[width=0.24\textwidth]{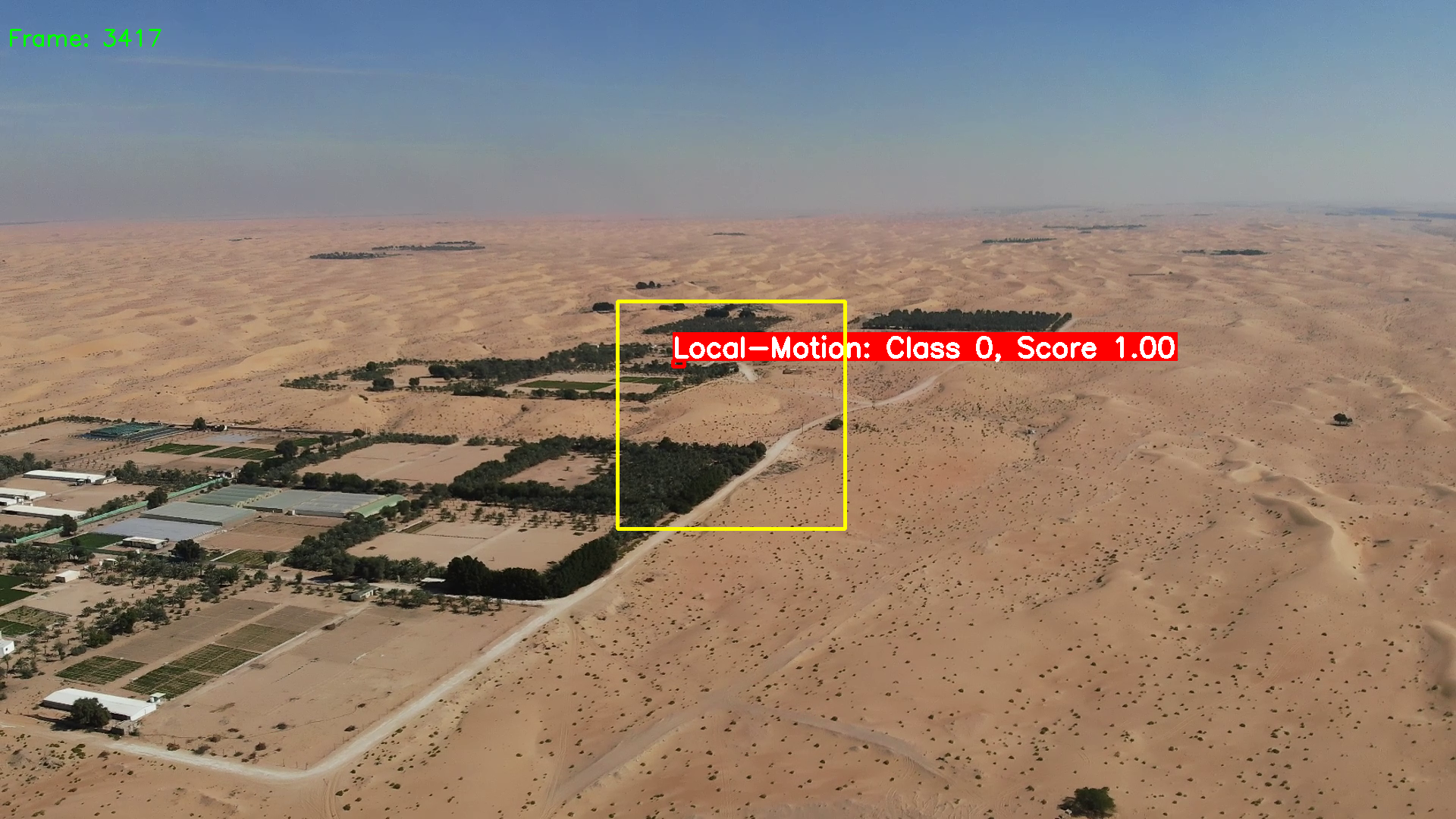} \\
        
        \multicolumn{1}{c}{\small{(a) Global-YOLO detection}} &
        \multicolumn{1}{c}{\small{(b) Global-Motion detection}} &
        \multicolumn{1}{c}{\small{(c) Local-YOLO detection}} &
        \multicolumn{1}{c}{\small{(d) Local-Motion detection}} \\
    \end{tabular}
    
    \caption{Some typical sample results from the Fixed-Wings dataset: Four different colors are used to distinguish the various detection outcomes: Global-YOLO results are displayed in green, Global-Motion results in light blue, Local-YOLO detections in dark blue, and Local-Motion results in red.}
    \label{Fig10}
\end{figure*}

\subsection{Ablation Experiment}
\subsubsection{YOLO Improvement Experiment}

To validate the performance of the improved YOLO Detector for small object detection, we conducted a comprehensive comparison with several state-of-the-art methods.  TABLE \ref{tab4} presents the performance metrics of each model under the complete image, where `P' denotes the addition of a detection head, `g' indicates the use of the Ghost module for lightweight design, and `a' represents the incorporation of multi-scale fusion features and attention mechanisms.

The optimized YOLO Detector significantly outperforms baseline models YOLOv5s, YOLOv8s, and YOLOv10s in recall, mAP50, and mAP50-95. The integration of a small object detection head effectively reduces information loss from downsampling, boosting recall by 4.4\% and mAP by 2.6\%. Attention mechanisms and multi-scale feature fusion further improve performance, with a 2\% increase in mAP50-95. The Ghost module enhances efficiency, cutting model size to 14.3M while reducing parameters and GFLOPs, making it ideal for resource-limited environments. Overall, the improved YOLO Detector excels in small object detection with strong results across recall, mAP, and model size.

Furthermore, TABLE \ref{tab5} presents a performance analysis of object detection on local image datasets, showcasing significant improvements with our YOLO Detector. Notably, it achieves the highest mAP50 of 0.880 and the best mAP50-95 score of 0.503, along with a 0.6\% increase in Precision compared to YOLOv8s-pga. While there are slight differences in Recall between the YOLO Detector and YOLOv8s-pga, our detector consistently outperforms in other metrics. Compared to global image detection, local image detection exhibits clear advantages, underscoring its potential for small object detection tasks and validating our approach's effectiveness in handling images with reduced background complexity.

\subsubsection{Motion Detection Experiment}

Table \ref{tab6} shows the effectiveness of the Motion Detector. Where G-YO refers to YOLO detection performed globally; G-YOMO refers to global YOLO combined with motion detection; GL-YO denotes YOLO detection at both global and local levels; and GL-YOMO represents global and local YOLO with motion detection. The results indicate that motion detection augments the sensitivity at the cost of some extra false positives, leading to a slight reduction in precision. However, this trade-off is compensated by significant improvements in other metrics. In global detection, adding a motion detector improved recall, F1-score, and AP by 5.7\%, 3.9\%, and 1.2\%, respectively. In local detection, the addition of a motion detector also brought notable gains, with increases of 3.2\% in recall and 1.5\% in F1-score. These results highlight the importance and effectiveness of motion detectors in object detection tasks, significantly enhancing performance in both global and global-local collaborative detection modes.

\begin{table}[htbp!]
\centering
\caption{Impact of Motion Detector Integration on Detection Performance}
\label{tab6}
\begin{tabular}{c|c|c|c|c}
\hline
Method & Precision & Recall & F1-Score & AP      \\
\hline
G-YO & 0.968 & 0.560 & 0.709 & 0.553 \\

G-YOMO & 0.948 & 0.617 & 0.748 & 0.565 \\

GL-YO & \textbf{0.989} & 0.829 & 0.902 & \textbf{0.828} \\

GL-YOMO & 0.981 & \textbf{0.861} & \textbf{0.917} & 0.822 \\
\hline
\end{tabular}
\end{table}

\subsection{Inference Time}

To comprehensively evaluate the deployment performance of our method on edge computing devices, we selected the NVIDIA Jetson Xavier NX as our testing platform.  By leveraging TensorRT for model optimization and acceleration, we achieved an average frame rate of 21.6 FPS using a $640\times640$ resolution model for inference on the test video, meeting the requirements for real-time applications. Fig. \ref{Fig10} illustrates some typical detection results of our method on the Fixed-Wings dataset.

\section{Conclusions}
In this paper, we proposed GL-YOMO, an advanced real-time detection method designed to tackle the challenges of detecting small-pixel UAV targets at long distances. GL-YOMO combines the YOLO detection algorithm with multi-frame motion analysis to achieve high-precision UAV detection. Our approach utilizes a global and local collaborative detection strategy: initially performing broad target localization, followed by an adaptive mechanism that refines the detection range for more precise local analysis. This method not only improves detection accuracy but also allows seamless transitions back to global detection when local detection struggles, enhancing overall system performance. Future work will focus on optimizing GL-YOMO for multi-UAV detection to handle more complex environments.

\bibliographystyle{IEEEtran}
\bibliography{GL-YOMO_UAV_Detection}

\end{document}